\documentclass{article}

\PassOptionsToPackage{numbers, compress}{natbib}

 \usepackage[eandd, final]{neurips_2026}

\usepackage[utf8]{inputenc}
\usepackage[T1]{fontenc}
\usepackage{microtype}

\usepackage{amsmath}
\usepackage{amssymb}
\usepackage{amsfonts}
\usepackage{amsthm}
\usepackage{mathtools}
\usepackage{nicefrac}

\usepackage{graphicx}
\usepackage{subcaption}
\usepackage{booktabs}
\usepackage{array}
\usepackage{multirow}
\usepackage{wrapfig}
\usepackage{enumitem}
\usepackage{pgfplots}
\pgfplotsset{compat=1.17}

\usepackage[dvipsnames,table]{xcolor}
\usepackage{pifont}
\usepackage[most]{tcolorbox}
\tcbuselibrary{breakable}
\usepackage{mdframed}
\usepackage{listings}

\usepackage{url}
\usepackage{hyperref}
\usepackage[capitalize,noabbrev]{cleveref}

\definecolor{mydarkblue}{rgb}{0,0.08,0.45}
\definecolor{darkgreen}{RGB}{50,100,0}
\definecolor{myblue}{RGB}{235,235,250}
\definecolor{lightblue}{RGB}{230, 235, 245}
\definecolor{lightgray}{RGB}{240, 240, 240}
\definecolor{darkgray}{RGB}{220, 220, 220}
\definecolor{superlightred}{rgb}{0.99, 0.92, 0.92}
\definecolor{darkred}{RGB}{230, 150, 170}

\theoremstyle{definition}

\theoremstyle{plain}

\newcommand{\name}{OmniCapBench}
\newcommand{\fullname}{Omni-Video Caption Benchmark}

\definecolor{uclablue}{RGB}{159, 195, 224}

\definecolor{uclagold}{RGB}{156, 172, 234}

\definecolor{grayred}{RGB}{30, 140, 224}

\definecolor{cyan}{rgb}{0.573, 0.675, 0.878}
\definecolor{limegreen}{rgb}{0.675, 0.843, 0.557}
\definecolor{table_red}{RGB}{255,52,88}
\definecolor{table_green}{RGB}{8,228,198}
\definecolor{table_gray}{RGB}{156,156,156}
\definecolor{morandigray}{RGB}{155,175,185}
\definecolor{morandipink}{RGB}{235,205,215}
\definecolor{casegreen}{RGB}{117, 189, 67}

\newcommand{\cmark}{\ding{51}}
\newcommand{\xmark}{\ding{55}}

\definecolor{iccvblue}{rgb}{0.16, 0.54, 0.86}
\newtcolorbox{ttcolorbox}[1][]{
  colframe=iccvblue,
  colback=iccvblue!5!white,
  title=#1,
  fonttitle=\bfseries\sffamily,
}

\newtcolorbox{remarkbox}[1][]{
  colback=uclagold!10,
  colframe=uclagold!80!black,
  coltitle=white,
  title=#1,
  fonttitle=\bfseries,
  colbacktitle=uclagold!80!black,
  left=6pt, right=6pt, top=6pt, bottom=6pt,
  breakable
}

\definecolor{codebg}{HTML}{F5F5F5}
\definecolor{codeframe}{HTML}{CCCCCC}
\definecolor{deepseek}{HTML}{4A6FA5}
\definecolor{qwen}{HTML}{7B2D8B}
\definecolor{olmo}{HTML}{2E7D32}
\definecolor{specialtoken}{HTML}{C62828}
\definecolor{commentgray}{HTML}{888888}

\lstdefinestyle{chattemplate}{
    basicstyle=\ttfamily\scriptsize,
    breaklines=true,
    breakatwhitespace=false,
    tabsize=2,
    showstringspaces=false,
    xleftmargin=0em,
    xrightmargin=0em,
    aboveskip=0em,
    belowskip=0em,
    literate={—}{{\textemdash}}1
             {→}{{$\rightarrow$}}1,
}

\definecolor{c-green-dark}{RGB}{18,120,102}
\definecolor{c-green-light}{RGB}{234,248,242}
\newtcolorbox{insightbox}[1]{
  enhanced,
  breakable,
  colback=white,
  colframe=c-green-dark,
  coltitle=black,
  colbacktitle=c-green-light,
  title={\small{\textcolor{c-green-dark}{\textbf{#1}}}},
  boxsep=2.7pt,
  top=1pt,
  bottom=1pt,
  left=4pt,
  right=4pt,
  fontupper=\small
}

\definecolor{deepMorandiBlue}{HTML}{0D47A1} 
\definecolor{deepMorandiRed}{RGB}{192, 57, 43}
\definecolor{deepMorandiTeal}{HTML}{00695C} 
\definecolor{deepMorandiGreen}{HTML}{2E7D32} 
\definecolor{algoBg}{HTML}{E3F2FD} 
\definecolor{promptBg}{HTML}{E8F5E9} 

\definecolor{promptBlue}{RGB}{159, 195, 224}
\definecolor{promptGreen}{RGB}{168, 204, 185}
\definecolor{promptPurple}{RGB}{180, 160, 200}
\definecolor{promptOrange}{RGB}{234, 190, 150}
\definecolor{promptRed}{RGB}{224, 160, 160}

\hypersetup{
    colorlinks=true,
    citecolor=deepMorandiTeal,
    linkcolor=deepMorandiRed,
    urlcolor=deepMorandiBlue
}

\title{\name{}: A Deep-Structured Evaluation Framework for Fine-Grained Audio-Visual Captioning}

\providecommand{\Letter}{\ding{41}}
\providecommand{\Heart}{\ding{170}}

\newsavebox{\envbox}
\sbox{\envbox}{\raisebox{0.9ex}{\scalebox{0.6}{\Letter}}}
\newcommand{\envmark}{\usebox{\envbox}}
\newsavebox{\heartbox}
\sbox{\heartbox}{\raisebox{0.7ex}{\scalebox{0.7}{\Heart}}}
\newcommand{\heartmark}{\usebox{\heartbox}}
\newcommand{\plmark}{\textsuperscript{\dag}}

\makeatletter
\newcommand{\blfootnote}[1]{%
  \begingroup
  \renewcommand{\thefootnote}{}%
  \renewcommand{\@makefnmark}{}%
  \footnotetext{#1}%
  \endgroup
}
\makeatother
\author{%
\textbf{Zhongyu Yang}\textsuperscript{1}\heartmark \quad
\textbf{Jiale Tao}\textsuperscript{1}\heartmark\plmark \quad
\textbf{Ruitao Chen}\textsuperscript{1}\heartmark \quad \\
\textbf{Zuhao Yang}\textsuperscript{2}\quad
\textbf{Yingfang Yuan}\textsuperscript{3} \quad 
\textbf{Xueliang Zhao}\textsuperscript{1} \quad
\textbf{Auden}\textsuperscript{1} \quad 
\textbf{Kai Wang}\textsuperscript{1}\quad
\textbf{Shuai Shao}\textsuperscript{1} \quad \\
\textbf{Biao Wang}\textsuperscript{1}\envmark \quad
\textbf{Steve Yves}\textsuperscript{1}\envmark \quad
\textbf{Qinglin Lu}\textsuperscript{1} \\
\textsuperscript{1}Hunyuan, Tencent \quad
\textsuperscript{2}Nanyang Technological University \quad
\textsuperscript{3}Northumbria University \\[3pt]
Project Website: \url{https://01yzzyu.github.io/OmniCapBench/}
}

\begin{document}
\addtocontents{toc}{\protect\setcounter{tocdepth}{0}}

\maketitle

\blfootnote{\heartmark\ Equal Contribution. 
\plmark\ Project Leader. 
\envmark\ Corresponding Author.}

\vspace{-10mm}
\begin{center}
    \includegraphics[width=\linewidth]{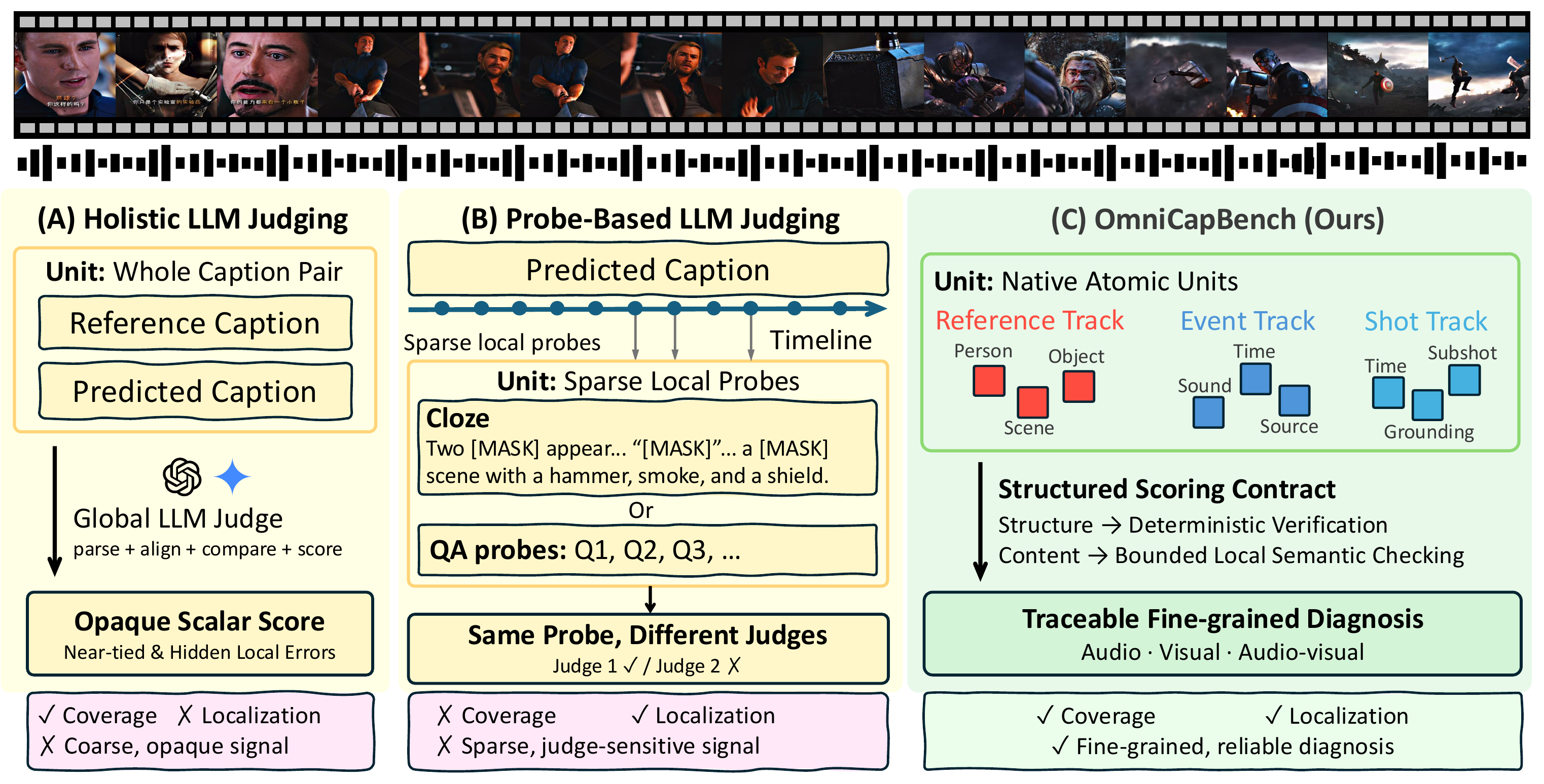}
    \captionsetup{hypcap=false}
    \captionof{figure}{\textbf{Comparison of existing evaluation frameworks and the deep-structured evaluation framework.} Both \textbf{(A) Holistic Evaluation} and \textbf{(B) Probe-Based Evaluation} derive scores from predicted unstructured, free-form text captions, but introduce distinct limitations. The former uses a global LLM-judge approach that treats the caption as a whole, resulting in an opaque score that masks local errors and lacks granularity, whereas the latter adopts a QA-probe approach to detect local errors but sacrifices global coverage and introduces instability across different LLMs. \textbf{(C) \name{} (Ours)} proposes a deep-structured evaluation framework. The caption is first natively represented as atomic, verifiable audio-visual units. Deterministic rules then verify structural and temporal relationships and dispatch these aligned units to localized, reliable LLMs to perform fine-grained semantic checks. This design preserves comprehensive coverage and achieves fine-grained error localization while eliminating the uncertainty associated with global LLM-based judges.}
    \label{fig:teaser}
\end{center}

\begin{abstract}
Multimodal large language models (MLLMs) are rapidly evolving toward continuous audio--visual reasoning, creating an urgent need for evaluations that expose their capability limits. Audio--visual captioning is an ideal diagnostic task, yet current benchmarks face a coupled trade-off: whole-caption scores provide coverage without localization, local probes provide localization without coverage, and unconstrained LLM judges introduce instability. We introduce OmniCapBench (Omni-Video Caption Benchmark), a benchmark that reframes audio--visual caption evaluation as a deep-structured diagnostic framework. OmniCapBench shifts the prediction target from free-form text to sets of atomic, verifiable evaluation units across three tracks: entity references, visual shots, and audio events, enabling reliable scoring with deterministic constraint checks and localized LLM-based semantic comparisons. With 786 densely annotated videos, OmniCapBench effectively distinguishes MLLM perception errors, including temporal grounding failures, identity drift, cross-modal misalignment, and hallucinated descriptions. Evaluating frontier MLLMs reveals strong local perception but weak long-horizon audio--visual reasoning, particularly in identity drift and cross-modal misalignment, providing a fine-grained roadmap for omnimodal development.
\end{abstract}

\section{Introduction}
\label{sec:introduction}

Multimodal large language models (MLLMs) are rapidly evolving to reason over continuous audio--visual streams~\citep{yao2024minicpmv, yang2025qwen3, xu2025qwen3omni, minicpmo45, qwen35omni}. This shift demands a corresponding evolution in evaluation. For representative tasks like audio--visual captioning~\citep{chai2025auroracap, zhang2026vcapsbench, geng2025longvale}, assessment must transcend coarse text-quality rankings. \textbf{In particular,} explicitly diagnosing specific capabilities, such as identity tracking, temporal grounding, and audio-visual association, is essential to expose fine-grained weaknesses and provide actionable insights for model improvement.



Existing caption benchmarks rarely co-design two tightly coupled components that enable fine-grained and reliable evaluation: \textbf{the evaluation unit} and \textbf{the scoring operator}. The unit determines the granularity of evaluation, whereas the scoring operator governs reliability over units.

For the \textbf{evaluation unit}, existing protocols are largely built around the predicted free-form caption, but differ in granularity. Whole-caption protocols treat the predicted–reference caption pair as a single unit (\textbf{Figure~\ref{fig:teaser}~(A)})~\citep{zhang2026vcapsbench, chai2025auroracap}, providing broad coverage but compressing fine-grained information into a global scalar that often fails to reveal local errors. Probe-based protocols instead define units as response–ground-truth pairs for question-answering (QA) probes or fact-related cloze tasks (\textbf{Figure~\ref{fig:teaser}~(B)})~\citep{xiao2021nextqa, xu2026carebench, ma2026omnicaptioner, geng2025longvale}, thereby accessing localized information within the predicted caption. This enables more localized and specific scoring, but the sparse units lack coverage. For the \textbf{scoring operator}, most existing evaluation approaches employ large language models (LLMs) to assess unstructured text captions, yet systematic efforts to obtain stable and reliable scores remain limited. Specifically, LLMs are required to reason over LLM-generated \emph{dense, monolithic text} under different evaluation protocols to compute global or local scores, introducing an inverse-engineering process that poses a substantial challenge for reliable judging. In detail, global scoring struggles to align fine-grained details within dense text, whereas local scoring is constrained by the need for accurate information extraction and matching. For instance, QA probes implicitly perform extraction and matching and are thus prone to distraction, which can lead to incorrect conclusions even when the caption is correct (Table~\ref{tab:judge_sensitivity}), while event-list scoring explicitly requires post hoc extraction and matching, which can also introduce reliability issues~\citep{geng2025longvale}. 

Consequently, the free-form, monolithic representation of model-generated captions fundamentally obstructs the joint design of the evaluation unit and the scoring operator. First, evaluation units that simultaneously provide global coverage and fine-grained local detail are not explicitly encoded. Second, the scoring operator must perform post hoc extraction prior to scoring, which introduces instability. Together, these limitations lead to additional errors and unreliable evaluation.

To close this gap, we propose \name{} (\fullname{}), a benchmark that redesigns fine-grained audio--visual captioning around deep-structured evaluation units (\textbf{Figure~\ref{fig:teaser}~(C)}). More specifically, rather than imposing a superficial format on global text, we decompose the evaluated object into native atomic units. This deep structure breaks the task into discrete, localizable targets, preserving comprehensive video coverage while eliminating the ambiguity of monolithic outputs. Crucially, it enables a reliable scoring operator: deterministic rules rigorously verify structural relationships, while LLMs are restricted to bounded semantic comparisons, avoiding the compounding uncertainty of unconstrained text-level judgments.

In practice, \name{} operationalizes this structure via three native evaluation tracks. The \emph{Reference} track catalogs persistent scenes and subjects, while the \emph{Event} track captures timestamped audio occurrences. Crucially, the \emph{Shot} track segments visual content, using explicit cross-links to temporally ground both references and events. This track design drives a two-stage scoring pipeline. First, deterministic rules and a bounded LLM matcher verify structural integrity, including valid IDs, temporal bounds, and cross-links. Second, localized LLMs exclusively assess the semantic equivalence of these isolated fields. By enforcing this strict scoring contract, \name{} elevates coarse text evaluation into a fine-grained diagnostic tool. Table~\ref{tab:benchmark_protocol_comparison} examines whether existing benchmarks natively support such verifiable atomic units.

\begin{table*}[t]
    \centering
    \vspace{-1.4em}
\caption{\textbf{Evaluation paradigm comparison.}
We compare whether existing protocols natively support the structural dimensions of omnimodal diagnosis. \emph{Identity Tracking} explicitly evaluates entity permanence across discontinuous shots (Visual). \emph{Temporal Grounding} enforces precise continuous time boundaries for actions and sounds (Audio/Visual). \emph{Audio-Visual Association} evaluates whether audio events are correctly attached to concurrent visual shots (Audio-Visual). \emph{Diagnostic Traceability} indicates whether the evaluation isolates explicit structural breakdowns from descriptive hallucinations.
\textcolor{table_green}{\cmark}: natively supported as a verifiable unit; \textcolor{table_gray}{\(\triangle\)}: implicitly judged; \textcolor{table_red}{\xmark}: not supported.}
    \label{tab:benchmark_protocol_comparison}
    \scriptsize
    \setlength{\tabcolsep}{4pt}
    \resizebox{\linewidth}{!}{
    \begin{tabular}{lcccccc}
    \toprule
    \textbf{Benchmark} &
    \textbf{Evaluated Unit} &
    \textbf{Scoring Operator} &
    \textbf{Identity Tracking} (Visual) &
    \textbf{Temporal Grounding} (Audio/Visual) &
    \textbf{Audio-Visual Association} (Audio-Visual) &
    \textbf{Diagnostic Traceability} \\
    \midrule\midrule
    \multicolumn{7}{c}{\cellcolor{gray!10}\textbf{Whole-Caption Paradigm}} \\
    AuroraCap~\citep{chai2025auroracap} &
    Caption &
    Global LLM &
    \textcolor{table_red}{\xmark} &
    \textcolor{table_red}{\xmark} &
    \textcolor{table_red}{\xmark} &
    \textcolor{table_red}{\xmark} \\
    VCapsBench~\citep{zhang2026vcapsbench} &
    Caption &
    Global LLM &
    \textcolor{table_red}{\xmark} &
    \textcolor{table_red}{\xmark} &
    \textcolor{table_red}{\xmark} &
    \textcolor{table_red}{\xmark} \\
    UGC-VideoCap~\citep{wu2025ugcvideocaptioner} &
    Caption &
    Global LLM &
    \textcolor{table_red}{\xmark} &
    \textcolor{table_red}{\xmark} &
    \textcolor{table_red}{\xmark} &
    \textcolor{table_red}{\xmark} \\
    video-SALMONN 2~\citep{tang2025videosalmonn2} &
    Caption &
    Global LLM &
    \textcolor{table_red}{\xmark} &
    \textcolor{table_red}{\xmark} &
    \textcolor{table_red}{\xmark} &
    \textcolor{table_red}{\xmark} \\
    \midrule
    \multicolumn{7}{c}{\cellcolor{gray!10}\textbf{Probe-Based Paradigm}} \\
    Omni-Cloze~\citep{ma2026omnicaptioner} &
    Cloze QA &
    LLM Scorer &
    \textcolor{table_gray}{\(\triangle\)} &
    \textcolor{table_gray}{\(\triangle\)} &
    \textcolor{table_gray}{\(\triangle\)} &
    \textcolor{table_gray}{\(\triangle\)} \\
    \midrule
    \multicolumn{7}{c}{\cellcolor{gray!10}\textbf{Parse-then-Score Paradigm}} \\
    LongVALE~\citep{geng2025longvale} &
    Event text &
    Parser + LLM &
    \textcolor{table_red}{\xmark} &
    \textcolor{table_green}{\cmark} &
    \textcolor{table_red}{\xmark} &
    \textcolor{table_red}{\xmark} \\
    TimeChat~\citep{yao2026timechatcaptioner} &
    Timed script &
    Parser + LLM &
    \textcolor{table_red}{\xmark} &
    \textcolor{table_green}{\cmark} &
    \textcolor{table_red}{\xmark} &
    \textcolor{table_gray}{\(\triangle\)} \\
    OmniScript~\citep{pu2026omniscript} &
    Script text &
    Parser + LLM &
    \textcolor{table_gray}{\(\triangle\)} &
    \textcolor{table_green}{\cmark} &
    \textcolor{table_gray}{\(\triangle\)} &
    \textcolor{table_gray}{\(\triangle\)} \\
    \midrule
    \multicolumn{7}{c}{\cellcolor{gray!10}\textbf{Native Atomic Paradigm}} \\
    \textbf{\name{} (Ours)} &
    Atomic units &
    Rule + Local LLM &
    \textcolor{table_green}{\cmark} &
    \textcolor{table_green}{\cmark} &
    \textcolor{table_green}{\cmark} &
    \textcolor{table_green}{\cmark} \\
    \bottomrule
    \end{tabular}
    }
\end{table*}

To validate \name{}, our experiments proceed in three parts. First, we establish a capability benchmark for state-of-the-art omnimodal models, decoupling evaluation into deterministic rules (Table~\ref{tab:main_results}) and bounded semantic checks (Table~\ref{tab:llm_judge_results}) to pinpoint localized bottlenecks. Second, we audit the holistic evaluation paradigm across models, metrics, and generation formats (Table~\ref{tab:caption_state_gap_result}), demonstrating that global text-level scores mask local errors and reward structurally flawed outputs. Finally, by contrasting weak and strong LLM judges (Table~\ref{tab:judge_sensitivity}), we show that global, unlocalized LLM scoring induces unstable conclusions, confirming the necessity of our decoupled framework.

Our contributions are three-fold: 
\textbf{(i)} We propose a \textbf{Deep-Structured Evaluation Paradigm} that decomposes audio--visual captioning into atomic, verifiable units, explicitly isolating semantic content, temporal grounding, identity tracking, and cross-modal association. 
\textbf{(ii)} We instantiate this paradigm into the \textbf{\name{} Benchmark}, a rigorous testbed comprising 786 densely annotated videos. Its native Reference, Shot, and Event tracks yield 5,818 entities, 6,537 audio events, and 11,419 visual shots to evaluate fine-grained omnimodal comprehension. 
\textbf{(iii)} We conduct an \textbf{Empirical Audit of Holistic Scoring}, exposing its vulnerability to judge instability and tendency to mask localized errors, thereby demonstrating the necessity of deep-structured metrics.

\section{Related Work}
\label{sec:related}

\paragraph{Omnimodal large language models.}
Omnimodal large language models are rapidly advancing toward reasoning over continuous audio--visual streams~\citep{yang2025qwen3, bai2025qwen25vl, xu2025qwen25omni, xu2025qwen3omni, qwenteam2026qwen35omni, li2024baichuanomni, liu2025ola, zhao2025humanomni, yang2025humanomniv2, sun2024videosalmonn, fu2025vita15, tang2025videosalmonn2, chen2025avocado, gemini25, gemini2024, nvidia2026nemotron3nanoomni,inex}. Accelerated by reinforcement learning and multi-agent systems~\citep{zhong2025omnir1, zhao2025r1omni, xing2025echoinkr1, chen2026omnivideor1, yang2026svagent, yang2026longvt,wu2026streamopd,zhang2026aero, Timeexpert, yang2026paravt}, audio--visual captioning has become a foundational task requiring the synthesis of visual, motion, and acoustic signals. While broad video benchmarks~\citep{fu2025videomme, hu2025videommmu, wu2024longvideobench, wang2025lvbench, zhou2025mlvu, hong2026worldsense, goel2026mmou, yang2026do, xu2026multihaystackbenchmarkingmultimodalretrieval,wu2026worldreasonbench, he2026vgi} track general progress, they often reduce complex behaviors to coarse text-level scores, lacking the diagnostic resolution needed to pinpoint localized audio--visual failures.

\paragraph{Audio--visual Captioning Evaluation.}
Moving beyond legacy metrics~\citep{vedantam2015cider, banerjee2005meteor, lin2004rouge, he2026omniduplexeval}, recent protocols explore different evaluation units. Holistic evaluation scores full captions across dimensions~\citep{zhang2026vcapsbench, chai2025auroracap, liu2025capability, chen2025vidcapbench, wu2025ugcvideocaptioner, li2026omnivideobench}, yet treating text as a monolithic unit masks specific temporal or relational errors. QA and temporal probes~\citep{xiao2021nextqa, xu2026carebench, ma2026omnicaptioner, cai2024temporalbench, li2026towards, omnicapif} localize errors via explicit questions but suffer from sparse coverage. Other methods parse free-form text into script items~\citep{geng2025longvale, yao2026timechatcaptioner, pu2026omniscript, chen2026diadem, fujita2020soda}, treating structure as a fragile post-hoc extraction layer. Crucially, when an LLM acts as a global judge—simultaneously parsing unstructured text, resolving references, and tracking time—these entangled tasks conflate factual hallucinations with stylistic variations, leading to unreliable scoring.

In short, existing methods face a strict tradeoff: they either sacrifice coverage for localization, or rely on opaque LLM judgments that obscure fine-grained errors. \name{} resolves this tension through \emph{deep-structured evaluation}. By mandating models to output native, atomic audio--visual units, our framework achieves both comprehensive coverage and exact localizability. Furthermore, we decouple the scoring process: deterministic probes verify structural relationships, while reliable localized LLMs are strictly bounded to local semantic checks. This eliminates the compounding uncertainty of global LLM judges, yielding a highly reliable diagnostic signal.

\begin{figure}[t]
    \centering
    \includegraphics[width=\linewidth]{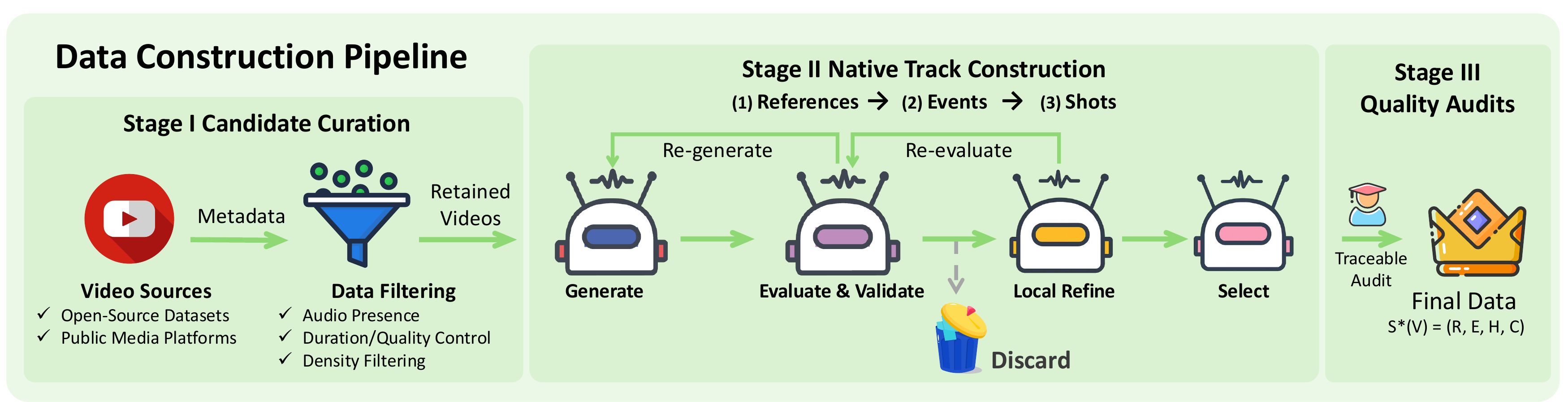}
    \caption{\textbf{\name{} data construction pipeline.} The construction proceeds in three stages. \textbf{Stage I} filters raw videos to isolate inputs with rich audio--visual complexity. \textbf{Stage II} generates atomic units (References $\rightarrow$ Events $\rightarrow$ Shots) through an iterative multi-model loop, enforcing structural integrity before semantic refinement. \textbf{Stage III} applies traceable human audits to finalize the rigorously verified audio--visual reference system.}
    \label{fig:pipeline}
\end{figure}

\section{\name{} Benchmark}
\label{sec:method}

To establish a fine-grained, reliable evaluation standard for omnimodal foundation models, we introduce the \name{} benchmark. We first present the data construction pipeline (Section~\ref{subsec:data_curation}), which converts raw videos into well-defined, verifiable atomic units. We then outline the evaluation goals and diagnostic task suite built on this structured representation (Section~\ref{subsec:eval_goals}).

\subsection{Data Construction and Native Track Generation}
\label{subsec:data_curation}

From an evaluation standpoint, rigorously verifying an AI's audio--visual understanding requires isolating \emph{what} happened from \emph{when} and \emph{where} it occurred. To achieve this fine-grained diagnostic evaluation, we inherently require a structured representation that systematically decouples distinct video elements, such as entities, auditory events, and visual states. The Multi-Stream Scene Script (MTSS)~\citep{mtss} naturally provides this factorized foundation, making it an ideal fit for our evaluation objectives. Therefore, we build \name{} upon MTSS, adopting its core decoupled design to independently track these information streams. To fully support precise benchmarking, we further optimize its coarse-grained limitations: specifically, we decompose visual shots into finer \emph{subshots} and replace discrete point timestamps with continuous \emph{time ranges}.

For a given video \(V\), we formalize its content into a structured representation \(\mathcal{S}^\star(V)=(\mathcal{R},\mathcal{E},\mathcal{H})\). To guarantee structural integrity, this representation is built across three interdependent tracks:
\begin{itemize}[leftmargin=1.5em, itemsep=0pt, parsep=0pt, topsep=2pt]
    \item \textbf{References (\(\mathcal{R}\)):} This track establishes persistent identities for key entities, including people, objects, and scenes, by assigning unique identifiers that enable consistent tracking of each entity throughout the video.
    \item \textbf{Events (\(\mathcal{E}\)):} This track isolates auditory events and spoken dialogue together with their precise temporal boundaries, for example, extracting a continuous dog bark from 10.5s to 12.0s.
    \item \textbf{Shots (\(\mathcal{H}\)):} This track segments the visual timeline into discrete camera shots and serves as the unifying structure by anchoring previously defined entities and audio events to specific frames, for example, linking the barking sound to the shot that shows the dog.
\end{itemize}

This atomic schema separates semantic content from the complex structure of audio–visual streams through explicit localization. In this way, the evaluation can precisely verify whether a predicted event occurs at the correct time and is linked to the appropriate visual entities, while enabling a localized LLM scorer to focus only on structurally aligned content for semantic verification.

The construction of this reference system proceeds in three stages (Figure~\ref{fig:pipeline}).

\begin{figure}[t]
\vspace{-4mm}
    \centering
    \begin{minipage}[t]{0.39\linewidth}
        \vspace{0pt}
        \centering
\scriptsize
\setlength{\tabcolsep}{2pt}
\renewcommand{\arraystretch}{1.05}
\begin{tabular}{@{}p{0.52\linewidth}>{\raggedleft\arraybackslash}p{0.22\linewidth}>{\raggedleft\arraybackslash}p{0.20\linewidth}@{}}
\toprule
\textbf{Statistic} & \textbf{Avg.\,/\,Pct.} & \textbf{Total} \\
\midrule
Videos ($<1$ min) & 61.32\% & 482 \\
Videos (1--3 min) & 29.01\% & 228 \\
Videos (3--5 min) & 9.67\% & 76 \\
Duration (hours) & -- & 12.8 \\
Categories (Level-1) & -- & 20 \\
Categories (Level-2) & -- & 125 \\
\midrule
References & 7.40 & 5,818 \\
Shots & 14.53 & 11,419 \\
Subshots & 49.82 & 39,160 \\
Events & 8.32 & 6,537 \\
\quad \textit{-- Dialogue} & 6.83 & 5,370 \\
\bottomrule
\end{tabular}

    \end{minipage}
    \hfill
    \begin{minipage}[t]{0.59\linewidth}
        \vspace{0pt}
        \centering
        \includegraphics[width=\linewidth]{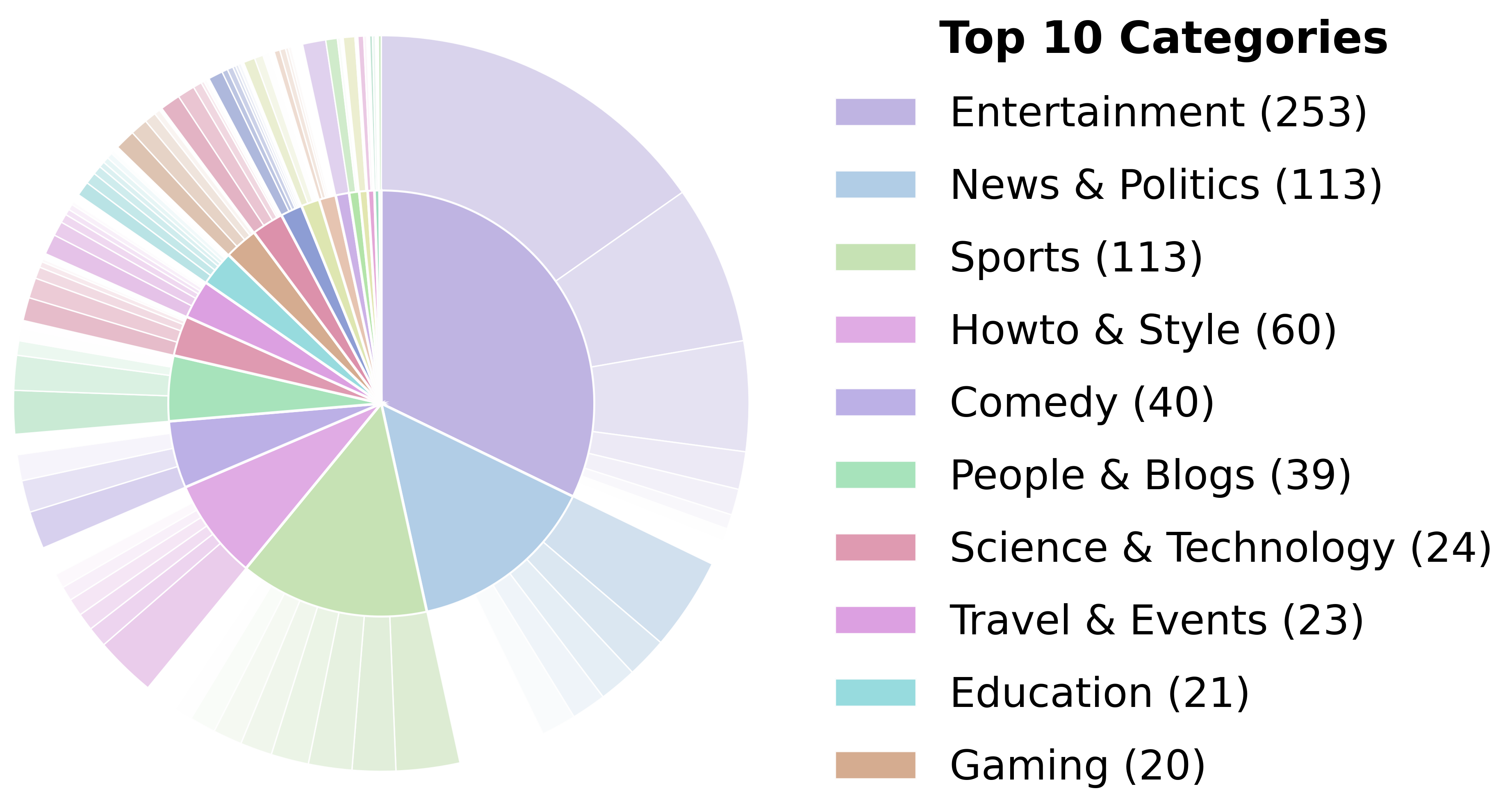}
    \end{minipage}
    
    \vspace{0.5em}
    
    \begin{minipage}[t]{0.39\linewidth}
        \captionsetup{hypcap=false}
        \captionof{table}{\textbf{Key statistics of \name{}}, comprising 786 videos. ``Avg.\,/\,Pct.'' denotes either the average or the percentage of total videos.}
        \label{tab:keystats}
    \end{minipage}
    \hfill
    \begin{minipage}[t]{0.57\linewidth}
        \captionsetup{hypcap=false}
        \captionof{figure}{\textbf{\name{} category distribution.} Inner and outer rings represent level-1 categories and level-2 subcategories, respectively. Segment area is proportional to video count, illustrating the benchmark coverage.}
        \label{fig:category_distribution}
    \end{minipage}
\vspace{-3mm}
\end{figure}
 
\noindent \textbf{Stage I: Candidate Curation.}
We source candidate videos from diverse open datasets and public platforms. Following initial preprocessing (e.g., metadata extraction, shot detection, and categorization), we apply rigorous filtering based on audio presence, duration, quality, and bucket-specific density (Appendix~\ref{app:benchmark_construction}). This curation ensures all retained videos exhibit sufficient audio--visual complexity to warrant our multi-unit structural evaluation, directly supporting our goal of probing deep omnimodal reasoning rather than isolated single-event recognition.

\noindent \textbf{Stage II: Native Track Construction.}
As defined previously, this stage iteratively builds the atomic units in the required dependency order (\(\mathcal{R} \rightarrow \mathcal{E} \rightarrow \mathcal{H}\)).

To autonomously guarantee structural integrity during data generation, we execute an iterative \emph{Local Refinement} loop: \textbf{Generate $\rightarrow$ Evaluate \& Validate $\rightarrow$ Local Refine $\rightarrow$ Select}. First, a generator proposes initial candidate tracks for the video. Next, an evaluator ensemble checks the factual accuracy of these candidates, while a programmatic validator rigorously enforces schema constraints, triggering automatic regeneration or discarding of invalid outputs. Crucially, once the foundational topology is validated, a local refinement step enriches the textual descriptions. During this process, the underlying structure, such as entity identifiers, temporal boundaries, and cross-links, remains strictly invariant. This constrained refinement explicitly prevents linguistic enhancements from introducing new hallucinations or corrupting the already verified topology. Finally, a selector model chooses the highest-quality refined version to form the final canonical tracks. Note that the frames used in this stage carry a red timestamp overlay purely as an annotation aid; evaluated models always receive the original frames without it (Appendix~\ref{app:benchmark_construction}).

\noindent \textbf{Stage III: Quality Audits.}
Because the native reference system \(\mathcal{S}^\star(V)\) directly dictates the evaluation scoring contract, its data integrity must be audited before scoring claims are reported. In this final stage, programmatic validators inspect schema validity, timestamp ranges, and cross-links, while targeted human audits inspect boundary and semantic disagreements. The finalized units provide each video with a reliable reference system \(\mathcal{S}^\star(V) = (\mathcal{R}, \mathcal{E}, \mathcal{H})\) for model evaluation.

\noindent \textbf{Dataset Statistics.} Table~\ref{tab:keystats} reports the verified dataset size and per-video averages, while Figure~\ref{fig:category_distribution} summarizes the category coverage.

\subsection{Evaluation Goals and Task Suite}
\label{subsec:eval_goals}
\label{subsec:state_scoring_principles}

The overarching goal of the \name{} benchmark is to evaluate omnimodal foundation models with \emph{high granularity} and \emph{reliability}. Instead of using an unconstrained text-centric generation process that inherently conflates a model's true audio--visual comprehension with its language style, \name{} derives its diagnostic tasks directly from the rigorous native reference system $\mathcal{S}^\star(V)$. Appendix~\ref{app:benchmark_construction} details the exact input--output formats.

\textbf{Task 1: Native Unit Generation.} In this primary task, a model takes a video $V$ and directly outputs the three prediction tracks (References, Shots, and Events) as a structured JSON object. Instead of writing a free-form paragraph, the model must explicitly construct the set of atomic units \(\hat{\mathcal{S}}(V)\). This forces the model to expose its internal reasoning in a highly structured format, directly aligning with our goal of fine-grained evaluation and removing the ambiguity of free-form text. We use this task to assess state-of-the-art models and quantify their structural and semantic accuracy.

\begin{table*}[t]
\vspace{-3mm}
\caption{\textbf{Rule-based structural evaluation.} Metrics quantify structural compliance across Visual, Audio, and Audio-Visual dimensions. \textbf{SGC} is an independent schema prerequisite. Scores are macro-averaged across videos. \textbf{Best} and \underline{second-best} results are highlighted.}
\label{tab:main_results}
\scriptsize
\setlength{\tabcolsep}{2pt}
\resizebox{\linewidth}{!}{
\begin{tabular}{lc cccccccc cc cc}
\toprule
\multirow{2}{*}{\textbf{Model}} &
\multirow{2}{*}{\textbf{SGC}} &
\multicolumn{8}{c}{\textbf{Visual}} &
\multicolumn{2}{c}{\textbf{Audio}} &
\multicolumn{2}{c}{\textbf{Audio-Visual}} \\
\cmidrule(lr){3-10} \cmidrule(lr){11-12} \cmidrule(lr){13-14}
& &
\textbf{RefUse} & \textbf{CCC} & \textbf{Ref Subject F1} & \textbf{Ref Scene F1} & \textbf{Shot F1} & \textbf{Shot tIoU} & \textbf{Subshot F1} & \textbf{Subshot tIoU} &
\textbf{Event F1} & \textbf{Event tIoU} &
\textbf{Speaker F1} & \textbf{EVSA F1} \\
\midrule
\rowcolor{limegreen!50} \multicolumn{14}{c}{\textbf{\textit{Proprietary Models}}} \\
\midrule
Gemini 3.1-Pro~\citep{gemini2024} & \textbf{97.36} & \underline{70.39} & \underline{34.43} & \underline{78.60} & \underline{82.68} & \textbf{76.24} & 70.74 & 65.27 & 48.76 & \textbf{63.32} & \textbf{71.92} & 91.18 & \textbf{51.46} \\
Gemini 2.5-Pro~\citep{gemini25} & \underline{96.73} & \textbf{70.75} & \textbf{37.81} & \textbf{80.70} & \textbf{83.65} & 70.93 & \textbf{72.58} & \underline{66.69} & \underline{49.18} & \underline{58.52} & \underline{69.35} & 89.72 & \underline{43.74} \\
Qwen3.5-Omni-Plus~\citep{qwenteam2026qwen35omni} & 96.09 & 66.92 & 29.55 & 73.57 & 79.97 & 71.25 & 72.35 & 65.69 & 48.29 & 53.92 & 67.77 & \underline{91.20} & 40.20 \\
Qwen3.5-Omni-Flash~\citep{qwenteam2026qwen35omni} & 95.58 & 63.10 & 24.31 & 72.27 & 66.07 & 65.65 & 68.99 & 58.94 & 46.52 & 50.68 & 69.34 & 89.91 & 35.58 \\
Seed2.0~\citep{bytedance2026seed2} & 87.68 & 64.69 & 31.88 & 65.29 & 75.13 & \underline{72.62} & \underline{72.56} & \textbf{68.82} & 47.94 & 4.54 & 62.78 & \textbf{91.80} & 3.52 \\
MiMo-2.5~\citep{xiaomi2026mimo25} & 96.58 & 61.99 & 23.91 & 61.19 & 69.27 & 66.58 & 65.53 & 60.37 & 46.51 & 43.82 & 61.82 & 86.62 & 31.69 \\
\midrule
\rowcolor{grayred!40} \multicolumn{14}{c}{\textbf{\textit{Open-source Omnimodal Models}}} \\
\midrule
Qwen3-Omni-Instruct~\citep{xu2025qwen3omni} & 89.02 & 40.96 & 9.85 & 54.84 & 64.55 & 49.98 & 67.92 & 29.51 & 47.95 & 41.70 & 40.99 & 88.35 & 13.33 \\
Qwen3-Omni-Captioner~\citep{xu2025qwen3omni} & 90.77 & 44.96 & 11.51 & 54.36 & 57.08 & 52.29 & 70.80 & 39.51 & \textbf{49.69} & 41.31 & 52.60 & 88.68 & 17.42 \\
MiniCPM-o-2.6~\citep{yao2024minicpmv} & 72.13 & 21.99 & 4.29 & 24.36 & 3.38 & 17.28 & 58.45 & 9.90 & 42.99 & 6.41 & 27.44 & 80.07 & 2.46 \\
\bottomrule
\end{tabular}
}
\vspace{-3mm}
\end{table*}

\textbf{Task 2: Structural Verification and Matching.}
Before semantic scoring, we first check whether each predicted unit is structurally valid. A valid unit must use defined identifiers, have a well-formed time span, and link only to existing and temporally compatible references, shots, or events. Predictions that violate these rules, such as undefined IDs, reversed start--end times, or impossible event--shot links, are removed by deterministic checks. We then match the remaining units to the ground truth. Shots and events are matched by temporal intersection-over-union (tIoU), while identity references and visual subshots are matched with a bounded LLM-assisted matcher that only compares local candidate pairs. This design separates structure from semantics: invalid or unmatched units are counted as structural errors, and the LLM scorer is used only on aligned unit pairs. Thus, errors such as invalid IDs, weak temporal overlap, or wrong event--shot links become traceable structural deficits rather than hidden components of a single caption-level score.

\textbf{Task 3: Localized Semantic Scoring.} In the final stage, we dispatch the matched unit pairs to an LLM judge. With structural complexities (e.g., timing and identity tracking) already resolved in Task 2, the LLM is strictly bounded to perform local semantic equivalence comparisons on isolated content fields. This ``structure-first, semantics-second'' approach isolates genuine descriptive fidelity from structural noise. Crucially, by decomposing the evaluation into deep-structured atomic units, the assessment becomes significantly simpler. This substantially reduces the dependency on advanced LLM reasoning capabilities, ensuring highly objective and reliable results.

\textbf{Bidirectional Evaluation: Precision vs.\ Recall.} Because open-ended video generation lacks a perfect one-to-one mapping between predictions and human annotations (for instance, a model might decompose a visual action into three dense subshots while the ground truth summarizes it in one), evaluating from a single direction is inherently biased. For open-vocabulary and continuous fields like References (Subject and Scene) and Subshots, \name{} enforces a \textbf{bidirectional matching and scoring} paradigm. Specifically, we execute the matching and scoring processes from two distinct perspectives with different focuses: 
(1) \textbf{The Recall Perspective} (Ground Truth $\rightarrow$ Prediction) uses the ground truth as the subject to check if the prediction covers it, strictly penalizing \emph{omissions}. 
(2) \textbf{The Precision Perspective} (Prediction $\rightarrow$ Ground Truth) uses the prediction as the subject to check if the ground truth supports it, strictly penalizing \emph{hallucinations}. 
The final \textbf{F1 scores} for categories like Subshot F1 are directly derived by computing the harmonic mean of these two distinct directional metrics. In our reporting, \textbf{Ref Subject} encompasses both persons and objects, while \textbf{Ref Scene} corresponds to scene references. This prevents models from gaming the evaluation through overly terse responses (which would fail recall) or excessively verbose, speculative dumps (which would fail precision). Appendix~\ref{app:bidirectional_matching} provides extensive details on this mechanism. 

\textbf{Structured Diagnostic Metrics.} \name{} provides a suite of deterministic metrics designed to expose specific structural vulnerabilities, moving beyond holistic text similarity. Exact formulations are detailed in Appendix~\ref{app:metrics_definitions}. 
In the \textbf{Visual} domain, \emph{Reference Subject/Scene F1} evaluate basic entity detection, while \emph{Reference Use (RefUse)} and \emph{Cross-Shot Coreference Consistency (CCC)} check identity permanence across video cuts. For temporal grounding, \emph{Shot/Subshot F1} and \emph{tIoU} measure the recovery and boundary alignment of visual states. 
In the \textbf{Audio} domain, \emph{Event F1} and \emph{tIoU} verify the recovery and precise temporal alignment of sound events. 
In the \textbf{Audio-Visual} domain, \emph{Event-Shot Association F1 (EVSA)} measures whether audio events are correctly attached to concurrent visual shots, while \emph{Speaker F1} verifies that dialogue is assigned to the correct visual speaker. 
Finally, \textbf{Localized Semantic} metrics (bounded \emph{Precision} and \emph{Recall}) quantify descriptive fidelity strictly on structurally valid units. This isolates factual comprehension from structural noise, preventing models from masking hallucinations or omissions behind fluent prose.

\begin{table*}[!t]
\vspace{-3mm}
\caption{\textbf{Localized semantic evaluation.} Semantic fidelity is assessed exclusively on structurally aligned units. Bidirectional scoring isolates distinct failure modes: Recall penalizes factual omissions, while Precision penalizes hallucinations. \textbf{Best} and \underline{second-best} results are highlighted.}
\label{tab:llm_judge_results}
\scriptsize
\setlength{\tabcolsep}{3pt}
\resizebox{\linewidth}{!}{
\begin{tabular}{l ccccccc cc}
\toprule
\multirow{3}{*}{\textbf{Model}} & \multicolumn{7}{c}{\textbf{Visual}} & \multicolumn{2}{c}{\textbf{Audio}} \\
\cmidrule(lr){2-8} \cmidrule(lr){9-10}
& \multirow{2}{*}{\textbf{Shot}} & \multicolumn{2}{c}{\textbf{Subject}} & \multicolumn{2}{c}{\textbf{Scene}} & \multicolumn{2}{c}{\textbf{Subshot}} & \multirow{2}{*}{\textbf{Dialogue}} & \multirow{2}{*}{\textbf{Non-dialogue}} \\
\cmidrule(lr){3-4} \cmidrule(lr){5-6} \cmidrule(lr){7-8}
& & \textbf{Recall} & \textbf{Precision} & \textbf{Recall} & \textbf{Precision} & \textbf{Recall} & \textbf{Precision} & & \\
\midrule
\rowcolor{limegreen!50} \multicolumn{10}{c}{\textbf{\textit{Proprietary Models}}} \\
\midrule
Gemini 3.1-Pro~\citep{gemini2024} & \textbf{59.05} & 43.78 & \textbf{67.07} & \underline{51.78} & \underline{75.62} & \textbf{53.29} & 70.92 & 91.18 & \textbf{41.49} \\
Gemini 2.5-Pro~\citep{gemini25} & \underline{55.58} & \textbf{51.53} & 57.08 & \textbf{56.80} & 65.12 & \underline{53.05} & \underline{71.21} & 89.72 & \underline{37.38} \\
Qwen3.5-Omni-Plus~\citep{qwenteam2026qwen35omni} & 54.17 & 45.41 & 62.84 & 50.75 & 69.80 & 50.22 & 70.66 & \underline{91.20} & 36.82 \\
Qwen3.5-Omni-Flash~\citep{qwenteam2026qwen35omni} & 47.66 & 43.87 & \underline{65.59} & 44.41 & 71.26 & 44.08 & 67.27 & 89.91 & 28.12 \\
Seed2.0~\citep{bytedance2026seed2} & 54.77 & 42.05 & 63.65 & 50.59 & 71.42 & 51.15 & \textbf{72.88} & \textbf{91.80} & 30.20 \\
MiMo-2.5~\citep{xiaomi2026mimo25} & 48.32 & \underline{46.70} & 58.63 & 46.72 & 65.47 & 44.93 & 67.27 & 86.62 & 30.85 \\
\midrule
\rowcolor{grayred!40} \multicolumn{10}{c}{\textbf{\textit{Open-source Omnimodal Models}}} \\
\midrule
Qwen3-Omni-Instruct~\citep{xu2025qwen3omni} & 34.63 & 40.31 & 62.02 & 36.07 & 70.73 & 29.42 & 57.84 & 88.35 & 20.51 \\
Qwen3-Omni-Captioner~\citep{xu2025qwen3omni} & 36.08 & 42.00 & 50.96 & 38.07 & 54.53 & 34.01 & 46.05 & 88.68 & 19.65 \\
MiniCPM-o-2.6~\citep{yao2024minicpmv} & 28.53 & 25.92 & 63.87 & 20.39 & \textbf{76.32} & 27.62 & 55.88 & 80.07 & 4.84 \\
\bottomrule
\end{tabular}
}
\vspace{-3mm}
\end{table*}

\section{Experiments}
\label{sec:experiments}

\subsection{Experimental Setup and Metrics}
\label{subsec:exp_setup}
\label{subsec:exp_metrics}

\textbf{Model.} 
We evaluate representative Omni MLLMs for OmniCapBench across three categories:
\begin{itemize}
\item Proprietary models: Gemini 3.1/2.5-Pro~\citep{gemini2024, gemini25}, Qwen3.5-Omni-Plus/Flash~\citep{qwenteam2026qwen35omni}, Seed2.0~\citep{bytedance2026seed2}, MiMo-2.5~\citep{xiaomi2026mimo25}
\item Open-source models: Qwen3-Omni-Instruct/Captioner~\citep{xu2025qwen3omni}, and MiniCPM-o-2.6~\citep{yao2024minicpmv}). 
\item Specialized video-captioning models:  ASID-Caption~\citep{li2026towards} and AVoCaDO~\citep{chen2025avocado}.
\end{itemize}

However, we exclude specialized video-captioning models from the main evaluation, as their supervised fine-tuning toward free-form text leads to poor instruction following on structured generation tasks; see Appendix~\ref{app:specialized_models_failure}.

\textbf{Metrics and Results Format.} Following Section~\ref{subsec:eval_goals}, our deep-structured evaluation reports fine-grained metrics across Visual (RefUse, CCC, Ref Subject/Scene F1, Shot/Subshot F1 and tIoU), Audio (Event F1 and tIoU), and Audio-Visual domains (Speaker F1 and EVSA F1). Table~\ref{tab:main_results} presents the rule-based match scores, while Table~\ref{tab:llm_judge_results} separately reports semantic equivalence scores computed only on structurally valid and aligned units.

\subsection{Benchmarking Omnimodal Model Capabilities}
\label{subsec:main_results}

Table~\ref{tab:main_results} reports the performance of evaluated models across the rule-based metric families. By shifting from global scalar text scores to deep-deep-structured evaluation units, we can pinpoint where models fail in fine-grained audio--visual comprehension. Table~\ref{tab:llm_judge_results} reports the corresponding bounded LLM-as-judge semantic scores on aligned units.

\paragraph{Diagnosing capabilities through deep structure.}
Our deep-structured evaluation exposes critical capability deficits that holistic text scores typically obscure. By decoupling evaluation into deterministic structural rules (Table~\ref{tab:main_results}) and bounded semantic checks (Table~\ref{tab:llm_judge_results}), we pinpoint three fundamental bottlenecks in current omnimodal models.

First, in the \textbf{visual} domain, \name{} shifts the focus from isolated entity recognition to continuous identity tracking. While frontier models like Gemini 2.5-Pro exhibit strong basic perception (84.80\% Ref Subject F1 on $<1$ min videos, Appendix Table~\ref{tab:per_duration_results}), their ability to track these identities across cuts degrades sharply. This is captured by the Cross-Shot Coreference Consistency (CCC) score, which drops to 37.81\% for Gemini 2.5-Pro and collapses below 12\% for open-source models. The insight is clear: current architectures struggle with long-term visual object permanence, a deficit hidden by traditional metrics that only check if an object was mentioned once.

Second, in the \textbf{audio and cross-modal} domains, our structural metrics reveal a persistent modality disconnect. While models excel at speech transcription (Dialogue semantics $\ge 80\%$), their auditory reasoning for environmental sounds remains weak. More critically, genuine omnimodal comprehension requires structurally binding these sounds to concurrent visual sources. However, Gemini 3.1-Pro's performance drops from 63.32\% (Event F1) to 51.46\% on Event-Shot Association (EVSA F1), with open-source models failing to surpass 20\%. Seed2.0 fails outright here, emitting almost no audio-event units (4.54\% Event F1, 3.52\% EVSA F1) despite top-ranked speaker attribution (91.80\%). This suggests that models still largely process audio and visual streams as unaligned pathways rather than a unified structural graph.

Finally, by restricting \textbf{semantic} evaluation exclusively to structurally valid units, we isolate true descriptive fidelity from structural noise. This localized audit uncovers a stark precision-recall trade-off that holistic scoring masks behind fluent prose. For instance, Gemini 3.1-Pro adopts a conservative strategy (67.07\% precision, 43.78\% recall on subjects), omitting details to avoid errors, whereas other models inflate recall by hallucinating unverified attributes. MiniCPM-o-2.6 is the extreme case: it tops Scene precision (76.32\%) while recalling only 20.39\% of scene content. By enforcing strict structural prerequisites, \name{} prevents models from gaming the evaluation, providing a clear roadmap for grounded omnimodal generation.

\begin{figure*}[t]
\vspace{-5mm}
    \centering
    \begin{subfigure}[t]{0.40\textwidth}
        \centering
        \includegraphics[width=\linewidth]{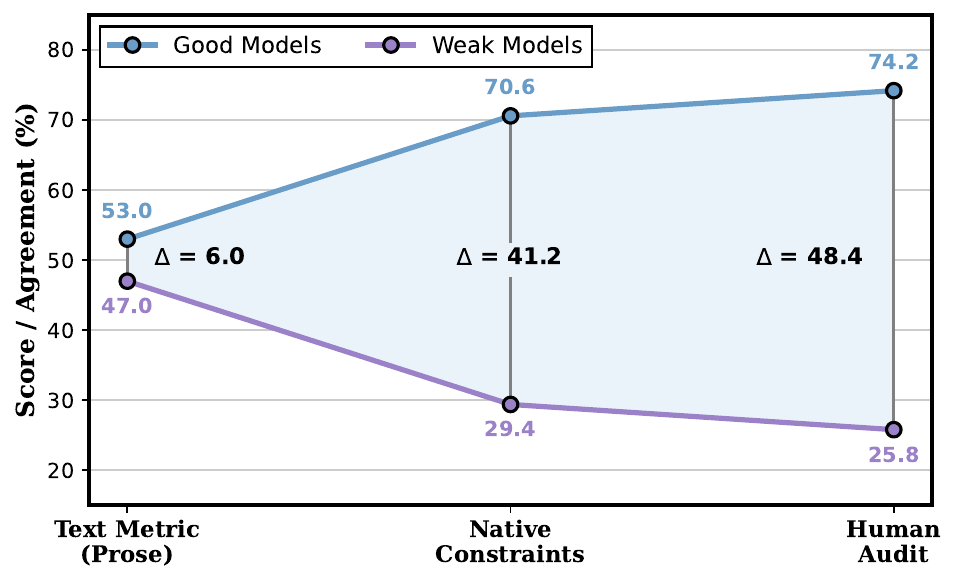}
        \caption{\textbf{The illusion of global scores.} Text metrics reward fluent but flawed outputs; structured metrics expose them.}
        \label{fig:caption_state_gap_scatter}
    \end{subfigure}
    \hfill
    \begin{subfigure}[t]{0.58\textwidth}
        \centering
        \includegraphics[width=\linewidth]{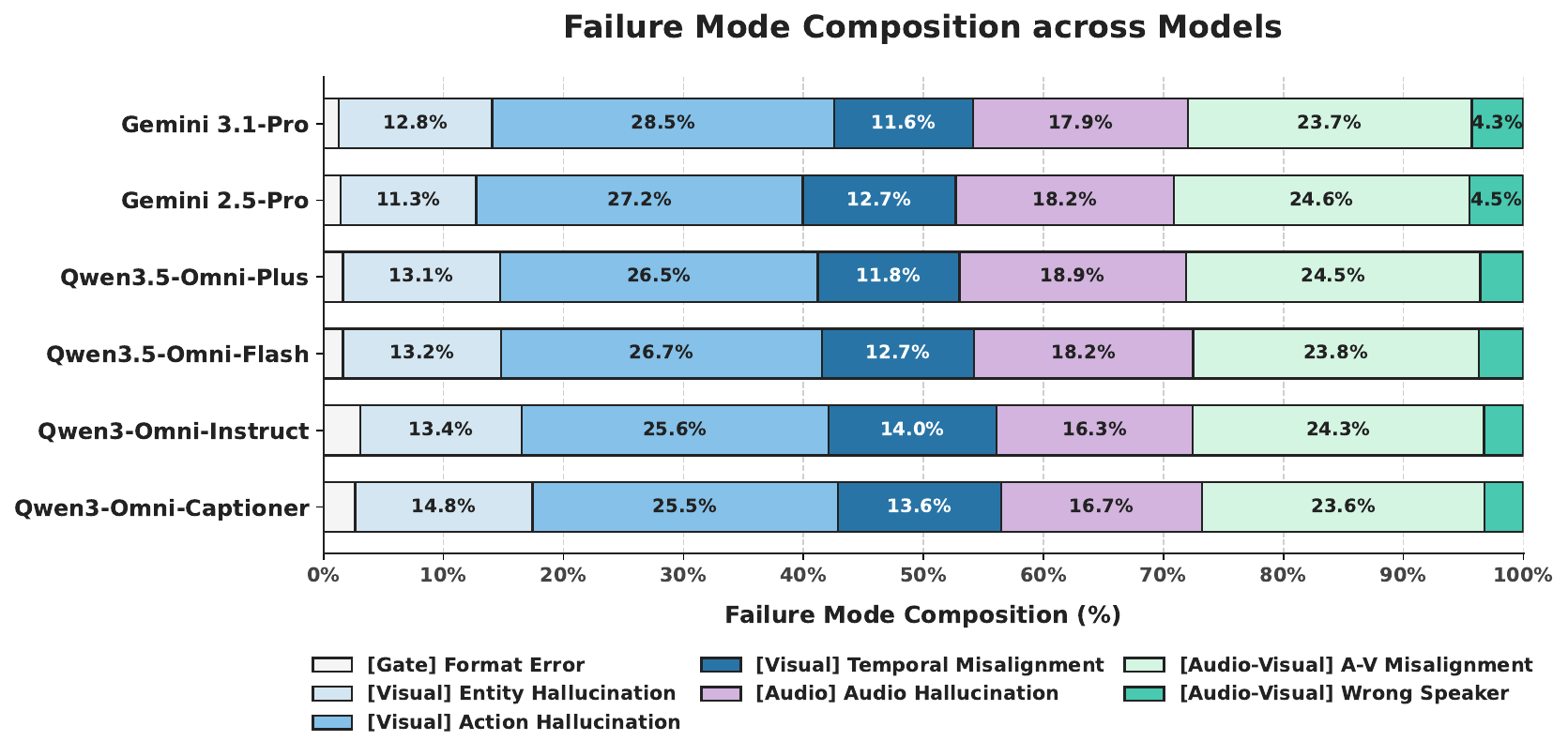}
        \caption{\textbf{Failure composition.} Distribution of structural errors, normalized by each model's total failures.}
        \label{fig:error_analysis_summary}
    \end{subfigure}
    
    \caption{\textbf{Evaluation paradigms and structural failures.}
    \textbf{(a)} Global text metrics make distinct models appear similar, whereas structured constraints better match human audit.
    \textbf{(b)} Error composition shows different bottlenecks: frontier models concentrate failures in fine-grained hallucination and audio-visual alignment, while open-source models exhibit broader degradation.}
    \label{fig:caption_gap_diagnostics}
    \vspace{-3mm}
\end{figure*}

\noindent \textbf{Error analysis.} Figure~\ref{fig:error_analysis_summary} decomposes model failures into strict structural violations. We derive this composition by converting the performance on each rule-based structural metric into an absolute capability deficit ($1 - \text{score}$) and normalizing these deficits to represent 100\% of each model's failure distribution. This detailed breakdown reveals insights invisible to holistic text metrics. First, \textbf{format errors} are negligible (1.3\% for Gemini 3.1-Pro), indicating that schema adherence is a solved prerequisite. Second, we observe distinct failure signatures across model capabilities. For frontier models like Gemini 3.1-Pro, errors are highly concentrated in deep cross-modal links and fine-grained action details: \textbf{action hallucination} accounts for a massive 28.5\% of its failures, followed by \textbf{A-V misalignment} (23.7\%) and \textbf{audio hallucination} (17.9\%). In contrast, open-source models suffer a systemic degradation, with errors distributed uniformly across both basic perception and complex audio-visual binding. This validates our core motivation: the true bottleneck in omnimodal video understanding lies not merely in describing isolated entities, but in avoiding deep-structured hallucinations like fabricating action details or assigning sounds to the wrong visual shot.

\subsection{Auditing the Legacy Evaluation Paradigm}
\label{subsec:caption_state_gap_analysis}

A fundamental motivation of \name{} is that existing holistic evaluations conflate linguistic fluency with genuine multimodal comprehension. This section tests whether global text-centric evaluations provide a reliable diagnostic signal by auditing two core vulnerabilities: the tendency of monolithic scores to obscure localized structural errors, and the severe instability introduced by unconstrained LLM judges.

\paragraph{Do global LLM judges introduce unstable conclusions?}
\begin{wraptable}{r}{0.5\textwidth}
\vspace{-1.2em}
\centering
\small
\caption{\textbf{Judge sensitivity.} Stability across judges.}
\label{tab:judge_sensitivity}
\vspace{-0.5em}
\resizebox{\linewidth}{!}{
\begin{tabular}{lccc}
\toprule
\textbf{Judge} & \textbf{Holistic} & \textbf{QA} & \textbf{\name{}} \\
\midrule
Weak & 71.24 & 61.17 & 56.82 \\
Mid & 83.08 & 53.49 & 58.24 \\
Strong & 93.92 & 56.57 & 57.60 \\
\bottomrule
\end{tabular}}
\vspace{-1.2em}
\end{wraptable}
Beyond masking errors, holistic evaluations introduce severe unreliability by forcing LLM judges to perform unconstrained global reasoning. To demonstrate this, we score identical model predictions using weak (Qwen3.6-27B), mid (GPT-4o), and strong (Gemini-2.5-Pro) judges across two baseline protocols: Holistic-style judging (UGC-VideoCap) and QA-style judging (Omni-Cloze). As Table~\ref{tab:judge_sensitivity} reports, both baselines exhibit extreme judge sensitivity—scores for the exact same outputs artificially inflate from 71.24 (weak judge) to 93.92 (strong judge). In contrast, \name{} isolates the LLM's role, restricting it exclusively to bounded semantic checks on structurally aligned units while leaving complex verification to deterministic rules. By eliminating implicit fact extraction and unconstrained reasoning, \name{} establishes a stable, objective scoring framework resilient to judge idiosyncrasies.

\begin{figure*}[t]
    \centering
    \includegraphics[width=\linewidth]{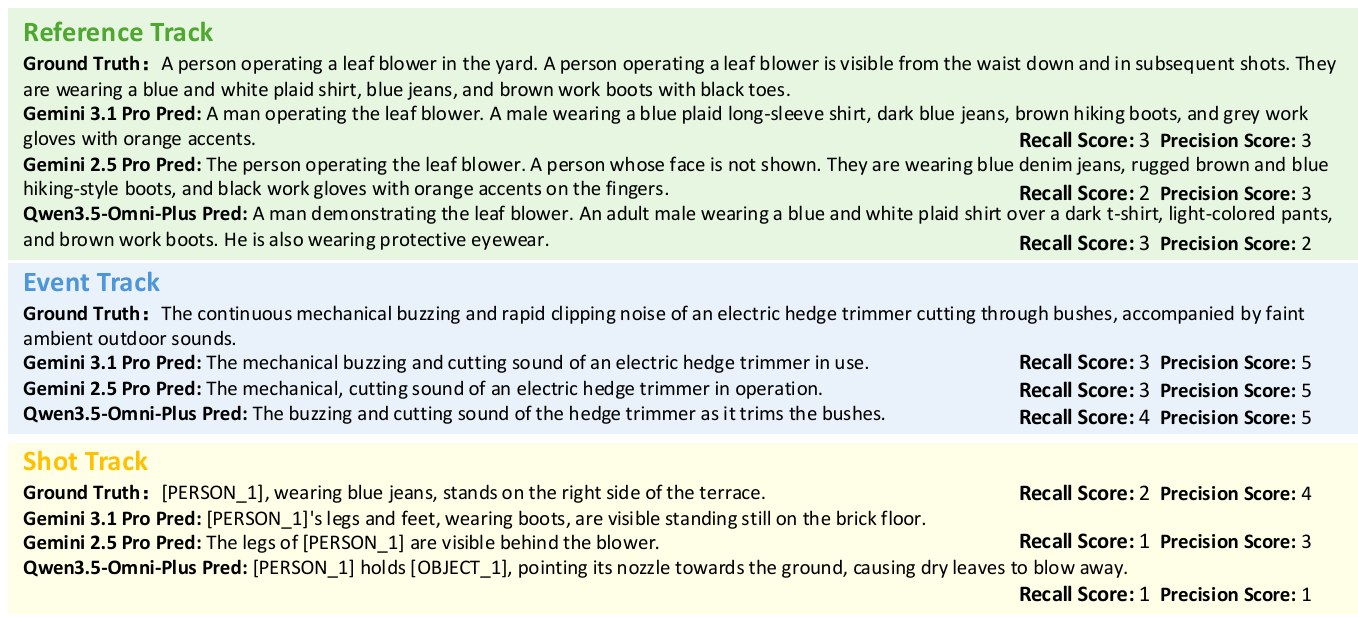}
    \caption{\textbf{Qualitative case study of deep-structured diagnostics.} Traditional holistic evaluations often over-reward fluent prose, missing critical errors like hallucinated ``protective eyewear'' or temporally unsupported actions. By performing bounded semantic checks on structurally valid units and assigning localized Recall and Precision, \name{} decouples genuine descriptive fidelity from structural noise, providing a high-resolution diagnostic trace.}
    \label{fig:case_study}
\end{figure*}

\paragraph{Does global text-level scoring hide localized errors?}
To empirically investigate whether compressing omnimodal events into a global scalar masks capability deficits, we construct hard-to-distinguish subsets of 200 videos from the UGC-VideoCap and video-SALMONN 2 benchmarks. Specifically, we define two evaluation pairs representing model upgrades: Qwen3.5-Omni-Flash vs.\ Qwen3-Omni, and Qwen3.5-Omni-Plus vs.\ Qwen3.5-Omni-Flash (details in Appendix~\ref{app:caption_state_gap_details}). For each pair, we strictly sample videos where the weaker and stronger models exhibit near-identical performance ($\sim$50\% vs. $\sim$50\%) under traditional text-centric metrics, creating a false illusion of parity. However, as Table~\ref{tab:caption_state_gap_result} and Figure~\ref{fig:caption_state_gap_scatter} demonstrate, shifting the evaluation to native structural constraints completely breaks this illusion. The average score of the stronger models climbs to perfectly align with human audit (74.2\%), whereas the weaker models collapse (25.8\%) when their structural flaws are exposed. This confirms our core hypothesis: global metrics are easily manipulated by linguistic fluency, inadvertently rewarding structurally flawed outputs as long as the generated prose is coherent. Relying on a monolithic text presentation layer severely degrades diagnostic resolution, necessitating a shift towards deep-structured evaluation.

\paragraph{Qualitative validation of deep-structured diagnostics.}
We qualitatively validate this decoupled scoring approach in Figure~\ref{fig:case_study}. Traditional paradigms are easily confounded by fluent hallucinations; for instance, unconstrained metrics frequently reward Qwen3.5-Omni-Plus for hallucinating detailed but unsupported attributes (e.g., ``protective eyewear'') because the overall prose reads well. Furthermore, models often generate plausible action descriptions that completely violate the continuous temporal boundaries of the visual shot. By strictly decomposing the evaluation into atomic units and enforcing structural alignment before any semantic checking, \name{} explicitly penalizes these specific omissions and temporal violations through localized Recall and Precision scores. This traceable diagnostic process guarantees that the final score reflects precise spatial-temporal grounding rather than mere text-generation capability.

\section{Conclusion}
\label{sec:conclusion}

The reliable evaluation of audio–visual captioning must provide fine-grained, localizable diagnostic signals. To address this need, we propose \name{}, which reframes the evaluation process through deep-structured evaluation units. The resulting atomic units across Reference, Shot, and Event tracks ensure that the evaluation achieves comprehensive video coverage while preserving fine-grained detail. Moreover, this atomic formulation enables a decoupled scoring process, in which deterministic rules rigorously verify structural relationships and LLMs are strictly restricted to local semantic checks. Beyond establishing a rigorous benchmark for current multimodal models, our empirical analysis reveals critical flaws in holistic paradigms. Global text scores often mask localized errors and reward structurally inconsistent outputs, while unconstrained LLM judges introduce severe ranking instability. Ultimately, \name{} demonstrates that reliable evaluation requires moving from monolithic text generation to verifiable atomic units, transforming performance measurement into an actionable diagnostic tool for multimodal systems.

\bibliography{mtss}

@inproceedings{fu2025videomme,
  title={Video-{MME}: The First-Ever Comprehensive Evaluation Benchmark of Multi-modal {LLM}s in Video Analysis},
  author={Fu, Chaoyou and Dai, Yuhan and Luo, Yongdong and Li, Lei and Ren, Shuhuai and Zhang, Renrui and Wang, Zihan and Zhou, Chenyu and Shen, Yunhang and Zhang, Mengdan and Chen, Peixian and Li, Yanwei and Lin, Shaohui and Zhao, Sirui and Li, Ke and Xu, Tong and Zheng, Xiawu and Chen, Enhong and Shan, Caifeng and He, Ran and Sun, Xing},
  booktitle={Proceedings of the IEEE/CVF Conference on Computer Vision and Pattern Recognition (CVPR)},
  pages={24108--24118},
  year={2025}
}

@article{hu2025videommmu,
  title={Video-{MMMU}: Evaluating Knowledge Acquisition from Multi-Discipline Professional Videos},
  author={Hu, Kairui and Wu, Penghao and Pu, Fanyi and Xiao, Wang and Zhang, Yuanhan and Yue, Xiang and Li, Bo and Liu, Ziwei},
  journal={arXiv preprint arXiv:2501.13826},
  eprint={2501.13826},
  doi={10.48550/arXiv.2501.13826},
  url={https://arxiv.org/abs/2501.13826},
  year={2025}
}

@article{wu2024longvideobench,
  title={{LongVideoBench}: A Benchmark for Long-context Interleaved Video-Language Understanding},
  author={Wu, Haoning and Li, Dongxu and Chen, Bei and Li, Junnan},
  journal={arXiv preprint arXiv:2407.15754},
  eprint={2407.15754},
  doi={10.48550/arXiv.2407.15754},
  url={https://arxiv.org/abs/2407.15754},
  year={2024}
}

@article{wang2025lvbench,
  title={{LVBench}: An Extreme Long Video Understanding Benchmark},
  author={Wang, Weihan and He, Zehai and Hong, Wenyi and Cheng, Yean and Zhang, Xiaohan and Qi, Ji and Gu, Xiaotao and Huang, Shiyu and Xu, Bin and Dong, Yuxiao and Ding, Ming and Tang, Jie},
  journal={arXiv preprint arXiv:2406.08035},
  eprint={2406.08035},
  doi={10.48550/arXiv.2406.08035},
  url={https://arxiv.org/abs/2406.08035},
  year={2025}
}

@article{zhou2025mlvu,
  title={{MLVU}: Benchmarking Multi-task Long Video Understanding},
  author={Zhou, Junjie and Shu, Yan and Zhao, Bo and Wu, Boya and Liang, Zhengyang and Xiao, Shitao and Qin, Minghao and Yang, Xi and Xiong, Yongping and Zhang, Bo and Huang, Tiejun and Liu, Zheng},
  journal={arXiv preprint arXiv:2406.04264},
  eprint={2406.04264},
  doi={10.48550/arXiv.2406.04264},
  url={https://arxiv.org/abs/2406.04264},
  year={2025}
}

@inproceedings{xiao2021nextqa,
  title={{NExT-QA}: Next Phase of Question-Answering to Explaining Temporal Actions},
  author={Xiao, Junbin and Shang, Xindi and Yao, Angela and Chua, Tat-Seng},
  booktitle={Proceedings of the IEEE/CVF Conference on Computer Vision and Pattern Recognition (CVPR)},
  month={June},
  pages={9777--9786},
  year={2021}
}

@article{cai2024temporalbench,
  title={{TemporalBench}: Benchmarking Fine-grained Temporal Understanding for Multimodal Video Models},
  author={Cai, Mu and Tan, Reuben and Zhang, Jianrui and Zou, Bocheng and Zhang, Kai and Yao, Feng and Zhu, Fangrui and Gu, Jing and Zhong, Yiwu and Shang, Yuzhang and Dou, Yao and Park, Jaden and Gao, Jianfeng and Lee, Yong Jae and Yang, Jianwei},
  journal={arXiv preprint arXiv:2410.10818},
  eprint={2410.10818},
  doi={10.48550/arXiv.2410.10818},
  url={https://arxiv.org/abs/2410.10818},
  year={2024}
}

@misc{wu2025ugcvideocaptioner,
      title={UGC-VideoCaptioner: An Omni UGC Video Detail Caption Model and New Benchmarks}, 
      author={Peiran Wu and Yunze Liu and Zhengdong Zhu and Enmin Zhou and Junxiao Shen},
      year={2025},
      eprint={2507.11336},
      archivePrefix={arXiv},
      primaryClass={cs.CV},
      url={https://arxiv.org/abs/2507.11336}, 
}

@article{yang2025qwen3,
  title={Qwen3 Technical Report},
  author={Yang, An and Li, Anfeng and Yang, Baosong and Zhang, Beichen and Hui, Binyuan and Zheng, Bo and Yu, Bowen and Gao, Chang and Huang, Chengen and Lv, Chenxu and Zheng, Chujie and Liu, Dayiheng and Zhou, Fan and Huang, Fei and Hu, Feng and Ge, Hao and Wei, Haoran and Lin, Huan and Tang, Jialong and Yang, Jian and Tu, Jianhong and Zhang, Jianwei and Yang, Jianxin and Yang, Jiaxi and Zhou, Jing and Zhou, Jingren and Lin, Junyang and Dang, Kai and Bao, Keqin and Yang, Kexin and Yu, Le and Deng, Lianghao and Li, Mei and Xue, Mingfeng and Li, Mingze and Zhang, Pei and Wang, Peng and Zhu, Qin and Men, Rui and Gao, Ruize and Liu, Shixuan and Luo, Shuang and Li, Tianhao and Tang, Tianyi and Yin, Wenbiao and Ren, Xingzhang and Wang, Xinyu and Zhang, Xinyu and Ren, Xuancheng and Fan, Yang and Su, Yang and Zhang, Yichang and Zhang, Yinger and Wan, Yu and Liu, Yuqiong and Wang, Zekun and Cui, Zeyu and Zhang, Zhenru and Zhou, Zhipeng and Qiu, Zihan},
  journal={arXiv preprint arXiv:2505.09388},
  eprint={2505.09388},
  doi={10.48550/arXiv.2505.09388},
  url={https://arxiv.org/abs/2505.09388},
  year={2025}
}

@article{bai2025qwen25vl,
  title={Qwen2.5-{VL} Technical Report},
  author={Bai, Shuai and Chen, Keqin and Liu, Xuejing and Wang, Jialin and Ge, Wenbin and Song, Sibo and Dang, Kai and Wang, Peng and Wang, Shijie and Tang, Jun and Zhong, Humen and Zhu, Yuanzhi and Yang, Mingkun and Li, Zhaohai and Wan, Jianqiang and Wang, Pengfei and Ding, Wei and Fu, Zheren and Xu, Yiheng and Ye, Jiabo and Zhang, Xi and Xie, Tianbao and Cheng, Zesen and Zhang, Hang and Yang, Zhibo and Xu, Haiyang and Lin, Junyang},
  journal={arXiv preprint arXiv:2502.13923},
  eprint={2502.13923},
  doi={10.48550/arXiv.2502.13923},
  url={https://arxiv.org/abs/2502.13923},
  year={2025}
}

@article{xu2025qwen25omni,
  title={Qwen2.5-Omni Technical Report},
  author={Xu, Jin and Guo, Zhifang and He, Jinzheng and Hu, Hangrui and He, Ting and Bai, Shuai and Chen, Keqin and Wang, Jialin and Fan, Yang and Dang, Kai and Zhang, Bin and Wang, Xiong and Chu, Yunfei and Lin, Junyang},
  journal={arXiv preprint arXiv:2503.20215},
  eprint={2503.20215},
  doi={10.48550/arXiv.2503.20215},
  url={https://arxiv.org/abs/2503.20215},
  year={2025}
}

@article{xu2025qwen3omni,
  title={Qwen3-Omni Technical Report},
  author={Xu, Jin and Guo, Zhifang and Hu, Hangrui and Chu, Yunfei and Wang, Xiong and He, Jinzheng and Wang, Yuxuan and Shi, Xian and He, Ting and Zhu, Xinfa and Lv, Yuanjun and Wang, Yongqi and Guo, Dake and Wang, He and Ma, Linhan and Zhang, Pei and Zhang, Xinyu and Hao, Hongkun and Guo, Zishan and Yang, Baosong and Zhang, Bin and Ma, Ziyang and Wei, Xipin and Bai, Shuai and Chen, Keqin and Liu, Xuejing and Wang, Peng and Yang, Mingkun and Liu, Dayiheng and Ren, Xingzhang and Zheng, Bo and Men, Rui and Zhou, Fan and Yu, Bowen and Yang, Jianxin and Yu, Le and Zhou, Jingren and Lin, Junyang},
  journal={arXiv preprint arXiv:2509.17765},
  eprint={2509.17765},
  doi={10.48550/arXiv.2509.17765},
  url={https://arxiv.org/abs/2509.17765},
  year={2025}
}

@article{qwenteam2026qwen35omni,
  title={Qwen3.5-Omni Technical Report},
  author={{Qwen Team}},
  journal={arXiv preprint arXiv:2604.15804},
  eprint={2604.15804},
  doi={10.48550/arXiv.2604.15804},
  url={https://arxiv.org/abs/2604.15804},
  year={2026}
}

@article{li2024baichuanomni,
  title={Baichuan-Omni Technical Report},
  author={Li, Yadong and Sun, Haoze and Lin, Mingan and Li, Tianpeng and Dong, Guosheng and Zhang, Tao and Ding, Bowen and Song, Wei and Cheng, Zhenglin and Huo, Yuqi and Chen, Song and Li, Xu and Pan, Da and Zhang, Shusen and Wu, Xin and Liang, Zheng and Liu, Jun and Zhang, Tao and Lu, Keer and Zhao, Yaqi and Shen, Yanjun and Yang, Fan and Yu, Kaicheng and Lin, Tao and Xu, Jianhua and Zhou, Zenan and Chen, Weipeng},
  journal={arXiv preprint arXiv:2410.08565},
  eprint={2410.08565},
  doi={10.48550/arXiv.2410.08565},
  url={https://arxiv.org/abs/2410.08565},
  year={2024}
}

@article{liu2025ola,
  title={Ola: Pushing the Frontiers of Omni-Modal Language Model},
  author={Liu, Zuyan and Dong, Yuhao and Wang, Jiahui and Liu, Ziwei and Hu, Winston and Lu, Jiwen and Rao, Yongming},
  journal={arXiv preprint arXiv:2502.04328},
  eprint={2502.04328},
  doi={10.48550/arXiv.2502.04328},
  url={https://arxiv.org/abs/2502.04328},
  year={2025}
}

@article{zhao2025humanomni,
  title={HumanOmni: A Large Vision-Speech Language Model for Human-Centric Video Understanding},
  author={Zhao, Jiaxing and Yang, Qize and Peng, Yixing and Bai, Detao and Yao, Shimin and Sun, Boyuan and Chen, Xiang and Fu, Shenghao and Chen, Weixuan and Wei, Xihan and Bo, Liefeng},
  journal={arXiv preprint arXiv:2501.15111},
  eprint={2501.15111},
  doi={10.48550/arXiv.2501.15111},
  url={https://arxiv.org/abs/2501.15111},
  year={2025}
}

@article{yang2025humanomniv2,
  title={HumanOmniV2: From Understanding to Omni-Modal Reasoning with Context},
  author={Yang, Qize and Yao, Shimin and Chen, Weixuan and Fu, Shenghao and Bai, Detao and Zhao, Jiaxing and Sun, Boyuan and Yin, Bowen and Wei, Xihan and Zhou, Jingren},
  journal={arXiv preprint arXiv:2506.21277},
  eprint={2506.21277},
  doi={10.48550/arXiv.2506.21277},
  url={https://arxiv.org/abs/2506.21277},
  year={2025}
}

@article{zhong2025omnir1,
  title={Omni-R1: Reinforcement Learning for Omnimodal Reasoning via Two-System Collaboration},
  author={Zhong, Hao and Zhu, Muzhi and Du, Zongze and Huang, Zheng and Zhao, Canyu and Liu, Mingyu and Wang, Wen and Chen, Hao and Shen, Chunhua},
  journal={arXiv preprint arXiv:2505.20256},
  eprint={2505.20256},
  doi={10.48550/arXiv.2505.20256},
  url={https://arxiv.org/abs/2505.20256},
  year={2025}
}

@article{zhao2025r1omni,
  title={R1-Omni: Explainable Omni-Multimodal Emotion Recognition with Reinforcement Learning},
  author={Zhao, Jiaxing and Wei, Xihan and Bo, Liefeng},
  journal={arXiv preprint arXiv:2503.05379},
  eprint={2503.05379},
  doi={10.48550/arXiv.2503.05379},
  url={https://arxiv.org/abs/2503.05379},
  year={2025}
}

@article{xing2025echoinkr1,
  title={EchoInk-R1: Exploring Audio-Visual Reasoning in Multimodal LLMs via Reinforcement Learning},
  author={Xing, Zhenghao and Hu, Xiaowei and Fu, Chi-Wing and Wang, Wenhai and Dai, Jifeng and Heng, Pheng-Ann},
  journal={arXiv preprint arXiv:2505.04623},
  eprint={2505.04623},
  doi={10.48550/arXiv.2505.04623},
  url={https://arxiv.org/abs/2505.04623},
  year={2025}
}

@inproceedings{xu2026multihaystackbenchmarkingmultimodalretrieval,
      title={MultiHaystack: Benchmarking Multimodal Retrieval and Reasoning over 40K Images, Videos, and Documents}, 
      author={Dannong Xu and Zhongyu Yang and Jun Chen and Yingfang Yuan and Ming Hu and Lei Sun and Luc Van Gool and Danda Pani Paudel and Chun-Mei Feng},
      booktitle={ECCV},
      year={2026},
      eprint={2603.05697},
      archivePrefix={arXiv},
      primaryClass={cs.CV},
      url={https://arxiv.org/abs/2603.05697}, 
}

@inproceedings{
yang2026do,
title={Do Vision and Text Cues Exhibit Evidential Coupling? {UFO}: A Benchmark for Compositional Multimodal Reasoning in Unified Models},
author={Zhongyu Yang and Dannong Xu and Yonghan Zhang and Kefan Chen and Xinyi Wang and Yang Xu and Wei Pang and Yingfang Yuan},
booktitle={Forty-third International Conference on Machine Learning},
year={2026},
url={https://openreview.net/forum?id=6UKaYYRM3h}
}

@misc{qwen35omni,
      title={Qwen3.5-Omni Technical Report}, 
      author={Qwen Team},
      year={2026},
      eprint={2604.15804},
      archivePrefix={arXiv},
      primaryClass={cs.CL},
      url={https://arxiv.org/abs/2604.15804}, 
}

@misc{minicpmo45,
      title={MiniCPM-o 4.5: Towards Real-Time Full-Duplex Omni-Modal Interaction}, 
      author={Junbo Cui and Bokai Xu and Chongyi Wang and Tianyu Yu and Weiyue Sun and Yingjing Xu and Tianran Wang and Zhihui He and Wenshuo Ma and Tianchi Cai and Jiancheng Gui and Luoyuan Zhang and Xian Sun and Fuwei Huang and Moye Chen and Zhuo Lin and Hanyu Liu and Qingxin Gui and Qingzhe Han and Yuyang Wen and Huiping Liu and Rongkang Wang and Yaqi Zhang and Hongliang Wei and Chi Chen and You Li and Kechen Fang and Jie Zhou and Yuxuan Li and Guoyang Zeng and Chaojun Xiao and Yankai Lin and Xu Han and Maosong Sun and Zhiyuan Liu and Yuan Yao},
      year={2026},
      eprint={2604.27393},
      archivePrefix={arXiv},
      primaryClass={cs.CL},
      url={https://arxiv.org/abs/2604.27393}, 
}

@misc{omnicapif,
      title={OmniCap-IF: Benchmarking and Improving Instruction Following Abilities for Omni-Video Captioning}, 
      author={Jiahao Wang and An Ping and Yanghai Wang and Yuanxing Zhang and Shihao Li and Hanyan Bian and Yichi Ren and Yize Zhang and Han Wang and Haowen Chen and Junze Li and Jiaqi Wang and Yiyang Hu and Zhuze Xu and Zijie Zhang and Jiaheng Liu},
      year={2026},
      eprint={2606.08572},
      archivePrefix={arXiv},
      primaryClass={cs.CV},
      url={https://arxiv.org/abs/2606.08572}, 
}

@misc{mtss,
      title={Script-a-Video: Deep Structured Audio-visual Captions via Factorized Streams and Relational Grounding}, 
      author={Tencent Hunyuan Team},
      year={2026},
      eprint={2604.11244},
      archivePrefix={arXiv},
      primaryClass={cs.CV},
      url={https://arxiv.org/abs/2604.11244}, 
}

@misc{gemini25,
      title={Gemini 2.5: Pushing the Frontier with Advanced Reasoning, Multimodality, Long Context, and Next Generation Agentic Capabilities}, 
      author={Gheorghe Comanici and Eric Bieber and Mike Schaekermann and others},
      year={2025},
      eprint={2507.06261},
      archivePrefix={arXiv},
      primaryClass={cs.CL},
      url={https://arxiv.org/abs/2507.06261}, 
}

@article{sun2024videosalmonn,
  title={video-SALMONN: Speech-Enhanced Audio-Visual Large Language Models},
  author={Sun, Guangzhi and Yu, Wenyi and Tang, Changli and Chen, Xianzhao and Tan, Tian and Li, Wei and Lu, Lu and Ma, Zejun and Wang, Yuxuan and Zhang, Chao},
  journal={arXiv preprint arXiv:2406.15704},
  eprint={2406.15704},
  doi={10.48550/arXiv.2406.15704},
  url={https://arxiv.org/abs/2406.15704},
  year={2024}
}

@article{fu2025vita15,
  title={VITA-1.5: Towards GPT-4o Level Real-Time Vision and Speech Interaction},
  author={Fu, Chaoyou and Lin, Haojia and Wang, Xiong and Zhang, Yi-Fan and Shen, Yunhang and Liu, Xiaoyu and Cao, Haoyu and Long, Zuwei and Gao, Heting and Li, Ke and Ma, Long and Zheng, Xiawu and Ji, Rongrong and Sun, Xing and Shan, Caifeng and He, Ran},
  journal={arXiv preprint arXiv:2501.01957},
  eprint={2501.01957},
  doi={10.48550/arXiv.2501.01957},
  url={https://arxiv.org/abs/2501.01957},
  year={2025}
}

@article{li2026omnivideobench,
  title={OmniVideoBench: Towards Audio-Visual Understanding Evaluation for Omni MLLMs},
  author={Li, Caorui and Chen, Yu and Ji, Yiyan and Xu, Jin and Cui, Zhenyu and Li, Shihao and Zhang, Yuanxing and Wang, Wentao and Song, Zhenghao and Zhang, Dingling and He, Ying and Liu, Haoxiang and Wang, Yuxuan and Wang, Qiufeng and Tang, Jiafu and Wu, Zhenhe and Luo, Jiehui and Pan, Zhiyu and Xie, Weihao and Zhang, Chenchen and Wang, Zhaohui and Tian, Jiayi and Wang, Yanghai and Cao, Zhe and Dai, Minxin and Wang, Ke and Wen, Runzhe and Ma, Yinghao and Pan, Yaning and Chang, Sungkyun and Taheri, Termeh and Xia, Haiwen and Plachouras, Christos and Benetos, Emmanouil and Li, Yizhi and Zhang, Ge and Yang, Jian and Peng, Tianhao and Wang, Zili and Liu, Minghao and Peng, Junran and Zhang, Zhaoxiang and Liu, Jiaheng},
  journal={arXiv preprint arXiv:2510.10689},
  eprint={2510.10689},
  doi={10.48550/arXiv.2510.10689},
  url={https://arxiv.org/abs/2510.10689},
  year={2026}
}

@article{yang2026paravt,
  title={ParaVT: Taming the Tool Prior Paradox for Parallel Tool Use in Agentic Video Reinforcement Learning},
  author={Yang, Zuhao and Zhang, Kaichen and Wang, Sudong and Wu, Keming and Yang, Zhongyu and Li, Bo and Qi, Xiaojuan and Lu, Shijian and Li, Xingxuan and Bing, Lidong},
  journal={arXiv preprint arXiv:2605.20342},
  year={2026}
}

@inproceedings{Timeexpert,
  author={Yang, Zuhao and Yu, Yingchen and Zhao, Yunqing and Lu, Shijian and Bai, Song},
  booktitle={2025 IEEE/CVF International Conference on Computer Vision (ICCV)}, 
  title={Timeexpert: an Expert-Guided Video Llm for Video Temporal Grounding}, 
  year={2025},
  volume={},
  number={},
  pages={24286-24296},
  doi={10.1109/ICCV51701.2025.02251}
}

@article{he2026vgi,
  title={VGI-BENCH: Probing Visual Intelligence in Video Generation Models},
  author={He, Xuan and Wei, Cong and Cheng, Yuhao and Ma, Linrui and Zhang, Yuxuan and Li, Zuojun and Wen, Yuhao and Liu, Zeyi and Hao, Yuren and Cai, Songcheng and others},
  journal={arXiv preprint arXiv:2608.19583},
  year={2026}
}

@article{wu2026worldreasonbench,
  title={WorldReasonBench: Human-Aligned Stress Testing of Video Generators as Future World-State Predictors},
  author={Wu, Keming and Cui, Yijing and Xue, Wenhan and Wang, Qijie and Luo, Xuan and Feng, Zhiyuan and Yang, Zuhao and Wang, Sudong and Jiang, Sicong and Zhu, Haowei and others},
  journal={arXiv preprint arXiv:2605.10434},
  year={2026}
}

@article{wu2026streamopd,
  title={StreamOPD: A Post-Training Recipe with Spatio-Temporal Cue Gating for Streaming Video Understanding},
  author={Wu, Keming and Wang, Baoyi and Zhang, Kaichen and An, Xiang and Yang, Zuhao and Wang, Sudong and Zhu, Haowei and Huang, Tingxuan and Gao, Hongcheng and Wang, Bin},
  journal={arXiv preprint arXiv:2608.16320},
  year={2026}
}

@article{zhang2026aero,
  title={Aero Realtime: Fully Aligned Input-Output Streams for Low-Latency Streaming Multimodal Generation},
  author={Zhang, Kaichen and Huang, Wei and Wu, Keming and Li, Bo and Qi, Xiaojuan},
  journal={arXiv preprint arXiv:2608.08469},
  year={2026}
}

@misc{he2026omniduplexeval,
      title={Omni-DuplexEval: Evaluating Real-time Duplex Omni-modal Interaction}, 
      author={Chaoqun He and Mingyang Xiang and Yingjing Xu and Bokai Xu and Junbo Cui and Jie Zhou and Yuan Yao and Lijie Wen},
      year={2026},
      eprint={2605.17360},
      archivePrefix={arXiv},
      primaryClass={cs.CV},
      url={https://arxiv.org/abs/2605.17360}, 
}

@inproceedings{
chen2026omnivideor1,
title={OmniVideo-R1: Reinforcing Audio-visual Reasoning with Query Intention and Modality Attention},
author={Zhangquan Chen and Jiale Tao and Ruihuang Li and Yihao Hu and Ruitao Chen and Zhantao Yang and Xinlei Yu and Haodong Jing and Manyuan Zhang and Shuai Shao and Biao Wang and Qinglin Lu and Ruqi Huang},
booktitle={Forty-third International Conference on Machine Learning},
year={2026},
url={https://openreview.net/forum?id=he06cvibXv}
}

@inproceedings{inex,
  title={InEx: Hallucination Mitigation via Introspection and Cross-Modal Multi-Agent Collaboration},
  author={Zhongyu Yang and Ying-Fang Yuan and Xuanming Jiang and Baoyi An and Wei Pang},
  booktitle={AAAI Conference on Artificial Intelligence},
  year={2025},
  url={https://api.semanticscholar.org/CorpusID:283458455}
}

@InProceedings{yang2026svagent,
    author    = {Yang, Zhongyu and Yang, Zuhao and Zhan, Shuo and Yue, Tan and Pang, Wei and Yuan, Yingfang},
    title     = {SVAgent: Storyline-guided Long Video Understanding via Cross-Modal Multi-Agent Collaboration},
    booktitle = {Proceedings of the IEEE/CVF Conference on Computer Vision and Pattern Recognition (CVPR)},
    month     = {June},
    year      = {2026},
    pages     = {24062-24072}
}

@inproceedings{yang2026longvt,
  title={LongVT: Incentivizing ``Thinking with Long Videos'' via Native Tool Calling},
  author={Yang, Zuhao and Wang, Sudong and Zhang, Kaichen and Wu, Keming and Leng, Sicong and Zhang, Yifan and Li, Bo and Qin, Chengwei and Lu, Shijian and Li, Xingxuan and Bing, Lidong},
  booktitle={CVPR},
  year={2026}
}

@article{hong2026worldsense,
  title={WorldSense: Evaluating Real-world Omnimodal Understanding for Multimodal LLMs},
  author={Hong, Jack and Yan, Shilin and Cai, Jiayin and Jiang, Xiaolong and Hu, Yao and Xie, Weidi},
  journal={arXiv preprint arXiv:2502.04326},
  eprint={2502.04326},
  doi={10.48550/arXiv.2502.04326},
  url={https://arxiv.org/abs/2502.04326},
  year={2026}
}

@misc{nvidia2026nemotron3nanoomni,
  title={Nemotron 3 Nano Omni: Efficient and Open Multimodal Intelligence},
  author={{NVIDIA} and Deshmukh, Amala Sanjay and Chumachenko, Kateryna and Rintamaki, Tuomas and Le, Matthieu and others},
  year={2026},
  eprint={2604.24954},
  archivePrefix={arXiv},
  primaryClass={cs.LG},
  url={https://arxiv.org/abs/2604.24954},
}

@article{pu2026omniscript,
  title={OmniScript: Towards Audio-Visual Script Generation for Long-Form Cinematic Video},
  author={Pu, Junfu and Chen, Yuxin and Wang, Teng and Shan, Ying},
  journal={arXiv preprint arXiv:2604.11102},
  eprint={2604.11102},
  doi={10.48550/arXiv.2604.11102},
  url={https://arxiv.org/abs/2604.11102},
  year={2026}
}

@article{li2026towards,
  title={Towards Universal Video MLLMs with Attribute-Structured and Quality-Verified Instructions},
  author={Li, Yunheng and Zhang, Hengrui and Guo, Meng-Hao and Gao, Wenzhao and Jia, Shaoyong and Jiao, Shaohui and Hou, Qibin and Cheng, Ming-Ming},
  journal={arXiv preprint arXiv:2602.13013},
  year={2026}
}

@article{chai2025auroracap,
  title={AuroraCap: Efficient, Performant Video Detailed Captioning and a New Benchmark},
  author={Chai, Wenhao and Song, Enxin and Du, Yilun and Meng, Chenlin and Madhavan, Vashisht and Bar-Tal, Omer and Hwang, Jenq-Neng and Xie, Saining and Manning, Christopher D.},
  journal={arXiv preprint arXiv:2410.03051},
  eprint={2410.03051},
  doi={10.48550/arXiv.2410.03051},
  url={https://arxiv.org/abs/2410.03051},
  year={2025}
}

@inproceedings{chen2025vidcapbench,
  title={VidCapBench: A Comprehensive Benchmark of Video Captioning for Controllable Text-to-Video Generation},
  author={Chen, Xinlong and Zhang, Yuanxing and Rao, Chongling and Guan, Yushuo and Liu, Jiaheng and Zhang, Fuzheng and Song, Chengru and Liu, Qiang and Zhang, Di and Tan, Tieniu},
  editor={Che, Wanxiang and Nabende, Joyce and Shutova, Ekaterina and Pilehvar, Mohammad Taher},
  booktitle={Findings of the Association for Computational Linguistics: ACL 2025},
  month={jul},
  address={Vienna, Austria},
  publisher={Association for Computational Linguistics},
  url={https://aclanthology.org/2025.findings-acl.449/},
  pages={8543--8563},
  doi={10.18653/v1/2025.findings-acl.449},
  isbn={979-8-89176-256-5},
  year={2025}
}

@article{zhang2026vcapsbench,
  title={VCapsBench: A Large-scale Fine-grained Benchmark for Video Caption Quality Evaluation},
  author={Zhang, Shi-Xue and Wang, Hongfa and Huang, Duojun and Li, Xin and Zhu, Xiaobin and Yin, Xu-Cheng},
  journal={Proceedings of the AAAI Conference on Artificial Intelligence},
  volume={40},
  number={15},
  pages={12726--12734},
  doi={10.1609/aaai.v40i15.38269},
  url={https://ojs.aaai.org/index.php/AAAI/article/view/38269},
  year={2026}
}

@article{liu2025capability,
  title={CAPability: A Comprehensive Visual Caption Benchmark for Evaluating Both Correctness and Thoroughness},
  author={Liu, Zhihang and Xie, Chen-Wei and Wen, Bin and Yu, Feiwu and Chen, Jixuan and Li, Pandeng and Zhang, Boqiang and Yang, Nianzu and Li, Yinglu and Gao, Zuan and Zheng, Yun and Xie, Hongtao},
  journal={arXiv preprint arXiv:2502.14914},
  eprint={2502.14914},
  doi={10.48550/arXiv.2502.14914},
  url={https://arxiv.org/abs/2502.14914},
  year={2025}
}

@inproceedings{xu2026carebench,
  title={CaReBench: A Fine-grained Benchmark for Video Captioning and Retrieval},
  author={Xu, Yifan and Li, Xinhao and Yang, Yichun and Meng, Desen and Huang, Rui and Wang, Limin},
  booktitle={ICLR},
  year={2026}
}

@article{tang2025videosalmonn2,
  title={video-SALMONN 2: Caption-Enhanced Audio-Visual Large Language Models},
  author={Tang, Changli and Li, Yixuan and Yang, Yudong and Zhuang, Jimin and Sun, Guangzhi and Li, Wei and Ma, Zejun and Zhang, Chao},
  journal={arXiv preprint arXiv:2506.15220},
  eprint={2506.15220},
  doi={10.48550/arXiv.2506.15220},
  url={https://arxiv.org/abs/2506.15220},
  year={2025}
}

@article{chen2025avocado,
  title={AVoCaDO: An Audiovisual Video Captioner Driven by Temporal Orchestration},
  author={Chen, Xinlong and Ding, Yue and Lin, Weihong and Hua, Jingyun and Yao, Linli and Shi, Yang and Li, Bozhou and Zhang, Yuanxing and Liu, Qiang and Wan, Pengfei and Wang, Liang and Tan, Tieniu},
  journal={arXiv preprint arXiv:2510.10395},
  eprint={2510.10395},
  doi={10.48550/arXiv.2510.10395},
  url={https://arxiv.org/abs/2510.10395},
  year={2025}
}

@article{goel2026mmou,
  title={MMOU: A Massive Multi-Task Omni Understanding and Reasoning Benchmark for Long and Complex Real-World Videos},
  author={Goel, Arushi and Ghosh, Sreyan and Agarwal, Vatsal and Anand, Nishit and Jayakumar, Kaousheik and Koroshinadze, Lasha and Xu, Yao and Lyons, Katie and Case, James and Sapra, Karan and Shih, Kevin J. and Gururani, Siddharth and Shrivastava, Abhinav and Duraiswami, Ramani and Manocha, Dinesh and Tao, Andrew and Catanzaro, Bryan and Shoeybi, Mohammad and Ping, Wei},
  journal={arXiv preprint arXiv:2603.14145},
  eprint={2603.14145},
  doi={10.48550/arXiv.2603.14145},
  url={https://arxiv.org/abs/2603.14145},
  year={2026}
}

@inproceedings{geng2025longvale,
  title={LongVALE: Vision-Audio-Language-Event Benchmark Towards Time-Aware Omni-Modal Perception of Long Videos},
  author={Geng, Tiantian and Zhang, Jinrui and Wang, Qingni and Wang, Teng and Duan, Jinming and Zheng, Feng},
  booktitle={Proceedings of the IEEE/CVF Conference on Computer Vision and Pattern Recognition (CVPR)},
  month={June},
  pages={18959--18969},
  year={2025}
}

@article{ma2026omnicaptioner,
  title={Omni-Captioner: Data Pipeline, Models, and Benchmark for Omni Detailed Perception},
  author={Ma, Ziyang and Xu, Ruiyang and Xing, Zhenghao and Chu, Yunfei and Wang, Yuxuan and He, Jinzheng and Xu, Jin and Heng, Pheng-Ann and Yu, Kai and Lin, Junyang and Chng, Eng Siong and Chen, Xie},
  journal={arXiv preprint arXiv:2510.12720},
  eprint={2510.12720},
  doi={10.48550/arXiv.2510.12720},
  url={https://arxiv.org/abs/2510.12720},
  year={2026}
}

@article{yao2026timechatcaptioner,
  title={TimeChat-Captioner: Scripting Multi-Scene Videos with Time-Aware and Structural Audio-Visual Captions},
  author={Yao, Linli and Wei, Yuancheng and Zhang, Yaojie and Li, Lei and Chen, Xinlong and Song, Feifan and Wang, Ziyue and Ouyang, Kun and Liu, Yuanxin and Kong, Lingpeng and Liu, Qi and Wan, Pengfei and Gai, Kun and Zhang, Yuanxing and Sun, Xu},
  journal={arXiv preprint arXiv:2602.08711},
  eprint={2602.08711},
  doi={10.48550/arXiv.2602.08711},
  url={https://arxiv.org/abs/2602.08711},
  year={2026}
}

@article{chen2026diadem,
  title={DiaDem: Advancing Dialogue Descriptions in Audiovisual Video Captioning for Multimodal Large Language Models},
  author={Chen, Xinlong and Lin, Weihong and Hua, Jingyun and Yao, Linli and Ding, Yue and Li, Bozhou and Zeng, Bohan and Shi, Yang and Liu, Qiang and Zhang, Yuanxing and Wan, Pengfei and Wang, Liang and Tan, Tieniu},
  journal={arXiv preprint arXiv:2601.19267},
  eprint={2601.19267},
  doi={10.48550/arXiv.2601.19267},
  url={https://arxiv.org/abs/2601.19267},
  year={2026}
}

@misc{gemini2024,
  title={Gemini 3.1: Best for complex tasks and bringing creative concepts to life},
  author={{Gemini Team}},
  year={2026},
  url={https://deepmind.google/models/gemini/pro/}
}

@inproceedings{vedantam2015cider,
  title={{CIDEr}: Consensus-based image description evaluation},
  author={Vedantam, Ramakrishna and Lawrence Zitnick, C and Parikh, Devi},
  booktitle={Proceedings of the IEEE conference on computer vision and pattern recognition},
  pages={4566--4575},
  year={2015}
}

@inproceedings{banerjee2005meteor,
  title={{METEOR}: An automatic metric for {MT} evaluation with improved correlation with human judgments},
  author={Banerjee, Satanjeev and Lavie, Alon},
  booktitle={Proceedings of the acl workshop on intrinsic and extrinsic evaluation measures for machine translation and/or summarization},
  pages={65--72},
  year={2005}
}

@article{lin2004rouge,
  title={{ROUGE}: A package for automatic evaluation of summaries},
  author={Lin, Chin-Yew},
  journal={Text summarization branches out},
  pages={74--81},
  year={2004}
}

@inproceedings{fujita2020soda,
  title={{SODA}: Story Oriented Dense Video Captioning Evaluation Framework},
  author={Fujita, Soichiro and Tsutsui, Satoshi and Ejiri, Yoshihiko and Shikida, Maris and Yamasaki, Toshihiko},
  booktitle={European Conference on Computer Vision},
  pages={517--531},
  year={2020}
}

@inproceedings{kwon2023efficient,
  title={Efficient Memory Management for Large Language Model Serving with {PagedAttention}},
  author={Kwon, Woosuk and Li, Zhuohan and Zhuang, Siyuan and Sheng, Ying and Zheng, Lianmin and Yu, Cody Hao and Gonzalez, Joseph E. and Zhang, Hao and Stoica, Ion},
  booktitle={Proceedings of the 29th ACM Symposium on Operating Systems Principles (SOSP)},
  pages={611--626},
  year={2023}
}

@misc{bytedance2026seed2,
      title={Seed2.0 Model Card: Towards Intelligence Frontier for Real-World Complexity}, 
      author={{ByteDance Seed Team}},
      year={2026},
      eprint={2607.00248},
      archivePrefix={arXiv},
      primaryClass={cs.CL},
      url={https://arxiv.org/abs/2607.00248}, 
}

@misc{xiaomi2026mimo25,
  title={{MiMo-V2.5}: A Native Omnimodal Model with Agentic Capabilities},
  author={{Xiaomi LLM-Core Team}},
  year={2026},
  howpublished={\url{https://huggingface.co/XiaomiMiMo/MiMo-V2.5}},
  note={Model card; language backbone described in the MiMo-V2-Flash technical report}
}

@misc{yao2024minicpmv,
      title={{MiniCPM-V}: A {GPT-4V} Level {MLLM} on Your Phone}, 
      author={Yuan Yao and Tianyu Yu and Ao Zhang and Chongyi Wang and Junbo Cui and Hongji Zhu and Tianchi Cai and Haoyu Li and Weilin Zhao and Zhihui He and others},
      year={2024},
      eprint={2408.01800},
      archivePrefix={arXiv},
      primaryClass={cs.CV},
      url={https://arxiv.org/abs/2408.01800}, 
}
\bibliographystyle{plainnat}

\newpage
\appendix

\renewcommand{\thefigure}{A.\arabic{figure}}
\renewcommand{\thetable}{A.\arabic{table}}
\renewcommand{\theequation}{A.\arabic{equation}}
\setcounter{figure}{0}
\setcounter{theorem}{0}
\setcounter{table}{0}
\setcounter{equation}{0}

\begin{center}
\Large\textbf{Appendix}
\end{center}

\addtocontents{toc}{\protect\setcounter{tocdepth}{1}}
\tableofcontents
\vspace{2em}


\appendix

\newpage
\section*{Overview of Appendix}
\label{app:overview}

This appendix provides comprehensive supplementary materials to ensure the reproducibility and transparency of the \name{} benchmark. The appendix is structured as follows:

\begin{itemize}[leftmargin=*]
    \item \textbf{Implementation Details (Appendix~\ref{app:implementation}):} Details the benchmark construction process, data statistics, and the full experimental setup used for evaluation.
    \item \textbf{More Experimental Results (Appendix~\ref{app:more_results}):} Provides additional quantitative results, detailed failure analyses, and qualitative Case Studies.
    \item \textbf{Limitations and Broader Impact (Appendix~\ref{app:broader_impact}):} Discusses the boundaries of our evaluation scope and the broader impacts of structured video benchmarking.
    \item \textbf{Ethical Considerations (Appendix~\ref{app:ethical_considerations}):} Addresses privacy, consent, and the ethical use of audio-visual datasets.
    \item \textbf{Dataset Licenses and Usage (Appendix~\ref{app:dataset_licenses}):} Documents the licenses and fair-use protocols for all underlying video sources and utilized models.
    \item \textbf{Instruction Templates and Prompts (Appendix~\ref{app:instruction_templates}):} Provides the complete, executable prompts used for unit generation, local semantic judging, and structural parsing.
\end{itemize}

\renewcommand{\thefigure}{A.\arabic{figure}}
\renewcommand{\thetable}{A.\arabic{table}}
\renewcommand{\theequation}{A.\arabic{equation}}
\setcounter{figure}{0}
\setcounter{table}{0}
\setcounter{equation}{0}

\section{Implementation Details}
\label{app:implementation}
\subsection{Benchmark Construction Details}
\label{app:benchmark_construction_new}
\subsubsection{Benchmark Construction Details}
\label{app:benchmark_construction}

Our construction pipeline combines data curation with a multi-model pipeline. Figure~\ref{fig:pipeline} illustrates this workflow.

\textbf{1. Candidate Curation Thresholds:} Before semantic annotation begins, candidate videos are filtered by technical quality, audio availability, and predefined duration buckets ($<1$ min, $1$--$3$ min, $3$--$5$ min). We additionally enforce minimum structural density thresholds, such as minimum shots, events, and characters per minute, so that selected videos contain enough audio-visual structure for deep-structured evaluation. The exact thresholds are detailed in Table~\ref{tab:curation_thresholds}.

\textbf{2. Multi-Model Pipeline:} We decompose the annotation into a multi-role pipeline:
\begin{itemize}
    \item \textbf{Generator:} Initially drafts candidate JSON for the current track, for instance, all Reference entities.
    \item \textbf{Format Verification:} A deterministic script verifies required keys, ID uniqueness, temporal bounds, cross-links, and support fields. Malformed drafts are immediately rejected and returned with structured error traces for regeneration.
    \item \textbf{Evaluators:} Independent frontier models review the drafted facts for factual accuracy, temporal consistency, and structural coherence against the raw video.
    \item \textbf{Refiners:} Accepted drafts are passed to secondary models to locally refine and enrich descriptive fields, for instance, expanding the visual details of an \texttt{appearance\_anchor}. Crucially, this refinement occurs under the strict condition that the established structure remains invariant; a programmatic validator ensures all IDs, timestamps, and cross-links are unmodified. If a textual refinement breaks a structural link, the pipeline falls back to the original verified draft.
    \item \textbf{Selector:} Acts as the final judge, selecting the most accurate verified version without permission to rewrite any fields.
\end{itemize}
This separation of roles maintains structural consistency. Candidate states remain mere ``drafts'' until they pass all Format Verifications and consensus checks, culminating in human review.

\textbf{3. Annotation-Only Timestamp Overlays:} Throughout reference construction, the frames presented to the pipeline models and to human reviewers carry a red timestamp burned into the bottom-right corner. This overlay lets annotators and Evaluators anchor shot and event boundaries to an exact time index, which is what makes the audited \texttt{time\_range} fields frame-accurate. The overlay is strictly an annotation aid confined to the construction and review interfaces: all evaluation inputs are re-rendered from the original videos, so no evaluated model is ever shown a burned-in timestamp, and no ablation over overlay removal applies to the reported temporal-grounding numbers (Appendix~\ref{app:implementation_protocols}).

\begin{table}[htbp]
\centering
\small
\caption{\textbf{\name{} Task Suite.} All diagnostic tasks derive from the same frozen system of atomic units \(\mathcal{S}^\star(V)\), but each exposes a distinct model failure mode.}
\begin{tabular}{llll}
\toprule
\textbf{Task} & \textbf{Input} & \textbf{Output} & \textbf{Diagnostic Target} \\
\midrule
Generation & Raw Video \(V\) & Scored structured units & Native recovery and precision \\
Probing & Predicted JSON \(\hat{\mathcal{S}}(V)\) & Probe pass/fail & Internal structural validity \\
Verification & \(V\) + Candidate JSON unit & Accept/reject boolean & Local contradiction detection \\
\bottomrule
\end{tabular}
\label{tab:task_suite_overview}
\end{table}

\begin{table}[htbp]
\centering
\caption{\textbf{Candidate curation thresholds.} Video-level filters applied before reference-unit construction. Per-minute intervals are inclusive and follow $[\text{center}-\text{std}, 3\times\text{center}]$; absolute thresholds apply only to the listed duration buckets.}
\begin{tabular}{llc}
\toprule
\textbf{Bucket} & \textbf{Metric} & \textbf{Required Range} \\
\midrule
All & shots/min & $[5,30]$ \\
All & events/min & $[2,15]$ \\
$<1$ min & subjects/min & $[3,12]$ \\
$<1$ min & scenes/min & $[3,15]$ \\
$<1$ min & total shots & $\ge 4$ \\
$<1$ min & total events & $\ge 2$ \\
$1$--$3$ min & subjects/min & $[2,9]$ \\
$1$--$3$ min & scenes/min & $[2,9]$ \\
$3$--$5$ min & main characters & $\ge 3$ \\
$3$--$5$ min & main scenes & $\ge 3$ \\
\bottomrule
\end{tabular}
\label{tab:curation_thresholds}
\end{table}

\subsubsection{Output Schema}
\label{app:schema_spec}

To evaluate models natively, we instruct them to generate outputs strictly conforming to the \name{} JSON schema. This schema explicitly defines the \texttt{references}, \texttt{events}, and \texttt{shots} domains, alongside their mandatory structural links and support fields. Table~\ref{tab:schema_fields} summarizes these required components. If a model fails to adhere to these structural instructions, such as outputting a continuous paragraph instead of a parseable JSON graph, it triggers a parse failure at the SGC gate and receives a score of 0 for all downstream metrics.

During inference, all generative models operate at a temperature of \textbf{0.0} (greedy decoding) to minimize structural hallucinations, with a maximum output limit of 4096 tokens. The detailed instruction templates used in our pipeline are provided at the end of this appendix.

\begin{table}[htbp]
\centering
\caption{\textbf{Schema Field Definitions.} Descriptions and typing rules for every structural component required from the evaluated model.}
\resizebox{\linewidth}{!}{
\begin{tabular}{lll}
\toprule
\textbf{Component} & \textbf{Fields} & \textbf{Description} \\
\midrule
\texttt{references} & \texttt{ref\_id}, \texttt{type}, \texttt{semantic\_description}, \texttt{appearance\_anchor} & Persistent entities or scenes with detailed visual descriptions under \texttt{id\_features} \\
\texttt{events} & \texttt{event\_id}, \texttt{type}, \texttt{time\_range}, \texttt{content} & Timed audio-centered units; spoken dialogue requires exact \texttt{content.line} \\
\texttt{shots} & \texttt{shot\_id}, \texttt{time\_range}, \texttt{shot\_transfer}, \texttt{visual\_description}, \texttt{camera} & Visual timeline segments containing fine-grained sub-shot units \\
Link fields & \texttt{active\_events}, \texttt{references\_in\_shot}, reference IDs in \texttt{content} & Checkable structural links governing event-shot membership and reference context \\
Support fields & \texttt{appearance\_anchor}, \texttt{visual\_description}, \texttt{time\_range} & Verifiable evidence fields utilized for local semantic diagnostics \\
\bottomrule
\end{tabular}}
\label{tab:schema_fields}
\end{table}

\subsubsection{Generation and Evaluation Protocols}
\label{app:implementation_protocols}

This section details the implementation mechanics of \name{}, including how models are invoked for structured generation and how LLM-assisted matching and judging are executed in code.

\paragraph{API Invocation and Fallback Mechanics}
Because \name{} strictly requires the generation of complex, interdependent JSON schemas, interacting with diverse model APIs requires robust programmatic handling. As implemented in our evaluation engine:
\begin{itemize}[leftmargin=*]
    \item \textbf{Modality Inputs:} Proprietary models, such as the Gemini family, receive the video via native API upload endpoints or HTTP URLs, while open-source checkpoints, including the Qwen-Omni series, receive interleaved Base64-encoded visual frames and raw audio waveforms alongside the task instruction.
    \item \textbf{Retry and Repair Logic:} The pipeline automatically strips hallucinated markdown formatting, for instance, \texttt{```json ... ```} delimiters. If the output remains structurally malformed, the system triggers up to 5 exponential-backoff retries. If the model consistently fails to produce a valid JSON schema, the output is permanently logged as a parse failure, and all downstream structural metrics for that sample evaluate to zero.
\end{itemize}

\paragraph{Native Track Generation Protocol}
To generate the required atomic units, models are guided by a comprehensive schema definition. Depending on the model's context window and reasoning capacity, the inference engine employs one of two strategies:
\begin{itemize}[leftmargin=*]
    \item \textbf{Single-Pass Generation:} The model receives the full video alongside a unified system instruction and is required to output a single, complete JSON object containing the \texttt{references}, \texttt{events}, and \texttt{shots} arrays. This approach tests the model's ability to globally reason over the entire narrative.
    \item \textbf{Progressive Pipeline:} For extremely long videos or models with limited context windows, generation is decoupled. First, the model processes the entire video to extract persistent entities. Then, it processes the video in sliding temporal segments to generate temporally grounded \texttt{events} and \texttt{shots}, actively referencing the global entities defined in the first step.
\end{itemize}

The generation instructions serve as the strict standard operating procedure for the models. Rather than allowing the model to freely interpret the task, we embed explicit annotation guidelines directly into the system instruction. To ensure models accurately ground their predictions, the protocol enforces several critical principles:

\begin{itemize}[leftmargin=*]
    \item \textbf{Timestamp Convention:} Models must report every temporal field as a \texttt{[start, end]} range in seconds relative to the beginning of the video, and are instructed to keep all boundaries inside the video duration; out-of-range predictions are clamped before scoring (Appendix~\ref{app:structural_alignment}). Evaluated models receive the original, unaltered frames: no timestamp is burned into the pixels and no per-frame time index is injected into the prompt, so every temporal boundary must be inferred from the video stream itself (Listing B.1). The red timestamp overlays used during reference construction (Appendix~\ref{app:benchmark_construction}) are an annotation aid only and never appear in any model-facing input.
    \item \textbf{Mute Video Handling:} If a video contains no audio track, the model is strictly forbidden from inferring auditory events based on visual cues; for instance, a person moving their mouth must not trigger a hallucinated dialogue event unless audio is actually present.
    \item \textbf{Minimum Granularity Principle (Visual Subshots):} Models are instructed to decompose visual narratives into the smallest indivisible atomic actions. For instance, rather than generating a compound description like \textit{``picked up the cup, drank, and put it down,''} the model is forced to split this into three distinct subshots.
    \item \textbf{Temporal Overlap Principle:} Crucially, models are explicitly encouraged to generate overlapping temporal boundaries for subshots. If Person A is walking while Person B is nodding, these are treated as concurrent independent streams rather than being artificially serialized.
    \item \textbf{Strict Objectivity:} Models must describe pure visual and auditory facts. We explicitly ban causal connective inferences, for instance, \textit{``The vase broke because the ball hit it.''}, in favor of sequential factual statements.
\end{itemize}

\subsubsection{Dataset and Domain Statistics}
\label{app:dataset_statistics}

\name{} evaluates models across diverse content domains and temporal scales. We report detailed dataset statistics not to claim unrestricted universal coverage, but to explicitly define the boundaries of our evaluation scope. As detailed in Table~\ref{tab:full_dataset_stats}, the benchmark comprises 786 videos with verified structural averages. We break down this diversity across multiple structural and semantic dimensions to ensure our testbed evaluates joint audio-visual comprehension robustly.

\paragraph{Data Source and Category Distributions}

\begin{figure*}[htbp]
    \centering
    \begin{subfigure}[t]{0.48\textwidth}
        \centering
        \includegraphics[width=\linewidth]{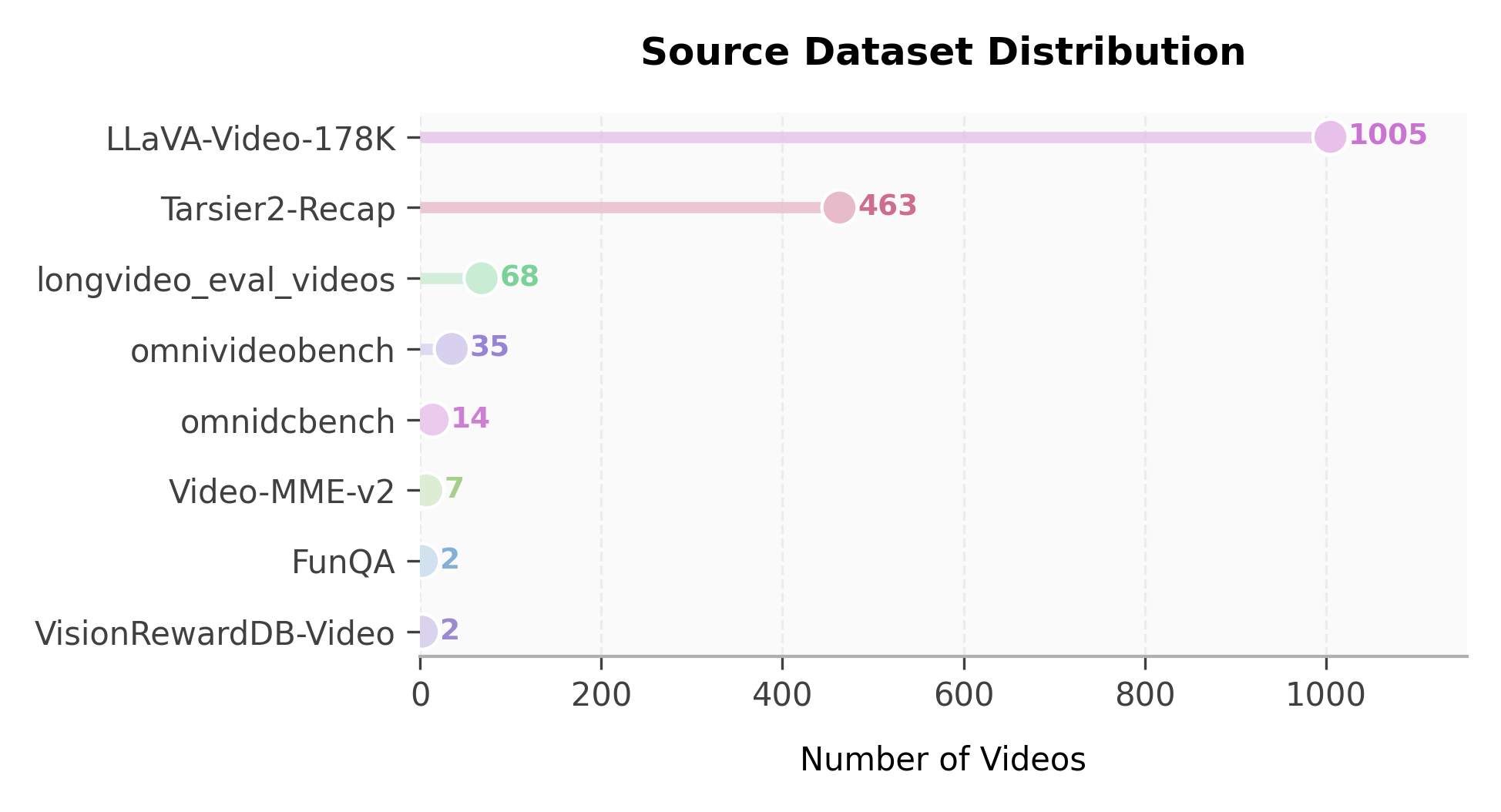}
        \caption{\textbf{Source Dataset Distribution.} A lollipop chart detailing the origins of the videos in the \name{} benchmark. The diverse sources ensure models are tested against varying production styles and visual characteristics.}
        \label{fig:app_datasets}
    \end{subfigure}\hfill
    \begin{subfigure}[t]{0.48\textwidth}
        \centering
        \includegraphics[width=\linewidth]{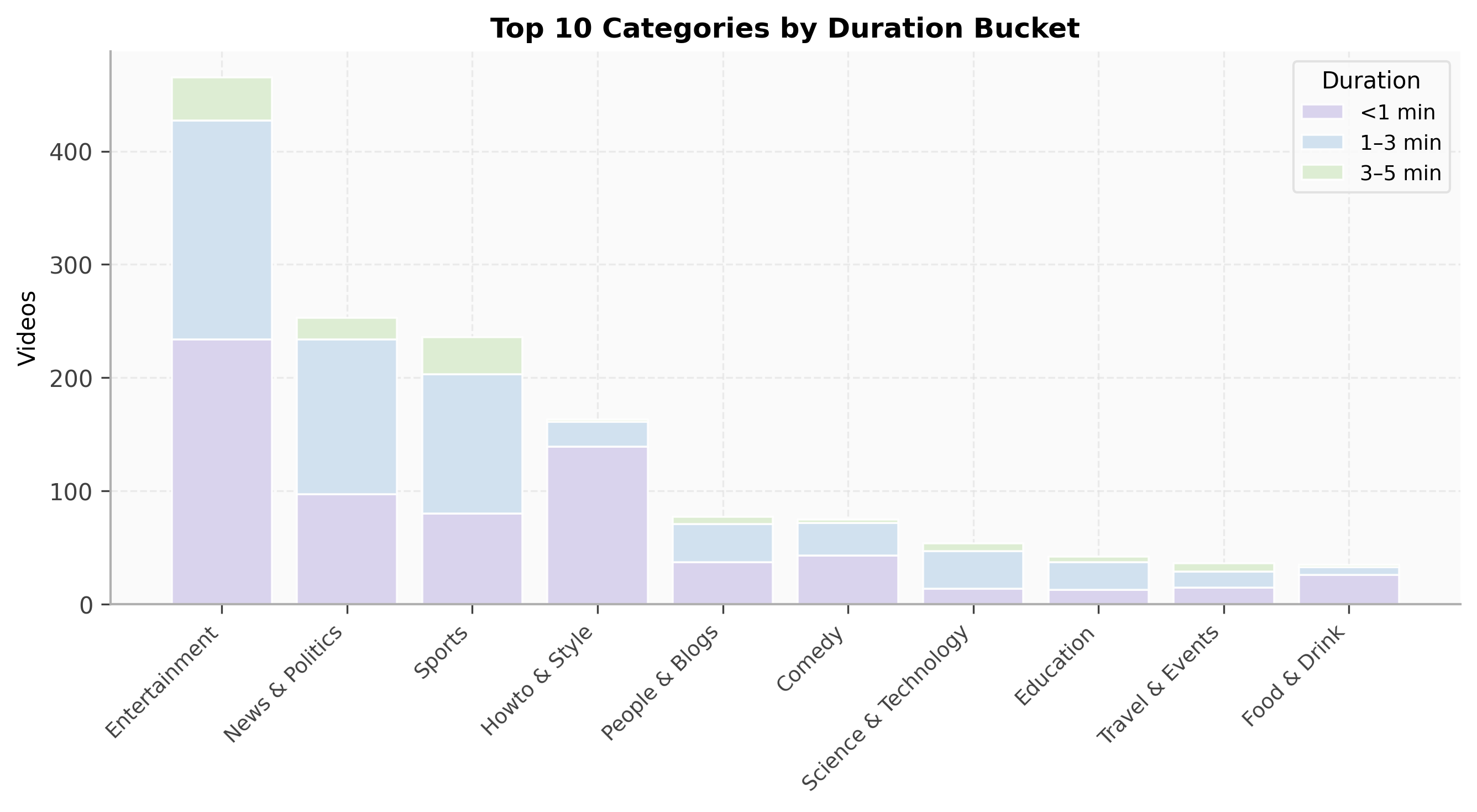}
        \caption{\textbf{Categories by Duration Bucket.} A stacked bar chart showing the composition of video durations, including $<1$ min, $1-3$ min, and $3-5$ min, across the top 10 semantic categories.}
        \label{fig:app_category_duration}
    \end{subfigure}
    \caption{\textbf{Data Source and Category Distributions.} The left panel illustrates the distribution of source datasets, ensuring domain diversity. The right panel shows the composition of video durations across the top 10 semantic categories.}
    \label{fig:app_datasets_and_categories}
\end{figure*}

To ensure domain diversity, the videos in \name{} are curated from multiple public datasets. Figure~\ref{fig:app_datasets_and_categories}(a) illustrates the distribution of source datasets. Beyond the source datasets, we analyze the semantic categories of the videos. Figure~\ref{fig:app_datasets_and_categories}(b) demonstrates the temporal footprint of these categories. Notably, categories such as "Entertainment" and "News \& Politics" exhibit distinct duration distributions, highlighting the varying temporal demands placed on models when processing different content types.

\begin{table}[htbp]
\centering
\caption{\textbf{Full Dataset Statistics.} Comprehensive macro averages and distribution metrics supplementing Table~\ref{tab:keystats}. These statistics explicitly define the coverage and complexity of the \name{} evaluation scope.}
\small
\begin{tabular}{lc}
\toprule
\textbf{Metric} & \textbf{Value} \\
\midrule
\multicolumn{2}{l}{\textit{Global Counts and Averages}} \\
Total Videos & 786 \\
Mean Duration (seconds) & 58.82 \\
Median Duration (seconds) & 28.32 \\
Avg. References / Video & 7.40 \\
Avg. Events / Video & 8.32 \\
Avg. Shots / Video & 14.53 \\
\midrule
\multicolumn{2}{l}{\textit{Video-Level Distributions}} \\
Train / Dev / Test Split & None (Test-only benchmark) \\
Audio Presence Rate (Videos with active audio) & 100.00\% \\
\bottomrule
\end{tabular}
\label{tab:full_dataset_stats}
\end{table}

\paragraph{Temporal and Density Characteristics}

\begin{figure*}[htbp]
    \centering
    \begin{subfigure}[t]{0.48\textwidth}
        \centering
        \includegraphics[width=\linewidth]{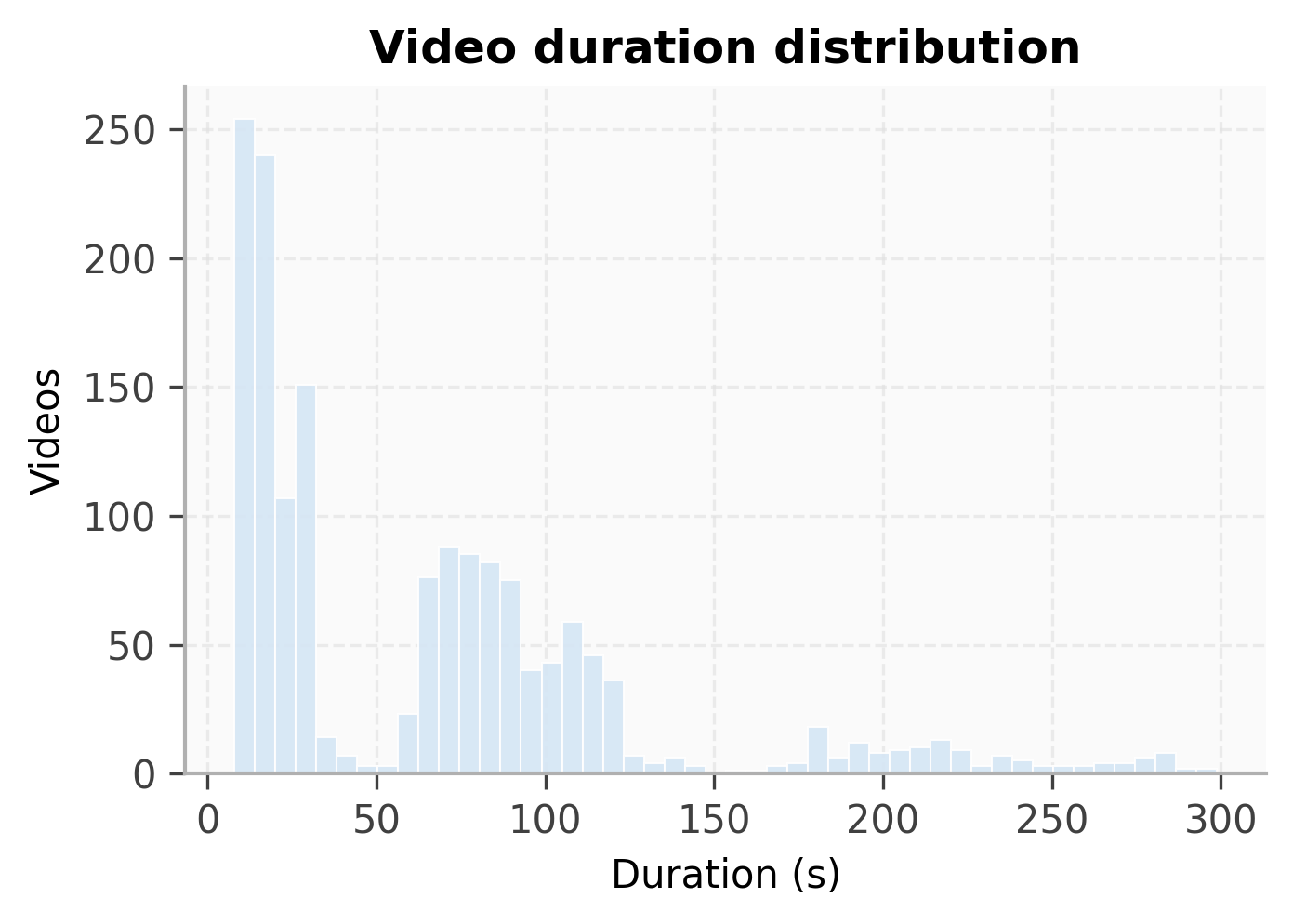}
        \caption{\textbf{Duration Distribution (Seconds).} The overall temporal footprint of the benchmark videos, bounded strictly within 5 minutes (300 seconds) to enable rigorous, exact analysis.}
        \label{fig:app_temporal}
    \end{subfigure}\hfill
    \begin{subfigure}[t]{0.48\textwidth}
        \centering
        \includegraphics[width=\linewidth]{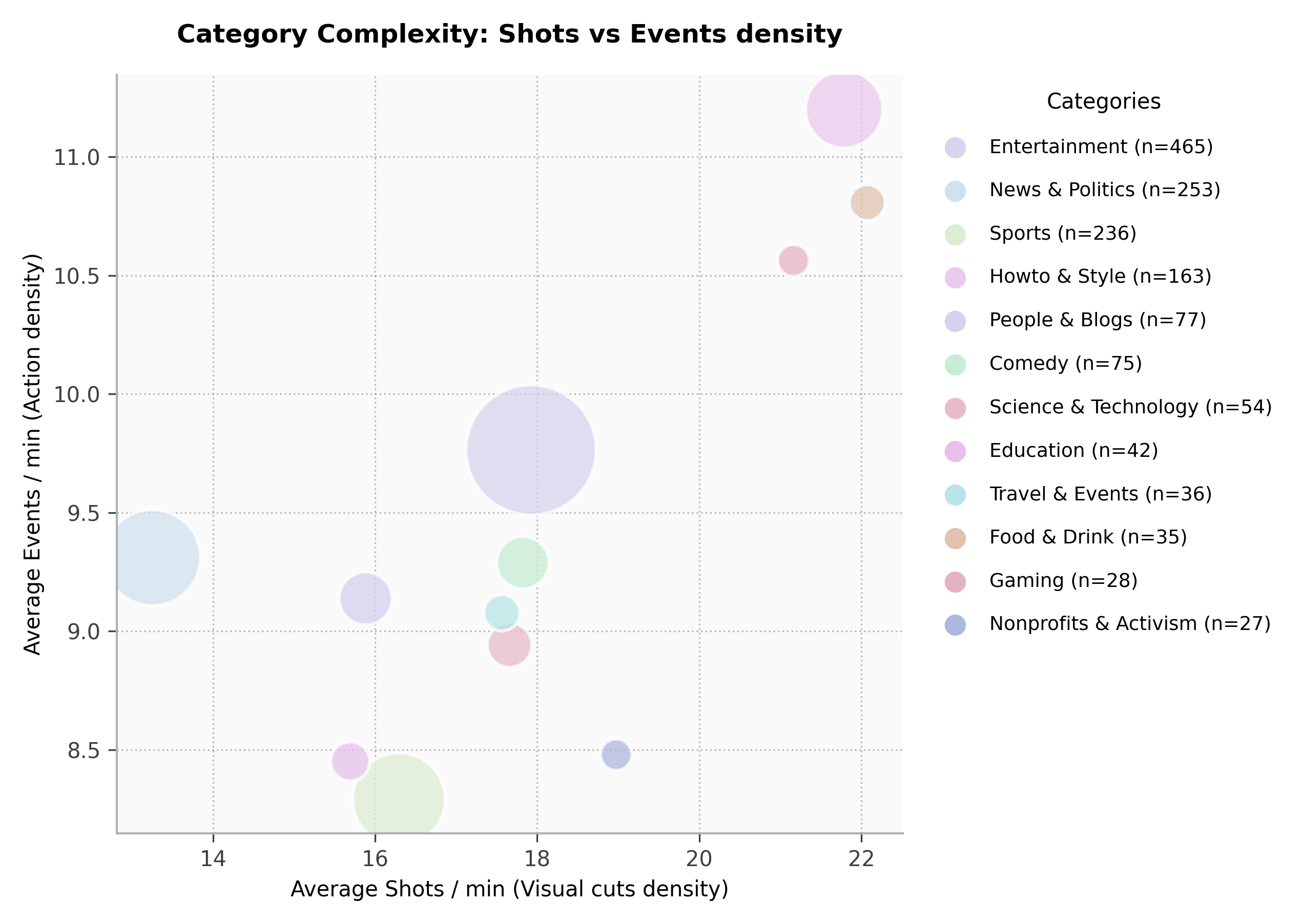}
        \caption{\textbf{Category Complexity.} A bubble chart correlating the visual cuts density (shots/min) with action density (events/min) across categories. Bubble sizes correspond to the number of videos.}
        \label{fig:app_bubble}
    \end{subfigure}
    \caption{\textbf{Temporal Footprint and Structural Complexity.} The left panel provides an exhaustive view of the video duration distribution. The right panel contextualizes the structural complexity by plotting visual cuts density against action density for different categories.}
    \label{fig:app_temporal_and_bubble}
\end{figure*}

\begin{figure*}[htbp]
    \centering
    \begin{subfigure}{0.48\textwidth}
        \centering
        \includegraphics[width=\linewidth]{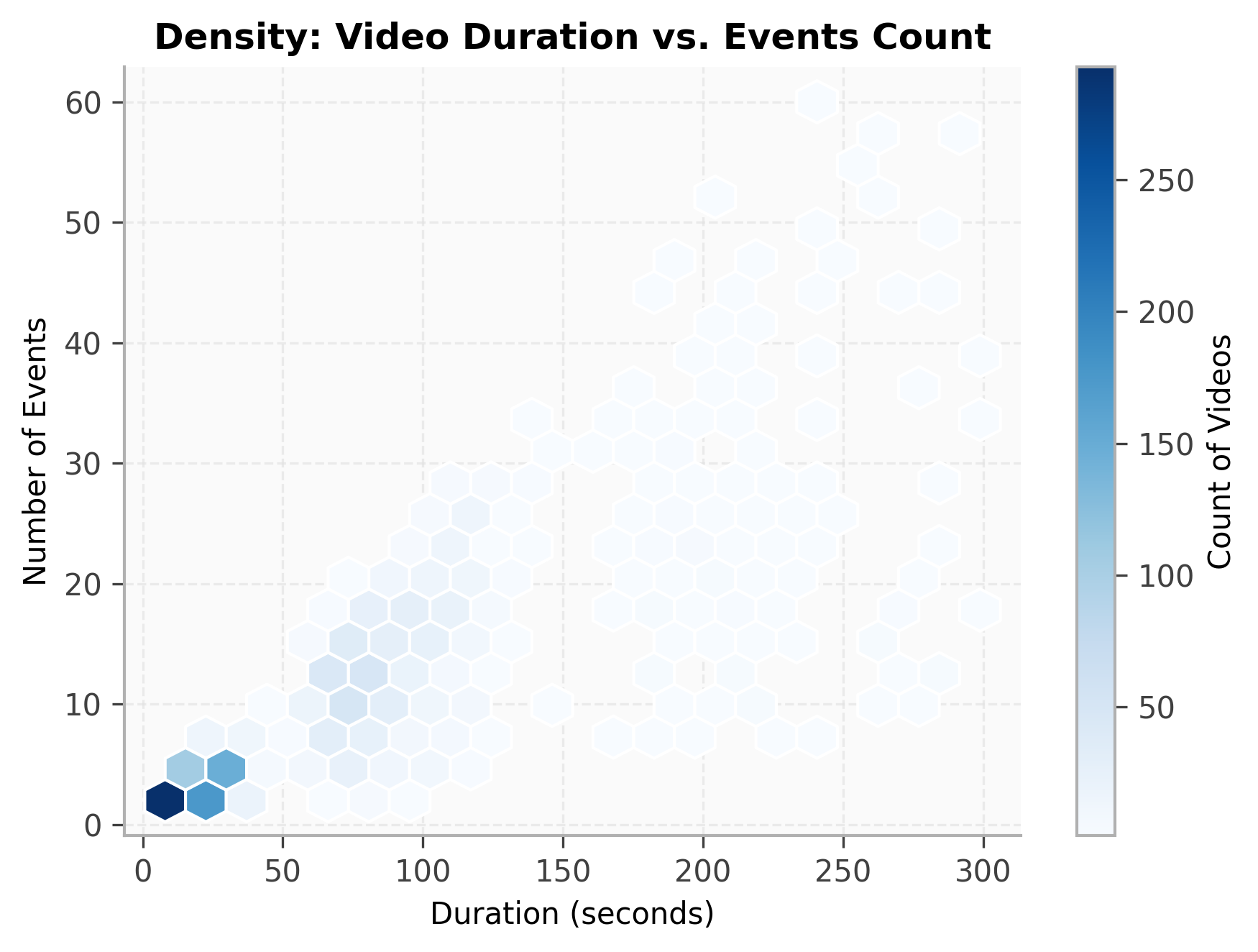}
        \caption{\textbf{Video Duration vs. Events Count.} A hexagonal binning plot displaying the correlation between the length of the video and the number of annotated audio-visual events. Darker bins represent higher concentrations of videos.}
        \label{fig:app_hexbin}
    \end{subfigure}\hfill
    \begin{subfigure}{0.48\textwidth}
        \centering
        \includegraphics[width=\linewidth]{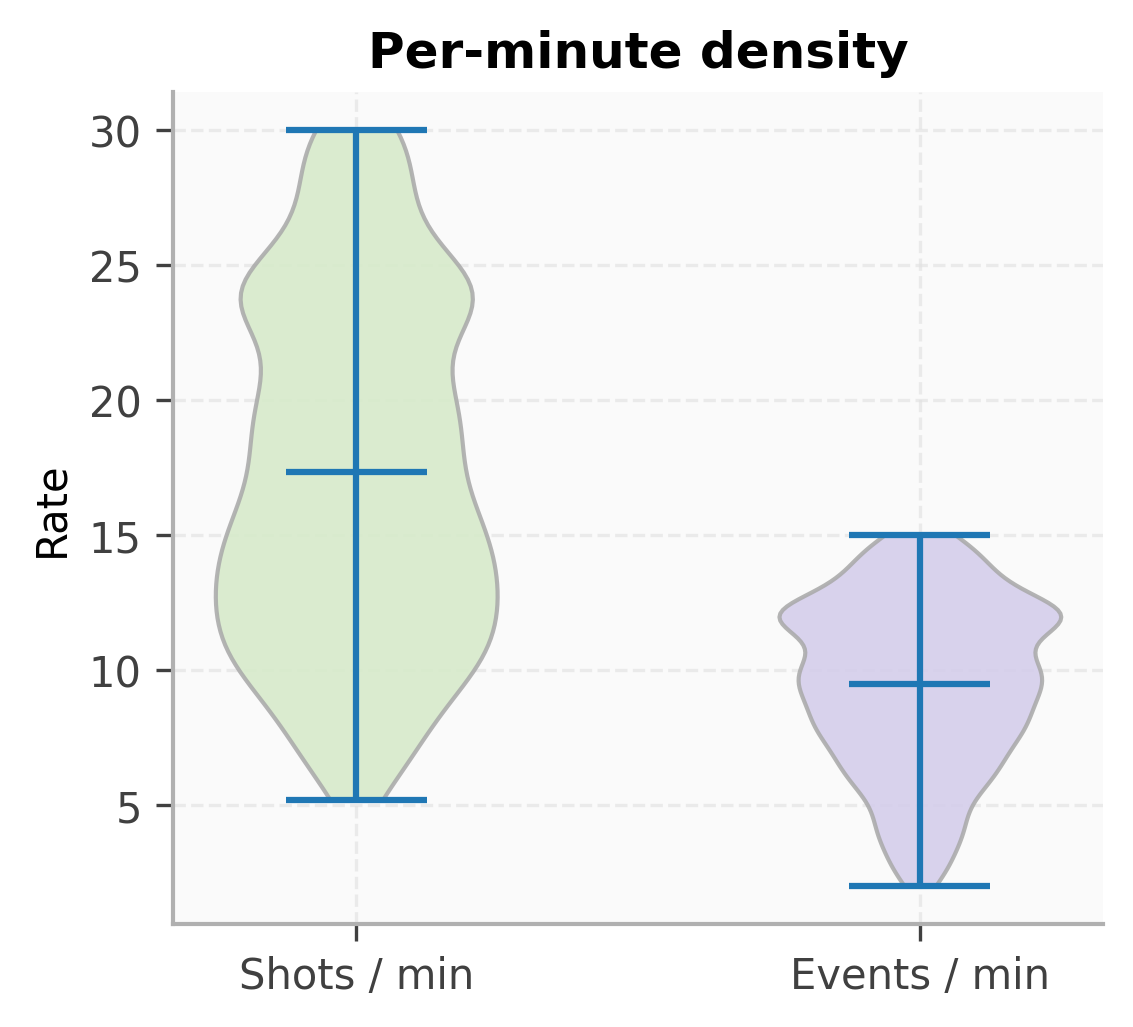}
        \caption{\textbf{Per-Minute Density.} Violin plots of the shots per minute and events per minute, visualizing the rapid pace of visual and narrative changes across the dataset.}
        \label{fig:app_rates}
    \end{subfigure}
    \caption{\textbf{Temporal Scale and Per-Minute Structural Density.} The left panel relates video duration to event count; the right panel summarizes shots-per-minute and events-per-minute distributions. These two views separate absolute temporal scale from normalized structural density.}
    \label{fig:app_hexbin_rates}
\end{figure*}

A core contribution of \name{} is its fine-grained structural evaluation, which necessitates characterizing how densely events and shots are annotated across videos. To contextualize this annotation regime, Figure~\ref{fig:app_temporal_and_bubble}(a) provides an exhaustive view of the benchmark's temporal distribution, spanning short clips to videos extending up to 5 minutes. Figure~\ref{fig:app_hexbin_rates} then offers a complementary perspective by separating two related but distinct properties: the absolute relationship between video duration and event count, and the normalized per-minute density of shots and events.

\paragraph{Structural Complexity Metrics}

\begin{figure*}[htbp]
    \centering
    \includegraphics[width=\textwidth]{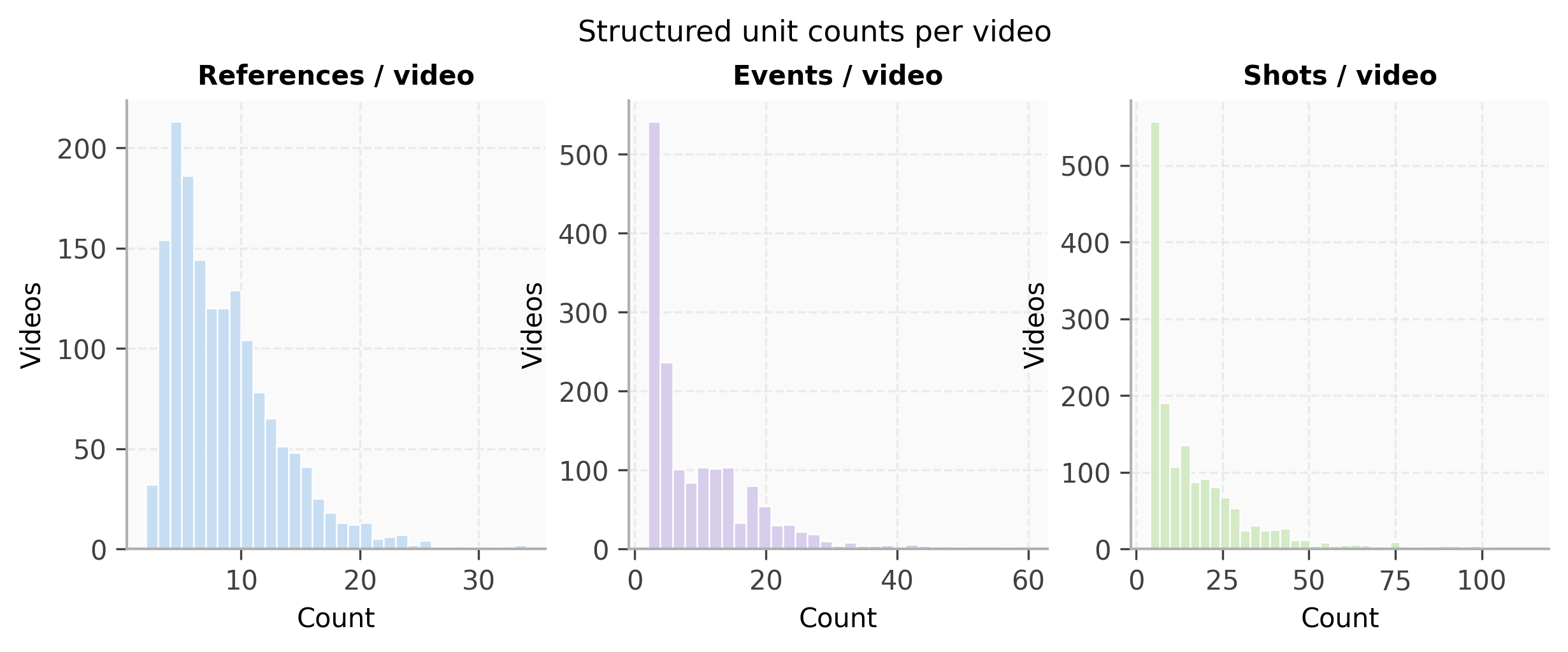}
    \caption{\textbf{Structured Unit Counts per Video.} Individual histograms of References, Events, and Shots per video, illustrating the dense tracking requirements placed on the models.}
    \label{fig:app_unit_hists}
\end{figure*}

We measure the complexity of \name{} by examining the rates of shots and events per minute, as well as the absolute counts of structural units per video. As shown in Figure~\ref{fig:app_unit_hists}, the unit counts form stable distributions, demanding models to manage multiple entities and events concurrently. Figure~\ref{fig:app_hexbin_rates} jointly presents the relationship between video duration and event count and the per-minute density of shots and events. Figure~\ref{fig:app_temporal_and_bubble}(b) further contextualizes this complexity by plotting the visual cuts density against action density for different categories.

\paragraph{Human Annotation and Quality Assurance Protocol}

To ensure the highest possible data quality for \name{}, we implement a rigorous, iterative human annotation and refinement protocol. Rather than simply calculating post-hoc agreement scores, our pipeline is designed around continuous error discovery and correction. This multi-stage quality assurance process guarantees that the final benchmark data is structurally flawless, semantically rich, and strictly faithful to the original video evidence.

\textbf{Professional Annotator Training.} We engaged a team of professional annotators with extensive experience in dense video captioning and complex multi-modal data processing. Prior to the formal annotation phase, all annotators underwent comprehensive training on the \name{} JSON schema and standardized corner cases. This training specifically addressed challenging scenarios such as heavily occluded characters, ambiguous shot boundaries, overlapping audio events, and fine-grained visual state transitions.

\textbf{Iterative Refinement Pipeline.} The core of our high-quality data generation is an iterative ``review-and-refine'' mechanism. Initial drafts, whether generated by annotators or derived from the multi-model pipeline, are subjected to meticulous human review using a synchronized video-audio playback interface. Reviewers scrutinize the drafts for:
\begin{itemize}[leftmargin=*]
    \item \emph{Structural Integrity:} Ensuring that all ID references (\texttt{ref\_id}, \texttt{event\_id}) are consistently maintained across temporal boundaries without hallucinated links.
    \item \emph{Temporal Precision:} Fine-tuning the start and end timestamps for visual shots and audio events down to the frame level.
    \item \emph{Semantic Granularity:} Expanding overly generic descriptions into highly specific factual statements, for instance, adding exact clothing details, specific character poses, or nuanced environmental audio cues.
\end{itemize}

Whenever an error, omission, or ambiguity is identified, the annotation is sent back for immediate refinement. This targeted correction loop is repeated until the JSON output perfectly reflects the video's ground truth.

\textbf{Multi-Stage Adjudication.} To resolve complex edge cases, we employ a multi-stage adjudication process. If annotators disagree on the interpretation of a scene, for example, determining whether a background noise constitutes a distinct event or how to segment a continuous camera movement, the case is escalated to senior reviewers. These domain experts finalize the annotations based strictly on the raw video and audio evidence. This protocol ensures that our evaluation references are not only format-compliant but set a gold standard for comprehensive, verifiable audio-visual understanding.

\subsection{Full Experimental Details}
\label{app:experimental_details}

\subsubsection{Experiments Compute Resources}
\label{app:compute_resources}
All of our experiments are conducted on compute nodes equipped with 8$\times$ NVIDIA H20 GPUs. For local model deployment and batched inference, we employ vLLM~\citep{kwon2023efficient} to optimize serving efficiency and memory usage.

\paragraph{Evaluation Matching Principles}
Before scoring, predicted units must be structurally aligned to ground-truth units (as formalized in Appendix~\ref{app:structural_alignment}). The programmatic matching principles are implemented as follows:
\begin{itemize}[leftmargin=*]
    \item \textbf{Reference Matching:} Performed via an LLM-assisted bipartite matcher. The matcher is restricted to comparing only the \texttt{appearance\_anchor.id\_features.detail\_description} of the entities. It strictly enforces a one-to-one match based on semantic equivalence, rejecting generic overlaps, empty descriptions, or contradictory identity/clothing facts.
    \item \textbf{Event Matching:} Handled deterministically. Dialogue events are matched using normalized Word Error Rate (WER) against the transcribed text (threshold $\ge 0.50$). Non-dialogue events, including sound effects and background music, are matched purely based on their temporal Intersection-over-Union (tIoU) (threshold $\ge 0.20$).
    \item \textbf{Shot and Subshot Matching:} Visual shots are matched one-to-one via temporal IoU (threshold $\ge 0.30$). Subshots are then matched locally within their aligned parent shots using an expanded time window ($\delta=1.0\text{s}$).
\end{itemize}

\paragraph{LLM-as-Judge Protocol}
As established in Section~\ref{subsec:eval_goals}, \name{} uses LLMs exclusively for bounded local semantic checks, avoiding opaque holistic evaluations. The following examples summarize the fixed local judging criteria used in our pipeline.

By injecting the exact matching scope into the instruction templates and decoupling the evaluation of different modalities, \name{} mathematically guarantees that the final diagnostic scores reflect specific capability deficits rather than arbitrary stylistic penalties.

\subsubsection{Metric Definitions}
\label{app:metrics_definitions}

This appendix formally defines the \name{} evaluation suite. Our core evaluation philosophy shifts video benchmarking from \emph{weak, holistic definitions} (where an entire paragraph is judged opaquely) to \emph{strong, structural definitions}. We define individual information points, including characters, sound events, and visual actions, as \textbf{verifiable atomic units}, whose interactions are strictly verified by deterministic rules before localized LLM scoring. 

\paragraph{Evaluation Protocol}
\label{app:bidirectional_matching}

\textbf{Why bidirectional evaluation is necessary.}
Unlike multiple-choice or short-answer tasks, audio-visual captioning is fundamentally open-ended: there is no single ``correct'' output string. When humans and models caption the exact same video, their structured representations rarely form a perfect one-to-one mapping. Three recurring sources of mismatch motivate our design:

\begin{itemize}[leftmargin=*]
    \item \emph{Granularity mismatch:} A model might decompose a continuous 5-second visual action into three dense, fine-grained subshots, whereas the ground truth summarizes the same sequence into a single cohesive subshot. Both decompositions can be factually correct.
    \item \emph{Attribute selection divergence:} A model might describe a character by clothing and posture, while the human annotator emphasizes hairstyle and facial features; both descriptions are valid, yet they share few lexical tokens.
    \item \emph{Structural topology variation:} Models may group audio events under different shots than humans, or split a single reference entity into multiple aliases, creating non-trivial alignment challenges even when the underlying facts are equivalent.
\end{itemize}

\textbf{The problem with unidirectional evaluation.}
Because of these mismatches, evaluating open-ended captioning from a single direction is intrinsically biased and gameable:

\begin{itemize}[leftmargin=*]
    \item \emph{Recall-only evaluation} (checking whether the prediction covers the ground truth) rewards excessively verbose, speculative outputs that guarantee coverage at the cost of significant hallucinations.
    \item \emph{Precision-only evaluation} (checking whether the ground truth supports the prediction) rewards overly terse, generic responses that are trivially correct but miss key video details.
\end{itemize}

\textbf{Our solution: bidirectional matching and scoring.}
To address this, \name{} applies a \textbf{bidirectional matching and scoring paradigm} specifically for open-vocabulary and continuous fields where the above mismatches are most pronounced, namely References (Subject and Scene appearances) and visual Subshots. We evaluate these elements from two complementary perspectives, both during the structural matching phase (determining \emph{which} predicted units correspond to which ground-truth units) and the subsequent local semantic scoring phase (measuring \emph{how well} a matched pair agrees on factual content).

\textbf{Tie-Breaking Rules:} For rule-based event and shot matching, candidate pairs are sorted by their matching score and consumed one-to-one. In instances where multiple predicted units achieve the exact same score against a ground-truth unit, we prioritize the prediction with the earlier temporal start time ($t_s$). If a tie persists, we break it using the lexical order of the predicted ID string. Reference matching is parsed from the LLM-produced bipartite map and then defensively checked for valid IDs, type compatibility, and one-to-one consistency.

Table~\ref{tab:matching_rules} details the matching rules for each unit layer.

\begin{itemize}
    \item \textbf{Reference Matching:} Typed one-to-one LLM-assisted matching over \texttt{appearance\_anchor.id\_features.detail\_description}. The matcher only considers \texttt{person}, \texttt{scene}, and \texttt{object} groups and does not use semantic summaries, attributes, timestamps, event links, shot links, or broader context.
    \item \textbf{Event Matching:} Typed one-to-one matching. Dialogue events are matched using normalized WER over \texttt{content.line} with threshold \(0.50\); non-dialogue events use temporal Intersection-over-Union (tIoU) with threshold \(0.20\).
    \item \textbf{Shot and Subshot Matching:} Shots are matched one-to-one using temporal IoU with threshold \(0.30\). Subshots are matched locally inside matched shot pairs: a ground-truth subshot creates a window expanded by \(\delta=1.0\text{s}\), and overlapping predicted subshots within the matched shot form its candidate group.
\end{itemize}

\begin{table}[htbp]
\centering
\caption{\textbf{Matching Rules Summary.} Parameters defining structural alignment across the reference, event, and shot layers.}
\resizebox{\linewidth}{!}{
\begin{tabular}{lllc}
\toprule
\textbf{Layer} & \textbf{Matching Metric} & \textbf{Default threshold} & \textbf{Tie-breaking} \\
\midrule
Reference & LLM bipartite matching over \texttt{detail\_description} & None (LLM output) & Defensive ID/type/one-to-one check \\
Dialogue event & $1-\mathrm{WER}$ over \texttt{content.line} & $\ge 0.50$ & Max line score, one-to-one \\
Non-dialogue event & Temporal IoU & $\ge 0.20$ & Max tIoU, one-to-one \\
Shot & Temporal IoU & $\ge 0.30$ & Max tIoU, one-to-one \\
\bottomrule
\end{tabular}}  
\label{tab:matching_rules}
\end{table}

Table~\ref{tab:open_world_policy} formally defines how matched, unmatched, invalid, and out-of-scope predictions contribute to credit, penalties, or diagnostics. This scope-bounded scoring policy is critical: it prevents penalizing models for discovering true facts that fall outside our annotation budget, while still strictly penalizing structural hallucinations.

\begin{table}[htbp]
\centering
\caption{\textbf{Open-World Scoring Policy.} This table outlines exactly how we credit, penalize, or ignore different types of predictions within our bounded evaluation scope. Crucially, while \name{} rewards models for recovering annotated truths, it does \emph{not} unfairly penalize models for correctly identifying true facts that merely fell outside our annotation budget. Unsupported extras are logged for diagnostic analysis but do not directly dock the main precision metrics.}
\small
\resizebox{\linewidth}{!}{
\begin{tabular}{p{3.1cm}cccp{2.8cm}p{2.4cm}}
\toprule
\textbf{Prediction type} & \textbf{Matched} & \textbf{Valid structure} & \textbf{Supported} & \textbf{Credit / penalty} & \textbf{Denominator} \\
\midrule
Matched unit & yes & yes & n/a & Full credit; proceeds to structural/semantic checks. & Matched metric denominators \\
Unmatched reference & no & yes/no & no & No credit; retained in diagnostic logs. & Appendix audit \\
Unmatched event & no & yes/no & no & No credit; retained in diagnostic logs. & Appendix audit \\
Unmatched sub-shot & no & yes/no & no & No credit; retained in diagnostic logs. & Appendix audit \\
Invalid or dangling link & no & no & no & Zero credit; penalized as a structural error. & Appendix audit \\
Valid but out-of-scope fact & no & yes & unknown & No manual credit, but no strict penalty; logged for auditing. & Reported diagnostics \\
\bottomrule
\end{tabular}}
\label{tab:open_world_policy}
\end{table}

\subsubsection{Full Model and Evaluation Settings}
\label{app:model_settings}

To ensure fair and transparent comparisons, this section explicitly documents the technical parameters used to evaluate each model. In total, we assessed 12 distinct models. To clarify the division between the main results and the robustness analysis:
\begin{itemize}
    \item \textbf{Main Benchmark Table:} Evaluates \textbf{7 native-capable omnimodal models} (three proprietary models via API and four open-source checkpoints). These models are capable of processing unstructured audio-visual streams and following instructions to generate structured outputs.
    \item \textbf{Appendix Robustness Table (Appendix~\ref{app:specialized_models_failure}):} Evaluates \textbf{5 specialized video-captioning baselines}. These models do not enter the main ranking because their SFT heavily degrades structured schema-following. We explicitly report these failures and their performance via the caption-posthoc route to demonstrate why free-form prose cannot substitute for native atomic unit prediction.
\end{itemize}

Because these models vary significantly in architecture, they operate under different hardware and context limitations. Table~\ref{tab:model_settings} meticulously units the full configuration settings required to reproduce our runs, including parameter counts, exact checkpoint hashes, frame sampling policies, max token limits, and decoding temperatures. We also explicitly log our evaluation pipeline policies, including retry limits and parse repair strategies. For open-source reproducibility, our released code artifacts replace internal endpoints and filesystem paths with sanitized placeholders.

\begin{table*}[!t]
\centering
\caption{\textbf{Model inference and configuration settings.} One row per evaluated model entry. All models use greedy decoding (temperature\,=\,0, top\_p\,=\,1) to minimize structural hallucination. Closed-source models are accessed via API with no frame cap; local Qwen3-Omni models (30B-A3B) are served via vLLM with TP\,=\,4 and max 192 frames at 2\,fps; videos exceeding 1\,min use a 2-pass strategy. Qwen2.5-Omni-based models (7B and fine-tuned variants) only evaluate videos $\leq$\,60\,s due to context limits.}
\label{tab:model_settings}
\scriptsize
\setlength{\tabcolsep}{3pt}
\resizebox{\linewidth}{!}{
\begin{tabular}{lp{2.6cm}ccccccc}
\toprule
\textbf{Model} & \textbf{Checkpoint / Endpoint} & \textbf{Params} & \textbf{Backend} & \textbf{Audio} & \textbf{Frames / FPS} & \textbf{Max Tokens} & \textbf{Duration} & \textbf{Long-Video Strategy} \\
\midrule
\rowcolor{limegreen!50} \multicolumn{9}{c}{\textbf{\textit{Proprietary Models}}} \\
\midrule
Gemini 3.1-Pro & gemini-3.1-pro-preview & -- & API & \cmark & no cap / 2\,fps & 65536 & All & Single-pass \\
Gemini 2.5-Pro & gemini-2.5-pro & -- & API & \cmark & no cap / 2\,fps & 65536 & All & Single-pass \\
Qwen3.5-Omni-Plus & qwen3.5-omni-plus & -- & API & \cmark & no cap / 2\,fps & 32768 & All & Single-pass \\
Qwen3.5-Omni-Flash & qwen3.5-omni-flash & -- & API & \cmark & no cap / 2\,fps & 32768 & All & Single-pass \\
\midrule
\rowcolor{grayred!40} \multicolumn{9}{c}{\textbf{\textit{Open-source Omnimodal Models}}} \\
\midrule
Qwen3-Omni-Instruct & Qwen3-Omni-30B-A3B-Instruct & 30B-A3B & vLLM (TP=4) & \cmark & 192 / 2\,fps & 32768 & All & 2-pass ($>$60\,s) \\
Qwen3-Omni-Captioner & Qwen3-Omni-30B-A3B-Captioner & 30B-A3B & vLLM (TP=4) & \cmark & 192 / 2\,fps & 32768 & All & 2-pass ($>$60\,s) \\
\midrule
\rowcolor{lightgray} \multicolumn{9}{c}{\textbf{\textit{Specialized Video-Captioning Models (Appendix~\ref{app:specialized_models_failure})}}} \\
\midrule
ASID-Caption-3B & ASID-Captioner-3B & 3B & Transformers & \cmark & 192 / 2\,fps & 32768 & $\leq$60\,s & -- \\
ASID-Caption-7B & ASID-Captioner-7B & 7B & Transformers & \cmark & 192 / 2\,fps & 32768 & $\leq$60\,s & -- \\
AVoCaDO & AVoCaDO & 7B & Transformers & \cmark & 192 / 2\,fps & 32768 & $\leq$60\,s & -- \\
TimeChat-Captioner & TimeChat-Captioner-GRPO-7B & 7B & Transformers & \cmark & 192 / 2\,fps & 32768 & $\leq$60\,s & -- \\
UGC-VideoCaptioner & UGC-VideoCaptioner & 7B & Transformers & \cmark & 192 / 2\,fps & 32768 & $\leq$60\,s & -- \\
\bottomrule
\end{tabular}
}
\end{table*}

\paragraph{Error Analysis Details}
\label{app:error_analysis_details}

While our rule-based metrics quantify overall model performance, diagnosing specific failure modes requires a deeper analysis of the error distribution. Table~\ref{tab:error_analysis} outlines the precise audit contract behind the error-composition chart (Figure~\ref{fig:error_analysis_summary}) in the main text. It defines exactly what constitutes a failure under each metric family, ensuring every error is interpretable as a concrete, broken structural link rather than a subjective LLM penalty.

\begin{table*}[!t]
\centering
\caption{\textbf{Error analysis summary.} This table defines exactly how we classify failures in structurally valid outputs. It explicitly links each failure mode to the corresponding rule-based metric. Counts and rates are derived from our evaluation logs over applicable samples for the Gemini 3.1-Pro model. This detailed breakdown ensures that every capability deficit is independently traceable rather than being obscured by a single aggregate score.}
\label{tab:error_analysis}
\small
\setlength{\tabcolsep}{3pt}
\resizebox{\linewidth}{!}{
\begin{tabular}{llp{4.2cm}cc}
\toprule
\textbf{Error Family (Category)} & \textbf{Primary Metric} & \textbf{What triggers this error?} & \textbf{Count} & \textbf{Rate} \\
\midrule
Schema Failure (Gate) & SGC & Output cannot be parsed into references, events, and shots, or is missing required JSON fields & 31 & 3.8\% \\
Reference Recovery (Visual) & Ref F1 & Required ground-truth reference is missed, or a hallucinated predicted reference cannot be matched & 3778 & 31.7\% \\
Reference-Use (Visual) & RefUse & A matched shot or subshot cites the wrong persistent reference ID & 8649 & 34.1\% \\
Identity Drift (Visual) & CCC & A single persistent reference is split, merged, or inconsistently tracked across multiple shots & 2656 & 65.2\% \\
Shot Recovery/Alignment (Visual) & Shot F1/tIoU & A visual micro-action or boundary is missed, hallucinated, or severely misaligned in time & 5156 & 26.2\% \\
Event Recovery/Alignment (Audio) & Event F1/tIoU & A required audio event is missed, hallucinated, or aligned to a completely wrong time segment & 4218 & 36.2\% \\
Event-Shot Link (Audio-Visual) & EVSA & A predicted audio event is attached to the completely wrong visual shot in the timeline & 1654 & 44.5\% \\
Dialogue Error (Audio-Visual) & Speaker F1 & A matched dialogue event fails to accurately transcribe the spoken text or assign it to the correct visual speaker & 260 & 7.0\% \\
\bottomrule
\end{tabular}}
\end{table*}

\noindent \textbf{How to read the diagnostic categories.}
\begin{itemize}
    \item \textbf{Schema or canonicalization failure (Gate):} The model completely failed instruction following. The output cannot be converted into the scored unit system, rendering downstream semantic checks mathematically impossible.
    \item \textbf{Visual reasoning failures (Visual):} The model misses persistent entities, hallucinates non-existent objects, uses the wrong mapped ID when referring to an object in a scene, fails to maintain the same identity for a character across disconnected scenes, or misaligns visual boundaries.
    \item \textbf{Audio reasoning failures (Audio):} Essential audio events are entirely missing, hallucinated, or aligned to the wrong time segment in the audio track.
    \item \textbf{Cross-modal and dialogue failures (Audio-Visual):} An event may be plausible in isolation but is attached to the wrong visual region (demonstrating a failure in cross-modal structural reasoning), or spoken dialogue is mis-transcribed or assigned to the wrong speaker.
\end{itemize}
These structured categories explain precisely why a fluent, holistic caption can appear superficially correct globally while failing rigorously on localized evaluations.

\begin{table}[htbp]
\centering
\caption{\textbf{Failure taxonomy for outputs exhibiting high text-level scores but low structural scores.} This table dissects exactly why a seemingly fluent holistic prose caption can fail rigorous deep-structured evaluation. We map specific failure modes to the exact metric designed to penalize them. All rates are derived from our validation logs for the Gemini 3.1-Pro model.}
\small
\resizebox{\linewidth}{!}{
\begin{tabular}{lllp{4.0cm}c}
\toprule
\textbf{Failure type} & \textbf{Category} & \textbf{Metric} & \textbf{What it looks like in practice} & \textbf{Rate} \\
\midrule
Broken identity persistence & Visual & CCC & The model correctly describes a character but assigns them a new ID, such as \texttt{PERSON\_2}, when they reappear in a later shot, losing track of their persistent identity. & 65.2\% \\
Wrong event reference & Visual & RefUse & An off-screen voice or a background action is attributed to the wrong character ID in the structured output. & 34.1\% \\
Event alignment error & Audio & Event tIoU & The model correctly identifies an event, like a dog barking, but aligns it to a completely wrong time segment where the dog is silent. & 42.0\% \\
Missing structural field & Gate & SGC & The JSON output is missing required fields like \texttt{time\_range} or \texttt{detail\_description}, rendering the unit untestable. & 3.8\% \\
Unsupported extra event & Audio & Event (Precision) & The model hallucinates an action or event that never actually occurred in the source video. & 36.2\% \\
\bottomrule
\end{tabular}}
\label{tab:failure_taxonomy}
\end{table}

\paragraph{Analysis of the Caption-State Gap.}
As revealed in Table~\ref{tab:failure_taxonomy}, evaluating models purely via dense text generation obscures severe structural failures. The highest failure rates occur in Identity Persistence (65.2\%) and Temporal Alignment (42.0\%). These numbers prove that while models can fluently generate nouns and verbs (yielding high text-level scores), they fundamentally struggle to maintain object permanence across shots or anchor events to precise timeline segments. Relying solely on holistic LLM judging masks these precise failure modes, confirming our motivation that reliable evaluation inherently requires structured, atomic verification.

\renewcommand{\thefigure}{B.\arabic{figure}}
\renewcommand{\thetable}{B.\arabic{table}}
\renewcommand{\theequation}{B.\arabic{equation}}
\setcounter{figure}{0}
\setcounter{table}{0}
\setcounter{equation}{0}
\subsubsection{Consistency Checks}
\label{app:state_consistency_probes}

As introduced in Task 2 (Section~\ref{subsec:eval_goals}), Structural Consistency Probes are the deterministic rules for auditing the structural relationships of a model's prediction \emph{before} invoking any LLM-as-judge. To transition these from abstract evaluation concepts into an executable test suite, we formalize them as logical predicates. These probes automatically verify whether the predicted units utilize valid reference IDs, possess logically compatible time ranges, and maintain consistent cross-modal links.

\paragraph{Failure Conditions}

We define the following formal predicates, each tied to an exact, programmatic failure trigger:
\begin{itemize}
    \item \(\mathrm{ValidRefUse}(x, r)\): For any event or shot \(x\), if \(x\) claims to feature reference \(r\), then \(r\) must exist in the top-level references list. \emph{Failure trigger}: A dangling ID is utilized.
    \item \(\mathrm{ValidEventShotAssociation}(e, h)\): If event \(e\) is linked to shot \(h\), they must exhibit a strictly positive temporal overlap (\(\mathrm{IoU}(time(e), time(h)) > 0\)). \emph{Failure trigger}: An event is linked to a completely disjoint visual shot.
    \item \(\mathrm{TemporalCompatible}(e, h)\): The duration of sub-elements, including subshots or active events, must be bounded by, or reasonably overlap with, their parent shot's duration. \emph{Failure trigger}: A subshot's timestamp significantly exceeds the boundaries of its parent shot.
    \item \(\mathrm{ConsistentCoreference}(r \text{ across } \mathcal{H})\): An ID mapped to a persistent entity must map consistently across the video's timeline without splitting into hallucinated variants. \emph{Failure trigger}: A tracked character abruptly changes ID assignment across sequential cuts.
    \item \(\mathrm{SupportFieldPresent}(r)\): Every reference \(r\) must possess its required evidence fields, most notably, the \texttt{appearance\_anchor}. \emph{Failure trigger}: The prediction omits the \texttt{appearance\_anchor} dictionary entirely.
    \item \(\mathrm{NoDanglingID}(\hat{\mathcal{S}})\): No unreferenced or circular links exist in the final prediction graph. \emph{Failure trigger}: A shot claims it contains an event ID that was never defined.
\end{itemize}

Table~\ref{tab:probe_types} summarizes these probe families and their targeted failure conditions.

\begin{table}[htbp]
\centering
\caption{\textbf{Structural Consistency Probe Types.} Each row defines a deterministic probe family, its formal predicate, and the exact condition counted as a failure. Proportions are computed over the subset of structurally applicable predictions for the Gemini 3.1-Pro model to isolate structural reasoning from raw detection capability.}
\small
\resizebox{\linewidth}{!}{
\begin{tabular}{p{3.3cm}p{3.5cm}p{4.1cm}c}
\toprule
\textbf{Probe type} & \textbf{Formal Predicate} & \textbf{Failure condition} & \textbf{Count / Prop.} \\
\midrule
Reference-use validity & \(\mathrm{ValidRefUse}(x, r)\) & Missing aligned region, dangling ID, or reference mismatch after alignment & 8649 / 34.1\% \\
Temporal compatibility & \(\mathrm{TemporalCompatible}(e, h)\) & Predicted interval is disjoint from the required shot/event support or violates ordering tolerance & 12240 / 42.0\% \\
Reference-support validity & \(\mathrm{SupportFieldPresent}(r)\) & Missing support field, malformed support field, or support field incompatible with the matched reference & 3778 / 31.7\% \\
Cross-reference integrity & \(\mathrm{ConsistentCoreference}(r)\) & Referenced ID is undefined, inconsistent across shots, or incompatible with the matched persistent reference & 2656 / 65.2\% \\
\bottomrule
\end{tabular}}
\label{tab:probe_types}
\end{table}

\paragraph{Isolating Structural Reasoning from Raw Detection.}
Table~\ref{tab:probe_types} audits predictions strictly on their structural consistency. The high failure rates in Cross-reference integrity (65.2\%) and Temporal compatibility (42.0\%) highlight that the true bottleneck for current omnimodal architectures is not isolated frame perception, but rather building a coherent spatiotemporal graph. Models often hallucinate entity IDs or assign disjoint time boundaries even when the basic visual components are correctly detected. This validates our design of deterministic consistency probes: structural compliance must be verified programmatically before invoking LLMs for semantic validation, otherwise LLMs may be inadvertently prompted to score physically impossible or internally contradictory scenes.

\subsubsection{Alignment Protocol}
\label{app:structural_alignment}

Before calculating any structural metrics, we canonicalize raw model outputs into a standardized evaluation graph and then align prediction units to the ground-truth units. This pipeline explicitly tracks and preserves the structural intent behind raw JSON outputs. Canonicalization, temporal matching, and rule checks are rule-based; reference alignment is a bounded LLM-assisted one-to-one matching step over reference detail descriptions. All thresholds and tie-breaking rules for rule-based matching are fixed \emph{prior} to computing model scores.

    \begin{tcolorbox}[
    enhanced,
    colback=uclablue!8,
    colframe=uclablue,
    colbacktitle=uclablue,
    coltitle=black,
    fonttitle={\bfseries\sffamily},
    title={\textbf{Algorithm 1:} Output Canonicalization and Structural Alignment},
    boxrule=1.2pt,
    arc=4pt,
    drop fuzzy shadow
]
    \small
    \textbf{Input:} Predicted JSON $\hat{V}$, Ground Truth $V^\star$ \\
    \textbf{Output:} Matched units $M_R, M_E, M_H, M_S$
    \vspace{0.5em}
    \hrule
    \vspace{0.5em}
    \begin{enumerate}[leftmargin=1.5em, itemsep=0.2em]
        \item $\hat{V} \gets \mathrm{FixMalformedIDs}(\hat{V})$ \hfill \textit{\color{gray}// Resolve formatting errors}
        \item $\hat{V} \gets \mathrm{ClipTimeRanges}(\hat{V}, V^\star_{max\_time})$ \hfill \textit{\color{gray}// Clamp out-of-bounds timestamps}
        \item $\hat{\mathcal{R}}, \hat{\mathcal{E}}, \hat{\mathcal{H}} \gets \mathrm{ParseAndFilterDanglingLinks}(\hat{V})$
        \item $M_R \gets \mathrm{AlignReferences}_{LLM}(\hat{\mathcal{R}}, \mathcal{R}^\star)$ \hfill \textit{\color{gray}// Bounded one-to-one over detail descriptions}
        \item $M_E \gets \mathrm{AlignEvents}(\hat{\mathcal{E}}, \mathcal{E}^\star)$ \hfill \textit{\color{gray}// WER for dialog, tIoU otherwise}
        \item $M_H \gets \mathrm{AlignShots}(\hat{\mathcal{H}}, \mathcal{H}^\star)$ \hfill \textit{\color{gray}// Temporal IoU}
        \item $M_S \gets \mathrm{LocalAlignSubshots}(M_H, \delta=1\text{s})$
        \item \textbf{return} $\mathrm{ComputeMetricFamilies}(M_R, M_E, M_H, M_S)$
    \end{enumerate}
    \end{tcolorbox}

\begin{tcolorbox}[colback=uclablue!8, colframe=uclablue, colbacktitle=uclablue, coltitle=black, boxrule=1pt, title={\textbf{Cross-Track Linking Instruction (Excerpt)}}]
\scriptsize\ttfamily
The most important step: Use the `active\_events` array to accurately link the current visual shot with the audio events happening simultaneously.\\
- Every `event\_id` listed in a shot's `active\_events` array MUST come from the previously defined `events` list.\\
- Fill this accurately based on temporal overlap: if any part of an event's `time\_range` overlaps with the current shot's `time\_range`, that `event\_id` MUST be included in `active\_events`.\\
- CROSS-VALIDATION: Check that the union of `active\_events` across all shots covers every single `event\_id` defined in the video. Do not leave any event dangling.
\end{tcolorbox}

\section{More Experimental Results}
\label{app:more_results}

\subsection{Intra-Family Metric Correlation}
\label{app:metric_correlation}

\begin{figure}[htbp]
    \centering
    \includegraphics[width=\linewidth]{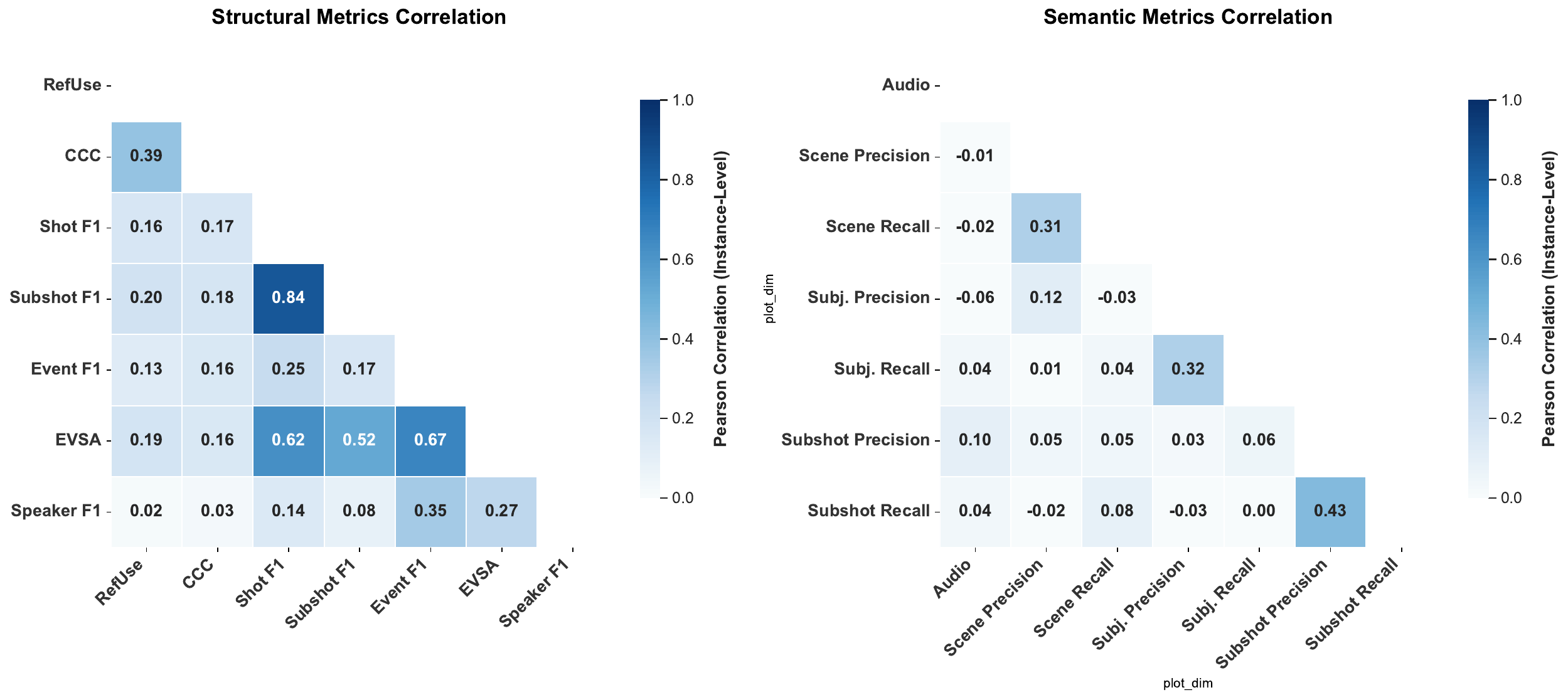}
    \caption{\textbf{Instance-level correlation among diagnostic metrics.} Pearson correlation coefficients computed across 786 videos (Gemini 3.1-Pro). \textbf{Left:} Structural correlations confirm that complex capabilities like identity tracking (\emph{CCC}) are decoupled from basic temporal grounding (\emph{Shot F1}, $r=0.17$). \textbf{Right:} Semantic LLM-as-judge scores exhibit near-zero correlation between Precision and Recall for Subject descriptions ($r=-0.03$), proving that hallucination tendency and factual coverage are independent failure modes.}
    \label{fig:metric_correlation_appendix}
\end{figure}

A critical design goal of \name{} is to provide fine-grained, decoupled diagnostic signals. To validate that our deep-structured atomic units measure distinct capabilities rather than redundant phenomena, we computed the instance-level (per-video) Pearson correlation among the metrics across all 786 test videos, using the predictions of Gemini 3.1-Pro. As illustrated in Figure~\ref{fig:metric_correlation_appendix}, we separate these into two distinct domains: Structural Metrics and Semantic Metrics.

The \textbf{Structural Metrics Correlation} heatmap reveals that fundamental spatial, temporal, and cross-modal capabilities are highly decoupled. For instance, the correlation between \emph{Event F1} (audio perception) and \emph{Shot F1} (visual temporal grounding) is remarkably low ($r=0.25$), indicating that models often succeed in one modality while failing in the other. While \emph{EVSA} (audio-visual association) naturally correlates with both its audio ($r=0.67$) and visual ($r=0.62$) prerequisites, complex tracking metrics remain isolated. \emph{CCC} (cross-shot tracking) shows almost no correlation with basic temporal grounding (\emph{Shot F1}, $r=0.17$), proving that locating an action in time is fundamentally separate from persistently tracking the actor's identity across discontinuous cuts.

The \textbf{Semantic Metrics Correlation} heatmap yields an equally crucial insight: descriptive fidelity is heavily fragmented. Strikingly, \emph{Subject Precision} and \emph{Subject Recall} exhibit zero correlation ($r=-0.03$). This proves that a model's tendency to hallucinate vivid details (low precision) is completely decoupled from its ability to comprehensively cover all ground-truth facts (recall). Furthermore, cross-domain semantic scores, such as \emph{Audio} descriptions versus \emph{Scene Recall} ($r=-0.02$), are entirely uncorrelated. This quantitative analysis confirms that \name{}'s suite of metrics effectively isolates distinct multimodal capabilities, exposing localized failures (e.g., severe hallucination despite high recall) that holistic scalars seamlessly average away.

\subsection{Detailed Setup}
\label{app:caption_state_gap_details}

In Section~\ref{sec:experiments}, we argue that traditional dense caption evaluation using text-centric metrics can falsely equate distinct models. To empirically validate this claim, we conducted an experiment using cases from the UGC-VideoCap~\citep{wu2025ugcvideocaptioner} and video-SALMONN 2~\citep{tang2025videosalmonn2} benchmarks, both of which rely on whole-caption evaluation paradigms.

Specifically, we constructed a diagnostic subset of 200 videos to serve as a ``blind spot'' dataset for traditional metrics. The construction followed these steps:
\begin{enumerate}[leftmargin=*]
    \item \textbf{Source Log Analysis:} We collected the dense caption outputs and corresponding global text-level scores for distinct frontier and open-source models evaluated on UGC-VideoCap and video-SALMONN 2.
    \item \textbf{Filtering for Pseudo-Equivalence:} We computed the variance in the text-level scores among the models for each individual video. We then specifically filtered for cases where the maximum score difference between models was strictly less than $10\%$.
    \item \textbf{Subset Construction:} We sampled 200 such cases to form our final test set. By definition, traditional global text scoring evaluates these models as exhibiting nearly identical performance on this subset.
    \item \textbf{Structured Re-Evaluation:} Finally, we evaluated the models' structural comprehension on these exact 200 videos using \name{}'s deep-structured atomic unit metrics.
\end{enumerate}

As discussed in the main text and shown in Table~\ref{tab:caption_state_gap_result}, while the models appeared identical under global text scoring, \name{}'s localized metrics revealed significant underlying quality gaps. This confirms that global text metrics frequently reward linguistically fluent but structurally flawed outputs, degrading diagnostic resolution.

\subsection{Full Main Results}
\label{app:full_results}

While the main text presents compact metric families to highlight macro-level capability gaps, rigorous diagnostic evaluation requires exposing the underlying structural submetrics. This section provides the fully expanded results for all evaluated models. Tables~\ref{tab:full_metrics_visual} and \ref{tab:full_metrics_audio} separate the rule-based compact columns into their constituent Precision, Recall, F1, and tIoU components. For instance, expanding the Audio dimension reveals whether a low Event F1 score is driven by hallucination (low precision) or omission (low recall). Table~\ref{tab:full_metrics_llm} reports the localized LLM-as-judge scores across specific descriptive fields. Finally, Tables~\ref{tab:per_duration_results} and \ref{tab:per_duration_llm_results} stratify performance by video duration ($<1$ min, $1$--$3$ min, and $3$--$5$ min), testing whether models maintain identity persistence and temporal grounding over extended contexts. Numeric tables use the same column-wise \textbf{bold}/\underline{underline} convention as the main paper (best and second-best, ties allowed).

\paragraph{Deep Analysis of Full Submetrics.}
By unpacking the compact F1 scores into their base Recall and Precision constituents, we uncover distinct behavioral paradigms across the models. In the \textbf{Visual} domain (Table~\ref{tab:full_metrics_visual}), Gemini 3.1-Pro and Gemini 2.5-Pro exhibit diverging strategies: Gemini 3.1-Pro dominates Reference Recall (77.02\%) but lags slightly behind in Reference Precision (70.56\%), suggesting an aggressive entity detection strategy. Conversely, Gemini 2.5-Pro achieves higher Precision (71.39\%) but lower Recall. Furthermore, the structural parsing of shots reveals that while models can generally pinpoint when actions occur (Shot tIoU > 70\%), their Subshot Precision plummets (e.g., Gemini 3.1-Pro drops from 88.77\% for Shots to 60.29\% for Subshots), proving that models hallucinate fine-grained temporal boundaries when forced to decompose continuous actions into structured sub-events. 

In the \textbf{Audio-Visual} domain (Table~\ref{tab:full_metrics_audio}), the expanded EVSA (Event-Shot Association) metric isolates the cross-modal bottleneck. Gemini 3.1-Pro's EVSA Precision is high (78.35\%): when the model does link an audio event to a visual shot, the topological link is usually correct. Its EVSA Recall, however, is severely low (42.14\%), confirming that the primary failure mode is omission—models perceive the audio and visual streams in isolation, miss their causal relationships, and drop the required cross-modal links entirely. This reinforces our core motivation: purely text-based evaluations cover up such missing links with generic paragraphs, whereas \name{} structurally mandates them.

\begin{table*}[!t]
\centering
\caption{\textbf{Expanded Visual rule-based submetrics.} Ref R/P are reference recall and precision. RefUse-shot and RefUse-sub report \texttt{WHO\_shot} and \texttt{WHO\_sub}. CCC is the macro average of cross-shot reference-trajectory IoU. Shot/Subshot R/P report alignment recall and precision, and tIoU reports temporal overlap (\texttt{WHEN\_H}, \texttt{WHEN\_S}). Within each column, \textbf{bold} is best and \underline{underline} is second-best among models (ties allowed).}
\label{tab:full_metrics_visual}
\scriptsize
\setlength{\tabcolsep}{2.2pt}
\resizebox{\linewidth}{!}{
\begin{tabular}{@{}l ccc ccc ccc ccc@{}}
\toprule
\textbf{Model} & \textbf{RefUse-shot} & \textbf{RefUse-sub} & \textbf{CCC} & \textbf{Ref R} & \textbf{Ref P} & \textbf{Shot R} & \textbf{Shot P} & \textbf{Shot tIoU} & \textbf{Subshot R} & \textbf{Subshot P} & \textbf{Subshot tIoU} \\
\midrule
\rowcolor{limegreen!50} \multicolumn{12}{c}{\textbf{\textit{Proprietary Models}}} \\
\midrule
Gemini 3.1-Pro~\citep{gemini2024} & \textbf{68.49} & \underline{68.77} & \underline{34.43} & \textbf{77.02} & 70.56 & \textbf{71.73} & 88.77 & 70.74 & \textbf{75.40} & 60.29 & 48.76 \\
Gemini 2.5-Pro~\citep{gemini25} & \underline{67.01} & \textbf{71.42} & \textbf{37.81} & \underline{76.96} & \textbf{71.39} & \underline{66.29} & 86.27 & \textbf{72.58} & \underline{70.26} & \textbf{66.89} & \underline{49.18} \\
Qwen3.5-Omni-Plus~\citep{qwenteam2026qwen35omni} & 63.99 & 65.22 & 29.55 & 65.02 & \underline{70.99} & 63.90 & \underline{89.62} & \underline{72.35} & 69.00 & \underline{66.66} & 48.29 \\
Qwen3.5-Omni-Flash~\citep{qwenteam2026qwen35omni} & 59.39 & 61.07 & 24.31 & 59.50 & 69.26 & 55.77 & \textbf{89.71} & 68.99 & 61.93 & 60.45 & 46.52 \\
\midrule
\rowcolor{grayred!40} \multicolumn{12}{c}{\textbf{\textit{Open-source Omnimodal Models}}} \\
\midrule
Qwen3-Omni-Instruct~\citep{xu2025qwen3omni} & 37.97 & 50.37 & 9.85 & 50.63 & 53.53 & 45.98 & 64.36 & 67.92 & 35.80 & 28.79 & 47.95 \\
Qwen3-Omni-Captioner~\citep{xu2025qwen3omni} & 40.88 & 50.80 & 11.51 & 49.44 & 46.97 & 47.79 & 71.63 & 70.80 & 44.62 & 42.29 & \textbf{49.69} \\
\bottomrule
\end{tabular}}
\end{table*}

\begin{table*}[!t]
\centering
\caption{\textbf{Expanded Audio and Audio-Visual rule-based submetrics.} Within each column, \textbf{bold} is best and \underline{underline} is second-best among models (ties allowed).}
\label{tab:full_metrics_audio}
\scriptsize
\setlength{\tabcolsep}{3.5pt}
\resizebox{\linewidth}{!}{
\begin{tabular}{@{}l ccc cc ccc@{}}
\toprule
\multirow{2}{*}{\textbf{Model}} & \multicolumn{3}{c}{\textbf{Audio}} & \multicolumn{4}{c}{\textbf{Audio-Visual}} \\
\cmidrule(lr){2-4} \cmidrule(lr){5-8}
& \textbf{Event R} & \textbf{Event P} & \textbf{Event tIoU} & \textbf{Speaker} & \textbf{EVSA-P} & \textbf{EVSA-R} & \textbf{EVSA} \\
\midrule
\rowcolor{limegreen!50} \multicolumn{8}{c}{\textbf{\textit{Proprietary Models}}} \\
\midrule
Gemini 3.1-Pro~\citep{gemini2024} & \textbf{63.96} & \textbf{67.20} & \textbf{71.92} & \underline{91.18} & \textbf{78.35} & \textbf{42.14} & \textbf{51.46} \\
Gemini 2.5-Pro~\citep{gemini25} & \underline{58.09} & \underline{64.23} & \underline{69.35} & 89.72 & \underline{74.05} & \underline{35.04} & \underline{43.74} \\
Qwen3.5-Omni-Plus~\citep{qwenteam2026qwen35omni} & 51.22 & 62.53 & 67.77 & \textbf{91.20} & 71.11 & 31.82 & 40.20 \\
Qwen3.5-Omni-Flash~\citep{qwenteam2026qwen35omni} & 48.01 & 58.91 & 69.34 & 89.91 & 66.28 & 27.69 & 35.58 \\
\midrule
\rowcolor{grayred!40} \multicolumn{8}{c}{\textbf{\textit{Open-source Omnimodal Models}}} \\
\midrule
Qwen3-Omni-Instruct~\citep{xu2025qwen3omni} & 39.73 & 49.67 & 40.99 & 88.35 & 28.26 & 9.92 & 13.33 \\
Qwen3-Omni-Captioner~\citep{xu2025qwen3omni} & 39.64 & 50.07 & 52.60 & 88.68 & 36.70 & 13.00 & 17.42 \\
\bottomrule
\end{tabular}}
\end{table*}

\begin{table*}[!t]
\centering
\caption{\textbf{Expanded LLM-as-judge submetrics.} Dialogue is the rule-based dialogue line structural score. Within each column, \textbf{bold} is best and \underline{underline} is second-best among models (ties allowed).}
\label{tab:full_metrics_llm}
\scriptsize
\setlength{\tabcolsep}{3pt}
\resizebox{\linewidth}{!}{
\begin{tabular}{@{}l ccc ccc c c c@{}}
\toprule
\multirow{3}{*}{\textbf{Model}} & \multicolumn{7}{c}{\textbf{Visual}} & \multicolumn{2}{c}{\textbf{Audio}} \\
\cmidrule(lr){2-8} \cmidrule(lr){9-10}
& \multicolumn{3}{c}{Subject Ref} & \multicolumn{3}{c}{Scene Ref} & Shot Fact & Dialogue & Non-dialogue \\
\cmidrule(lr){2-4} \cmidrule(lr){5-7} \cmidrule(lr){8-8} \cmidrule(lr){9-9} \cmidrule(lr){10-10}
& \textbf{R} & \textbf{P} & \textbf{Mean} & \textbf{R} & \textbf{P} & \textbf{Mean} & \textbf{Mean} & \textbf{Mean} & \textbf{Mean} \\
\midrule
\rowcolor{limegreen!50} \multicolumn{10}{c}{\textbf{\textit{Proprietary Models}}} \\
\midrule
Gemini 3.1-Pro~\citep{gemini2024} & 43.78 & \textbf{67.07} & \textbf{55.43} & \underline{51.78} & \textbf{75.62} & \textbf{63.70} & \textbf{59.05} & \underline{91.18} & \textbf{41.49} \\
Gemini 2.5-Pro~\citep{gemini25} & \textbf{51.53} & 57.08 & 54.30 & \textbf{56.80} & 65.12 & \underline{60.96} & \underline{55.58} & 89.72 & \underline{37.38} \\
Qwen3.5-Omni-Plus~\citep{qwenteam2026qwen35omni} & \underline{45.41} & 62.84 & 54.13 & 50.75 & 69.80 & 60.27 & 54.17 & \textbf{91.20} & 36.82 \\
Qwen3.5-Omni-Flash~\citep{qwenteam2026qwen35omni} & 43.87 & \underline{65.59} & \underline{54.73} & 44.41 & \underline{71.26} & 57.84 & 47.66 & 89.91 & 28.12 \\
\midrule
\rowcolor{grayred!40} \multicolumn{10}{c}{\textbf{\textit{Open-source Omnimodal Models}}} \\
\midrule
Qwen3-Omni-Instruct~\citep{xu2025qwen3omni} & 40.31 & 62.02 & 51.16 & 36.07 & 70.73 & 53.40 & 34.63 & 88.35 & 20.51 \\
Qwen3-Omni-Captioner~\citep{xu2025qwen3omni} & 42.00 & 50.96 & 46.48 & 38.07 & 54.53 & 46.30 & 36.08 & 88.68 & 19.65 \\
\bottomrule
\end{tabular}}
\end{table*}

\begin{table*}[!t]
\centering
\caption{\textbf{Per-duration rule-based structural results.} Stratification follows the three duration buckets ($<1$ min, $1$--$3$ min, and $3$--$5$ min); Ref Subj./Ref S. are recomputed per video in each bucket as in the main table. Within each column, \textbf{bold} is best and \underline{underline} is second-best among models (ties allowed).}
\label{tab:per_duration_results}
\scriptsize
\setlength{\tabcolsep}{2.0pt}
\resizebox{\textwidth}{!}{%
\begin{tabular}{@{}l c ccccccccc cc cc@{}}
\toprule
\multirow{2}{*}{\textbf{Model}} &
\multirow{2}{*}{\textbf{SGC}} &
\multicolumn{8}{c}{\textbf{Visual}} &
\multicolumn{2}{c}{\textbf{Audio}} &
\multicolumn{2}{c}{\textbf{Audio-Visual}} \\
\cmidrule(lr){3-10} \cmidrule(lr){11-12} \cmidrule(lr){13-14}
& &
\textbf{RefUse} & \textbf{CCC} & \textbf{Ref Subj. F1} & \textbf{Ref S. F1} & \textbf{Shot F1} & \textbf{Shot tIoU} & \textbf{Sub F1} & \textbf{Sub tIoU} &
\textbf{Evt F1} & \textbf{Evt tIoU} &
\textbf{Spk F1} & \textbf{EVSA F1} \\
\midrule
\rowcolor{lightgray} \multicolumn{14}{c}{\textbf{\textit{Duration: $<$1 min}}} \\
\midrule
Gemini 3.1-Pro & \underline{97.46} & \textbf{74.13} & \underline{38.41} & \underline{83.09} & \underline{86.54} & \textbf{77.90} & 69.76 & 67.55 & 47.73 & \textbf{63.09} & \textbf{70.65} & \textbf{91.15} & \textbf{53.30} \\
Gemini 2.5-Pro & \textbf{97.57} & \underline{73.40} & \textbf{40.60} & \textbf{84.80} & \textbf{87.39} & 76.61 & \underline{71.72} & \textbf{70.94} & \underline{48.36} & \underline{60.46} & \underline{67.46} & 89.08 & \underline{49.31} \\
Qwen3.5-Omni-Plus & 96.90 & 70.94 & 32.71 & 80.28 & 84.99 & \underline{77.60} & \textbf{73.57} & \underline{69.65} & \textbf{48.73} & 57.12 & 64.57 & \underline{90.77} & 46.87 \\
Qwen3.5-Omni-Flash & 96.26 & 66.83 & 26.81 & 77.37 & 66.70 & 72.86 & 69.94 & 64.57 & 47.07 & 52.72 & 66.90 & 89.34 & 41.65 \\
Qwen3-Omni-Instruct & 93.91 & 50.85 & 13.42 & 68.94 & 76.29 & 40.06 & 52.58 & 27.29 & 38.83 & 44.41 & 47.30 & 88.32 & 17.09 \\
Qwen3-Omni-Captioner & 94.15 & 54.17 & 16.45 & 69.03 & 76.25 & 51.50 & 55.64 & 42.08 & 41.09 & 46.57 & 55.78 & 88.54 & 24.51 \\
\midrule
\rowcolor{lightgray} \multicolumn{14}{c}{\textbf{\textit{Duration: 1--3 min}}} \\
\midrule
Gemini 3.1-Pro & \textbf{97.84} & \underline{65.61} & \underline{29.59} & \underline{72.09} & \underline{76.14} & \textbf{78.87} & 73.84 & \underline{65.76} & 52.17 & \textbf{67.31} & \textbf{77.44} & \textbf{92.22} & \textbf{54.76} \\
Gemini 2.5-Pro & \underline{96.47} & \textbf{66.99} & \textbf{35.43} & \textbf{72.70} & \textbf{77.32} & \underline{69.25} & 75.17 & \textbf{66.68} & 52.01 & \underline{57.77} & 73.12 & 91.06 & \underline{39.80} \\
Qwen3.5-Omni-Plus & 95.52 & 60.98 & 25.64 & 64.09 & 70.40 & 68.70 & 71.65 & 65.75 & 49.32 & 53.40 & 72.80 & \underline{92.07} & 35.03 \\
Qwen3.5-Omni-Flash & 94.99 & 56.99 & 21.17 & 63.76 & 63.72 & 61.31 & 68.57 & 55.76 & 47.21 & 49.38 & \underline{75.16} & 90.83 & 29.70 \\
Qwen3-Omni-Instruct & 83.25 & 26.52 & 4.51 & 35.84 & 48.05 & 67.62 & \textbf{94.30} & 35.87 & \textbf{63.60} & 38.09 & 32.53 & 88.84 & 9.12 \\
Qwen3-Omni-Captioner & 86.87 & 33.13 & 4.82 & 36.29 & 32.86 & 55.32 & \underline{93.30} & 37.98 & \underline{63.48} & 34.25 & 49.30 & 88.72 & 8.98 \\
\midrule
\rowcolor{lightgray} \multicolumn{14}{c}{\textbf{\textit{Duration: 3--5 min}}} \\
\midrule
Gemini 3.1-Pro & \textbf{95.31} & \underline{60.57} & \underline{23.21} & \underline{69.11} & \underline{77.09} & \textbf{57.79} & 67.76 & \textbf{49.21} & 45.18 & \textbf{53.01} & 63.06 & 88.26 & \textbf{30.02} \\
Gemini 2.5-Pro & 92.12 & \textbf{64.89} & \textbf{27.02} & \textbf{78.16} & \textbf{78.35} & 39.43 & 70.38 & 39.42 & 46.03 & \underline{48.22} & \underline{69.53} & 89.24 & \underline{19.83} \\
Qwen3.5-Omni-Plus & 92.55 & 58.82 & 20.96 & 58.84 & 76.43 & 38.10 & 66.63 & \underline{40.08} & 42.37 & 34.95 & \textbf{73.16} & \textbf{90.95} & 13.14 \\
Qwen3.5-Omni-Flash & \underline{93.04} & 57.36 & 17.62 & 64.84 & 69.00 & 32.29 & 64.08 & 32.30 & 40.74 & 41.44 & 66.60 & \underline{89.87} & 14.36 \\
Qwen3-Omni-Instruct & 76.59 & 24.15 & 4.18 & 26.16 & 42.34 & \underline{57.39} & \textbf{95.00} & 23.98 & \underline{62.77} & 36.12 & 25.88 & 86.97 & 3.28 \\
Qwen3-Omni-Captioner & 83.97 & 29.92 & 4.44 & 28.07 & 23.09 & 47.58 & \underline{93.51} & 29.99 & \textbf{63.43} & 33.67 & 45.74 & 89.16 & 4.16 \\
\bottomrule
\end{tabular}%
}
\end{table*}

\begin{table*}[!t]
\centering
\caption{\textbf{Per-duration local semantic-equivalence score results.} Stratification follows the three duration buckets ($<1$ min, $1$--$3$ min, and $3$--$5$ min). Within each column, \textbf{bold} is best and \underline{underline} is second-best among models.}
\label{tab:per_duration_llm_results}
\scriptsize
\setlength{\tabcolsep}{3pt}
\resizebox{\textwidth}{!}{%
\begin{tabular}{@{}l ccccccc cc@{}}
\toprule
\multirow{3}{*}{\textbf{Model}} & \multicolumn{7}{c}{\textbf{Visual}} & \multicolumn{2}{c}{\textbf{Audio}} \\
\cmidrule(lr){2-8} \cmidrule(lr){9-10}
& \multirow{2}{*}{\textbf{Shot}} & \multicolumn{2}{c}{\textbf{Subject}} & \multicolumn{2}{c}{\textbf{Scene}} & \multicolumn{2}{c}{\textbf{Subshot}} & \multirow{2}{*}{\textbf{Dialogue}} & \multirow{2}{*}{\textbf{Non-dialogue}} \\
\cmidrule(lr){3-4} \cmidrule(lr){5-6} \cmidrule(lr){7-8}
& & \textbf{Recall} & \textbf{Precision} & \textbf{Recall} & \textbf{Precision} & \textbf{Recall} & \textbf{Precision} & & \\
\midrule
\rowcolor{lightgray} \multicolumn{10}{c}{\textbf{\textit{Duration: $<$1 min}}} \\
\midrule
Gemini 3.1-Pro & \textbf{62.84} & 46.74 & \textbf{68.18} & \underline{52.62} & \textbf{78.01} & \underline{53.90} & \underline{71.75} & \textbf{91.15} & \textbf{43.39} \\
Gemini 2.5-Pro & \underline{59.70} & \textbf{54.97} & 57.11 & \textbf{57.66} & 67.97 & \textbf{54.19} & \textbf{72.40} & 89.08 & \underline{40.03} \\
Qwen3.5-Omni-Plus & 58.92 & \underline{48.27} & \underline{62.41} & 51.83 & 71.96 & 51.88 & 70.08 & \underline{90.77} & 38.23 \\
Qwen3.5-Omni-Flash & 51.38 & 44.16 & 62.10 & 46.15 & 72.85 & 45.58 & 66.84 & 89.34 & 28.94 \\
Qwen3-Omni-Instruct & 39.62 & 39.99 & 62.06 & 36.46 & \underline{73.31} & 34.64 & 65.12 & 88.32 & 23.71 \\
Qwen3-Omni-Captioner & 42.82 & 41.44 & 51.46 & 38.73 & 56.76 & 36.28 & 49.78 & 88.54 & 23.21 \\
\midrule
\rowcolor{lightgray} \multicolumn{10}{c}{\textbf{\textit{Duration: 1--3 min}}} \\
\midrule
Gemini 3.1-Pro & \textbf{59.05} & 40.96 & \underline{65.67} & \underline{51.34} & \textbf{73.61} & \textbf{54.35} & \underline{70.90} & \textbf{92.22} & \textbf{40.03} \\
Gemini 2.5-Pro & \underline{54.75} & \textbf{49.53} & 57.48 & \textbf{56.33} & 62.88 & \underline{52.95} & 70.14 & 91.06 & 33.44 \\
Qwen3.5-Omni-Plus & 53.57 & 42.86 & 63.97 & 50.39 & 68.11 & 50.20 & \textbf{71.75} & \underline{92.07} & \underline{34.91} \\
Qwen3.5-Omni-Flash & 47.16 & \underline{43.29} & \textbf{68.36} & 42.76 & \underline{69.67} & 44.45 & 67.48 & 90.83 & 24.12 \\
Qwen3-Omni-Instruct & 34.11 & 40.66 & 62.37 & 35.70 & 66.24 & 30.31 & 47.68 & 88.84 & 12.50 \\
Qwen3-Omni-Captioner & 34.85 & 42.64 & 50.00 & 38.03 & 50.59 & 33.82 & 42.20 & 88.72 & 9.38 \\
\midrule
\rowcolor{lightgray} \multicolumn{10}{c}{\textbf{\textit{Duration: 3--5 min}}} \\
\midrule
Gemini 3.1-Pro & \textbf{54.15} & 43.41 & \underline{68.01} & \underline{49.32} & \textbf{71.23} & \underline{49.68} & 68.88 & 88.26 & 33.85 \\
Gemini 2.5-Pro & \underline{48.07} & \textbf{47.52} & 55.83 & \textbf{54.59} & 60.37 & \textbf{50.25} & \textbf{70.16} & 89.24 & \textbf{35.29} \\
Qwen3.5-Omni-Plus & 43.49 & 43.09 & 60.71 & 46.43 & 64.29 & 45.43 & \underline{69.19} & \textbf{90.95} & \underline{34.03} \\
Qwen3.5-Omni-Flash & 39.16 & \underline{44.60} & \textbf{69.17} & 42.20 & \underline{69.50} & 38.80 & 68.37 & \underline{89.87} & 32.81 \\
Qwen3-Omni-Instruct & 33.04 & 41.13 & 60.22 & 34.43 & 68.85 & 23.54 & 63.59 & 86.97 & 14.71 \\
Qwen3-Omni-Captioner & 32.36 & 42.13 & 52.36 & 30.49 & 48.17 & 30.81 & 46.04 & 89.16 & 21.30 \\
\bottomrule
\end{tabular}%
}
\end{table*}

\subsection{Case Studies}
\label{app:case_study}

Our core motivation is to evaluate video captions through explicit, verifiable units rather than through a single holistic text score. This section presents two qualitative case studies that illustrate what this structured representation makes observable. The first case focuses on \emph{identity persistence}: whether a model uses the same entity identifier for the same visual subject across temporally separated shots. The second case focuses on \emph{event-shot association}: whether a recognized audio event is linked to the visual shot that supports its source. Together, these examples show how \name{} turns broad caption quality into localized diagnostic evidence.

\subsubsection{Identity Persistence: Cross-Shot Coreference Consistency}

Cross-Shot Coreference Consistency (CCC) measures whether a model maintains a stable identity assignment for the same entity across multiple shots. This is different from detecting that a person or object appears somewhere in the video. A model may recognize the entity locally in each shot, but still fragment the entity into different identifiers after viewpoint changes, occlusions, or temporal gaps. CCC therefore evaluates whether the predicted reference graph is temporally consistent.

\begin{figure}[htbp]
    \centering
    \includegraphics[width=0.9\linewidth]{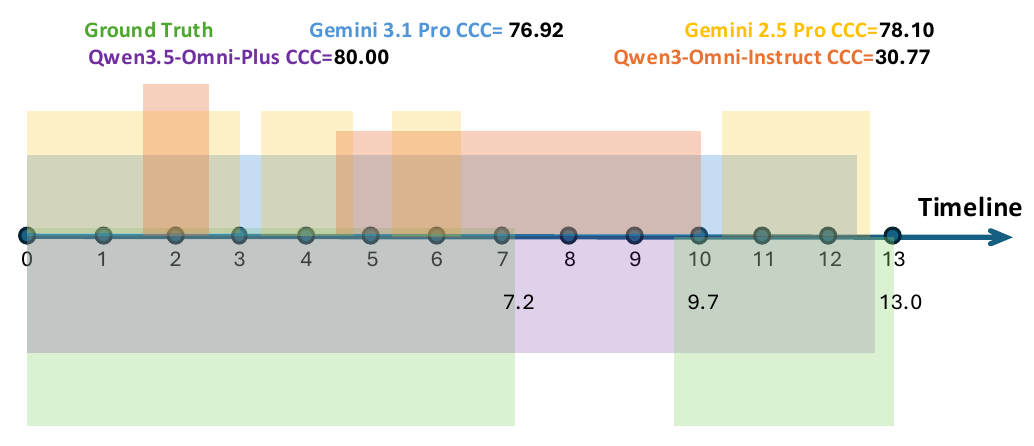}
    \caption{\textbf{Case study of identity persistence measured by CCC.}
    The timeline compares identity assignments across models over the same video interval. Colored spans show ground-truth and predicted identity coverage. Higher CCC indicates more consistent cross-shot identity reuse; lower CCC indicates identity fragmentation. In this case, Gemini 3.1 Pro, Gemini 2.5 Pro, and Qwen3.5-Omni-Plus score 76.92, 78.10, and 80.00, while Qwen3-Omni-Instruct scores 30.77, revealing a clear failure in preserving identity continuity.}
    \label{fig:case_study_good}
\end{figure}

Figure~\ref{fig:case_study_good} shows that CCC provides a direct diagnostic signal for identity persistence. The key comparison is not whether a model can mention the visible subject, but whether it preserves the same identity assignment over time. In the displayed example, three models maintain relatively consistent identity traces, with CCC scores above 76, while Qwen3-Omni-Instruct obtains 30.77. This indicates that its prediction is more fragmented at the structural level. The implication is that CCC isolates a specific failure mode: identity drift across the video timeline. This failure would be difficult to identify from a fluent caption alone, because a paragraph can describe the visible subject in each shot without exposing whether the model has maintained a stable entity graph.

\subsubsection{Audio-Visual Association: Event-Shot Association}

Event-Shot Association (EVSA) evaluates a different type of structure. It checks whether each audio event is linked to the visual shot or shots that support its source. This distinction matters because recognizing an audio category and grounding that audio category in the concurrent visual timeline are different capabilities. A model may detect that a dialogue or mechanical sound exists, but still omit the shot-level edge that makes the audio source verifiable.

\begin{tcolorbox}[
    colback=blue!4,
    colframe=uclablue,
    boxrule=0.6pt,
    title={\textbf{Case study of event-shot association measured by EVSA}},
    fonttitle=\small\bfseries,
    left=5pt,
    right=5pt,
    top=5pt,
    bottom=5pt,
    breakable
]
\scriptsize
\textbf{Shots:}\\[-0.3ex]
\begin{itemize}[leftmargin=1.6em,itemsep=0.15ex,topsep=0.2ex]
    \item \textbf{Shot 1} [0.0--0.6]: static outdoor medium shot.
    \item \textbf{Shot 2} [0.6--3.2]: close-up high-angle view into an open wooden hive.
    \item \textbf{Shot 3} [3.2--5.5]: indoor medium shot with \texttt{PERSON\_1} speaking.
    \item \textbf{Shot 4} [5.5--8.1]: close-up of the top mechanism of the object.
    \item \textbf{Shot 5} [8.1--10.0]: extreme close-up of a white plastic gate valve.
\end{itemize}

\textbf{Audio events:}\\[-0.3ex]
\begin{itemize}[leftmargin=1.6em,itemsep=0.15ex,topsep=0.2ex]
    \item \textbf{DIALOGUE\_1} [0.0--9.8]: \texttt{PERSON\_1} speaks in an instructional tone.
    \item \textbf{SFX\_1} [5.5--8.0]: mechanical whirring and rhythmic geared turning.
\end{itemize}

\textbf{Ground-truth event-shot links:} DIALOGUE\_1 $\rightarrow$ Shots 1--5; SFX\_1 $\rightarrow$ Shot 4.
\vspace{0.6ex}

\textbf{Model predictions:}\\[-0.3ex]
\begin{itemize}[leftmargin=1.6em,itemsep=0.2ex,topsep=0.2ex]
    \item \textbf{Gemini 3.1-Pro} (F1=1.000): all six links are correct, including SFX\_1 $\rightarrow$ Shot 4.
    \item \textbf{Gemini 2.5-Pro} (F1=0.909): all dialogue links are correct, but SFX\_1 $\rightarrow$ Shot 4 is missed.
    \item \textbf{Qwen3.5-Omni-Plus} (F1=0.909): all dialogue links are correct, but SFX\_1 $\rightarrow$ Shot 4 is missed.
    \item \textbf{Qwen3.5-Omni-Flash} (F1=0.727): detects SFX\_1 $\rightarrow$ Shot 4, but adds the false link SFX\_1 $\rightarrow$ Shot 3 and misses DIALOGUE\_1 $\rightarrow$ Shots 4--5.
    \item \textbf{Qwen3-Omni-Instruct} (F1=0.800): keeps DIALOGUE\_1 $\rightarrow$ Shots 1--4, but misses DIALOGUE\_1 $\rightarrow$ Shot 5 and SFX\_1 $\rightarrow$ Shot 4.
    \item \textbf{Qwen3-Omni-Captioner} (F1=0.800): follows the same omission pattern as Qwen3-Omni-Instruct, missing the final dialogue shot and the mechanical SFX link.
\end{itemize}

\vspace{0.8ex}
\hrule
\vspace{0.6ex}
\textit{This example separates two EVSA failure modes. Gemini 2.5-Pro and Qwen3.5-Omni-Plus preserve the long dialogue-to-shot coverage but omit the localized mechanical sound link. Qwen3.5-Omni-Flash instead detects the mechanical sound link but over-extends it to a neighboring shot and loses later dialogue coverage. These errors are small graph edits, but they change which visual evidence is considered to support each audio event.}
\end{tcolorbox}

This case study shows why EVSA is evaluated as explicit event-shot edges rather than as part of a holistic caption score. The dialogue event spans almost the full clip and should therefore be linked to all five shots. This is an easy relation to hide in prose, because a caption can simply state that a person is explaining the process throughout the video. EVSA makes the requirement explicit: each covered shot must be represented as a separate verifiable edge. The mechanical sound is more localized. It occurs from 5.5 to 8.0 seconds and is supported by Shot 4, so missing SFX\_1 $\rightarrow$ Shot 4 indicates an omission of a short cross-modal relation, while predicting SFX\_1 $\rightarrow$ Shot 3 indicates temporal over-extension. The six model traces therefore expose three distinct behaviors: Gemini 3.1-Pro recovers the full graph, Gemini 2.5-Pro and Qwen3.5-Omni-Plus miss the localized SFX edge, and the remaining Qwen variants either truncate the long dialogue coverage or add an unsupported neighboring SFX edge.

Taken together, CCC and EVSA reveal two complementary forms of structured video understanding. CCC tests whether a visual entity remains the same node across discontinuous shots; EVSA tests whether an audio event is attached to the correct visual evidence in the timeline. Both metrics convert a fluent caption into checkable graph constraints. The resulting insight is that current models can often describe local content, but still fail when the evaluation asks whether references and events are consistently bound across time and modality. This is precisely the failure mode that a global caption-quality score would tend to smooth over.

\subsection{Analysis of Specialized Captioning Models}
\label{app:specialized_models_failure}

As stated in Section~\ref{subsec:exp_setup}, we purposefully excluded highly specialized video-captioning models (such as ASID-Caption~\citep{li2026towards}, AVoCaDO~\citep{chen2025avocado}, TimeChat-Captioner~\citep{yao2026timechatcaptioner}, and UGC-VideoCaptioner~\citep{wu2025ugcvideocaptioner}) from the primary capability benchmarking. This exclusion is not because these models are incapable of perceiving videos, but because their rigorous supervised fine-tuning (SFT) heavily biases them toward generating free-form, holistic text paragraphs.

This intensive, domain-specific linguistic alignment degrades their general instruction-following capabilities in ways that are visible before any semantic scoring is applied. When required to output the deep-structured atomic units of \name{} (i.e., a structured JSON with Reference, Event, and Shot arrays), representative outputs show four recurring failure modes, separated into instruction-following failures in Figure~\ref{fig:sft_instruction_failures} and structural-output failures in Figure~\ref{fig:sft_structural_failures}:

\begin{itemize}[leftmargin=*]
    \item \textbf{Schema rejection:} ASID-Caption models often ignore the JSON contract, emitting timestamped prose. In one 17.1-second clip, the raw response repeats its description template until \texttt{At 1037s}, while the parsed \texttt{references}, \texttt{events}, and \texttt{shots} arrays remain empty (Figure~\ref{fig:sft_instruction_failures}(a)).
    \item \textbf{Fixed caption schema:} TimeChat-Captioner does produce machine-readable JSON, but it follows its own captioning template with fields such as \texttt{timestamp}, \texttt{segment\_detail\_caption}, \texttt{camera\_state}, \texttt{storyline}, and \texttt{shooting\_style}. These fields are fluent but do not instantiate the \name{} evaluation graph: there are no persistent reference IDs, no event objects, and no shot-level cross-modal links (Figure~\ref{fig:sft_instruction_failures}(b)).
    \item \textbf{Incomplete graph construction:} AVoCaDO superficially follows the requested key names, producing two references and two shots, but leaves \texttt{events} empty, assigns repeated zero-length \texttt{sub\_shot\_time} spans, and links shots to an undefined \texttt{SCENE\_1}. The resulting object is a caption-like shot summary rather than a connected Reference--Event--Shot graph (Figure~\ref{fig:sft_structural_failures}(a)).
    \item \textbf{Degenerate structured output:} UGC-VideoCaptioner also emits target keys, but the structure is not usable. It repeatedly assigns \texttt{sub\_shot\_time=[0.0,0.0]} and restates near-identical subject facts such as \texttt{[PERSON\_1] is wearing a white shirt and blue jeans} or \texttt{[PERSON\_1] is holding a metal tool to remove honeycomb frames} (Figure~\ref{fig:sft_structural_failures}(b)).
\end{itemize}

These failures occur at the interface between instruction following and structural generation, not at the final semantic-matching stage. The outputs either cannot be parsed into \name{} units, follow a model-specific caption template, or collapse the temporal event graph into disconnected, repeated fragments. Reporting downstream structural scores would therefore conflate video perception with schema compliance; we instead treat these traces as evidence that caption-specialized SFT does not provide the controllable, graph-structured generation required by \name{}.

\begin{figure*}[htbp]
    \centering
    \begin{subfigure}[t]{0.485\linewidth}
        \begin{tcolorbox}[
            colback=red!4,
            colframe=table_red,
            boxrule=0.6pt,
            title={\textbf{(a) Schema rejected: prose loop}},
            fonttitle=\small,
            left=5pt,
            right=5pt,
            top=5pt,
            bottom=5pt,
            height=5.85cm,
            valign=top
        ]
        {\scriptsize\ttfamily
        model: asid\_caption\_7b\\
        video duration: 17.1s; shots: 8\\
        parsed target units: refs=0, events=0, shots=0\\[5pt]
        At 0s, the video opens in a dimly lit room...\\
        At 8s, ... a slight, knowing smirk...\\
        At 16s, ... finger still pointed at the bald man's face.\\[2pt]
        \textcolor{table_red}{At 1037s, the same template is still being generated.}\\
        \textcolor{table_red}{No JSON object or Reference--Event--Shot arrays are produced.}
        }
        \end{tcolorbox}
        \caption{ASID-Caption ignores the target schema and continues a repetitive timestamped prose loop far beyond the 17.1-second video.}
    \end{subfigure}
    \hfill
    \begin{subfigure}[t]{0.485\linewidth}
        \begin{tcolorbox}[
            colback=orange!6,
            colframe=orange!80!black,
            boxrule=0.6pt,
            title={\textbf{(b) Instruction obeyed only superficially}},
            fonttitle=\small,
            left=5pt,
            right=5pt,
            top=5pt,
            bottom=5pt,
            height=5.85cm,
            valign=top
        ]
        {\scriptsize\ttfamily
        model: timechat\_captioner\\
        video duration: 10.0s; model emits segments to 54s\\
        parsed target units: refs=0, events=0, shots=0\\[5pt]
        [\{ ``timestamp'': ``00:00-00:05'',\\
        \quad ``segment\_detail\_caption'': ``Two men ...'',\\
        \quad ``camera\_state'': ``high angle ...'',\\
        \quad ``storyline'': ``The video begins ...'',\\
        \quad ``shooting\_style'': ``establishing shot'' \}, ...]\\[2pt]
        \textcolor{table_red}{Valid caption template; no ref\_id, event\_id, or shot graph.}
        }
        \end{tcolorbox}
        \caption{TimeChat-Captioner returns its fixed dense-caption format instead of the requested \name{} unit schema.}
    \end{subfigure}
    \caption{\textbf{Instruction-following failures induced by caption-specialized SFT.} The two models either reject the requested JSON contract outright or satisfy only the generic request for machine-readable captions while missing the required Reference--Event--Shot structure.}
    \label{fig:sft_instruction_failures}
\end{figure*}

\begin{figure*}[htbp]
    \centering
    \begin{subfigure}[t]{0.485\linewidth}
        \begin{tcolorbox}[
            colback=blue!4,
            colframe=blue!65!black,
            boxrule=0.6pt,
            title={\textbf{(a) Target keys, disconnected graph}},
            fonttitle=\small,
            left=5pt,
            right=5pt,
            top=5pt,
            bottom=5pt,
            height=6.25cm,
            valign=top
        ]
        {\scriptsize\ttfamily
        model: avocado\\
        video duration: 24.1s; shots: 4\\
        parsed target units: refs=2, events=0, shots=2\\[5pt]
        references: PERSON\_1, PERSON\_2\\
        events: [ ]\\
        shot\_1.active\_events: [ ]\\
        shot\_1.references\_in\_shot: [PERSON\_1, PERSON\_2, SCENE\_1]\\[2pt]
        sub\_shot\_time: [0.0, 0.0] repeated inside shot\_1\\
        \textcolor{table_red}{SCENE\_1 is linked but never defined; actions are prose, not events.}
        }
        \end{tcolorbox}
        \caption{AVoCaDO uses several target field names, but leaves the event layer empty and produces disconnected shot summaries with undefined references and zero-length sub-shot spans.}
    \end{subfigure}
    \hfill
    \begin{subfigure}[t]{0.485\linewidth}
        \begin{tcolorbox}[
            colback=yellow!8,
            colframe=orange!80!black,
            boxrule=0.6pt,
            title={\textbf{(b) Pseudo-structure degenerates}},
            fonttitle=\small,
            left=5pt,
            right=5pt,
            top=5pt,
            bottom=5pt,
            height=6.25cm,
            valign=top
        ]
        {\scriptsize\ttfamily
        model: ugc\_videocaptioner\\
        video duration: 10.0s; shots: 4\\
        parsed target units: refs=1, events=1, shots=1\\[5pt]
        sub\_shot\_time: [0.0, 0.0]\\
        {}[PERSON\_1] is wearing a white shirt and blue jeans.\\
        sub\_shot\_time: [0.0, 0.0]\\
        {}[PERSON\_1] is holding a metal tool to remove honeycomb frames.\\
        sub\_shot\_time: [0.0, 0.0]\\
        {}[PERSON\_1] is pointing at the honeycomb frame.\\[2pt]
        \textcolor{table_red}{The same zero-time subject facts repeat instead of local events.}
        }
        \end{tcolorbox}
        \caption{UGC-VideoCaptioner produces target-looking JSON, but temporal grounding collapses and the visual descriptions repeat the same subject attributes and actions.}
    \end{subfigure}
    \caption{\textbf{Structural-output failures after partial schema compliance.} Even when caption-specialized models emit \name{}-like keys, the generated objects do not form usable evaluation graphs: events are missing, references can be undefined, and local descriptions collapse into repeated zero-duration fragments.}
    \label{fig:sft_structural_failures}
\end{figure*}

\begin{table}[t]
\centering
\caption{\textbf{The illusion of global text scores.} We construct two hard-to-distinguish subsets where a weaker baseline and a stronger upgraded model exhibit near-identical text-centric scores. Across these pairs, we track performance through progressive strictness: (1) traditional text-centric metrics on dense captions, (2) strict structural verification using our native atomic units, and (3) human audit. Global text metrics falsely penalize the stronger models and overestimate the weaker ones, creating an illusion of parity. However, evaluating structural constraints directly resolves this gap, perfectly aligning with human judgment.}
\small
\resizebox{0.85\linewidth}{!}{
\begin{tabular}{lccc}
\toprule
\textbf{Model}
& \textbf{Text Metric (Prose)}
& \textbf{Native Constraints}
& \textbf{Human Audit} \\
\midrule
\multicolumn{4}{l}{\textit{Pair 1: Qwen3.5-Omni-Flash vs.\ Qwen3-Omni}} \\
\midrule
Weaker (Qwen3-Omni)         & 52.2  & 37.5  & 21.6 \\
Stronger (Qwen3.5-Omni-Flash) & 47.8  & 62.5  & 78.4\\
\midrule
\multicolumn{4}{l}{\textit{Pair 2: Qwen3.5-Omni-Plus vs.\ Qwen3.5-Omni-Flash}} \\
\midrule
Weaker (Qwen3.5-Omni-Flash) & 41.7  & 21.4  & 29.9 \\
Stronger (Qwen3.5-Omni-Plus)  & 58.3 & 78.6  & 70.1 \\
\midrule
\midrule
\textbf{Average (Stronger Models)}     & 53.0 & 70.6  &  74.2 \\
\textbf{Average (Weaker Models)}     & 47.0 &   29.4   &  25.8 \\
\bottomrule
\end{tabular}}
\label{tab:caption_state_gap_result}
\vspace{-2mm}
\end{table}

\subsection{Post-hoc Parsing Baseline}
\label{app:posthoc_parser}

Appendix~\ref{app:specialized_models_failure} shows that prose-optimized captioners cannot emit \name{} units directly, which raises a fair question: do their low scores measure perception, or only instruction following? To separate the two, we add a fixed \emph{post-hoc} track. A single LLM parser receives the model's free-form caption and never the video, converts that caption into \name{} units, and marks unsupported fields as \texttt{unknown} (Listing B.2). The parser is identical for every model, so it contributes no visual evidence and no per-model tuning.

Tables~\ref{tab:posthoc_rule} and~\ref{tab:posthoc_llm} show that this route buys format compliance, not information. Structural well-formedness rises: SGC improves from 97.36 to 99.19 for Gemini 3.1-Pro, and Shot F1 and Shot tIoU exceed 95 for every post-hoc entry, because a text-only parser emits clean, non-overlapping segments by construction. Every metric that requires grounding to the video degrades at the same time. Gemini's CCC falls from 34.43 to 5.20 and its EVSA from 51.46 to 38.29, Qwen3-Omni-Instruct's Event F1 falls from 41.70 to 12.82, and the localized semantic scores drop across the board, for instance Gemini's Shot score from 59.05 to 35.84. This joint pattern is itself a diagnostic result: near-saturated shot metrics paired with single-digit CCC is precisely the profile that a global caption score hides and that our deep-structured units are built to expose.

The post-hoc track also makes specialized captioners evaluable instead of simply scoring them zero. ASID-Captioner-7B now produces valid units and reaches 96.17 SGC, yet stays at 9.65 CCC, 11.09 Event F1, and 10.87 EVSA. Format repair therefore widens the benchmark's coverage, but the cross-shot identity and audio-visual relations that \name{} targets are absent from the original dense captions and cannot be reconstructed after the fact.

\begin{table*}[!t]
\centering
\caption{\textbf{Native generation vs.\ the post-hoc parsing track (rule-based metrics).} \emph{Native} asks the model for \name{} units directly; \emph{Post-hoc} lets the model write a free-form caption that a fixed, video-blind LLM parser converts into units. Post-hoc raises structural well-formedness (SGC, Shot F1, Shot tIoU) while every grounding-dependent metric (CCC, Event F1, EVSA) degrades, showing that format repair does not recover information the caption never contained.}
\label{tab:posthoc_rule}
\scriptsize
\setlength{\tabcolsep}{3pt}
\resizebox{\linewidth}{!}{
\begin{tabular}{@{}ll cccc cc cc cc@{}}
\toprule
\textbf{Model} & \textbf{Track} & \textbf{SGC} & \textbf{RefUse} & \textbf{CCC} & \textbf{Ref Subj} & \textbf{Shot F1} & \textbf{Shot tIoU} & \textbf{Evt F1} & \textbf{Evt tIoU} & \textbf{Speaker} & \textbf{EVSA} \\
\midrule
\multirow{2}{*}{Gemini 3.1-Pro~\citep{gemini2024}} & Native & 97.36 & 70.39 & 34.43 & 78.60 & 76.24 & 70.74 & 63.32 & 71.92 & 91.18 & 51.46 \\
 & Post-hoc & 99.19 & 46.91 & 5.20 & 57.11 & 97.39 & 95.62 & 40.73 & 56.65 & 87.36 & 38.29 \\
\midrule
\multirow{2}{*}{Qwen3-Omni-Instruct~\citep{xu2025qwen3omni}} & Native & 89.02 & 40.96 & 9.85 & 54.84 & 49.98 & 67.92 & 41.70 & 40.99 & 88.35 & 13.33 \\
 & Post-hoc & 96.61 & 40.93 & 9.62 & 49.21 & 96.44 & 95.59 & 12.82 & 54.40 & 81.65 & 12.23 \\
\midrule
ASID-Captioner-7B~\citep{li2026towards} & Post-hoc & 96.17 & 39.58 & 9.65 & 41.89 & 96.51 & 95.73 & 11.09 & 54.83 & 87.85 & 10.87 \\
\bottomrule
\end{tabular}
}
\end{table*}

\begin{table*}[!t]
\centering
\caption{\textbf{Native generation vs.\ the post-hoc parsing track (localized LLM-judge metrics).} Every local semantic score drops under post-hoc conversion, including for the caption-specialized model that the track is meant to accommodate. Precision rises on some fields only because the parser emits fewer, more conservative units.}
\label{tab:posthoc_llm}
\scriptsize
\setlength{\tabcolsep}{3pt}
\resizebox{\linewidth}{!}{
\begin{tabular}{@{}ll c cc cc cc cc@{}}
\toprule
\textbf{Model} & \textbf{Track} & \textbf{Shot} & \textbf{Subj-R} & \textbf{Subj-P} & \textbf{Scene-R} & \textbf{Scene-P} & \textbf{Subsh-R} & \textbf{Subsh-P} & \textbf{Dialogue} & \textbf{Non-dial} \\
\midrule
\multirow{2}{*}{Gemini 3.1-Pro~\citep{gemini2024}} & Native & 59.05 & 43.78 & 67.07 & 51.78 & 75.62 & 53.29 & 70.92 & 91.18 & 41.49 \\
 & Post-hoc & 35.84 & 24.75 & 75.90 & 24.72 & 82.59 & 36.15 & 62.94 & 87.36 & 21.42 \\
\midrule
\multirow{2}{*}{Qwen3-Omni-Instruct~\citep{xu2025qwen3omni}} & Native & 34.63 & 40.31 & 62.02 & 36.07 & 70.73 & 29.42 & 57.84 & 88.35 & 20.51 \\
 & Post-hoc & 28.67 & 28.49 & 71.82 & 29.32 & 77.37 & 33.77 & 50.75 & 81.65 & 13.15 \\
\midrule
ASID-Captioner-7B~\citep{li2026towards} & Post-hoc & 22.87 & 25.24 & 65.66 & 32.96 & 64.35 & 33.25 & 40.17 & 87.85 & 10.37 \\
\bottomrule
\end{tabular}
}
\end{table*}

\subsection{Threshold Sensitivity}
\label{app:threshold_sensitivity}

\name{} fixes four matching cutoffs (Table~\ref{tab:matching_rules}): non-dialogue event tIoU $\ge 0.20$, shot tIoU $\ge 0.30$, subshot window $\delta = 1.0$\,s, and dialogue $1-\mathrm{WER} \ge 0.50$. Since these values decide which predictions count as matched, we test whether our conclusions survive perturbation. We vary one cutoff at a time over the grid in Table~\ref{tab:threshold_sweep} for the six models listed there, holding predictions, references, and the one-to-one matching rule fixed, so no inference is re-run and only the alignment decision changes.

Absolute scores move as expected---tightening shot tIoU from 0.30 to 0.50 costs Gemini 3.1-Pro 12.2 Shot F1 points---but the induced ranking does not. Spearman's $\rho$ against the default, averaged over the two alternatives per sweep, is 1.000 for every metric except Shot F1 in the shot sweep (0.971) and Speaker in the dialogue sweep (0.914). No sweep produces a systematic open- versus closed-source reversal, and the only top-1 change within this set is a near tie on Speaker between Gemini 3.1-Pro (91.18) and Qwen3.5-Omni-Plus (91.20), two proprietary models 0.02 points apart. The structural bottlenecks we report, CCC and EVSA, are therefore properties of the evaluated models rather than artifacts of a particular cutoff.

\begin{table*}[!t]
\centering
\caption{\textbf{Matching-threshold sweep.} Each block varies one cutoff while predictions, references, and the one-to-one matching rule stay fixed; no inference is re-run. Default settings are shown in \textbf{bold}. Absolute scores shift with the cutoff, but the induced ranking does not: Spearman's $\rho$ against the default ranges from 0.914 to 1.000 (Appendix~\ref{app:threshold_sensitivity}).}
\label{tab:threshold_sweep}
\scriptsize
\setlength{\tabcolsep}{4pt}
\resizebox{\linewidth}{!}{
\begin{tabular}{@{}llc cccccc@{}}
\toprule
\textbf{Sweep} & \textbf{Metric} & \textbf{Setting} & \shortstack{\textbf{Gemini}\\\textbf{3.1-Pro}} & \shortstack{\textbf{Gemini}\\\textbf{2.5-Pro}} & \shortstack{\textbf{Qwen3.5-}\\\textbf{Omni-Plus}} & \shortstack{\textbf{Qwen3.5-}\\\textbf{Omni-Flash}} & \shortstack{\textbf{Qwen3-Omni-}\\\textbf{Instruct}} & \shortstack{\textbf{Qwen3-Omni-}\\\textbf{Captioner}} \\
\midrule
\multirow{3}{*}{Non-dialogue tIoU} & \multirow{3}{*}{Event F1} & 0.10 & 64.02 & 59.54 & 54.63 & 51.43 & 42.29 & 41.88 \\
 & & \textbf{0.20} & 63.32 & 58.52 & 53.92 & 50.68 & 41.70 & 41.31 \\
 & & 0.40 & 61.86 & 56.34 & 52.38 & 48.95 & 40.71 & 40.07 \\
\midrule
\multirow{3}{*}{Shot tIoU} & \multirow{3}{*}{Shot F1} & 0.20 & 77.84 & 72.32 & 72.96 & 67.81 & 54.22 & 55.34 \\
 & & \textbf{0.30} & 76.24 & 70.93 & 71.25 & 65.65 & 49.98 & 52.29 \\
 & & 0.50 & 64.07 & 60.87 & 60.37 & 52.79 & 37.41 & 38.82 \\
\midrule
\multirow{3}{*}{Subshot window $\delta$} & \multirow{3}{*}{Subshot F1} & 0.5\,s & 64.64 & 66.07 & 64.79 & 57.91 & 29.18 & 38.89 \\
 & & \textbf{1.0\,s} & 65.27 & 66.69 & 65.69 & 58.94 & 29.51 & 39.51 \\
 & & 2.0\,s & 65.90 & 67.35 & 66.38 & 59.92 & 29.78 & 40.17 \\
\midrule
\multirow{3}{*}{Dialogue $1-\mathrm{WER}$} & \multirow{3}{*}{Speaker} & 0.40 & 89.92 & 88.10 & 89.16 & 87.54 & 85.82 & 86.64 \\
 & & \textbf{0.50} & 91.18 & 89.72 & 91.20 & 89.91 & 88.35 & 88.68 \\
 & & 0.60 & 94.08 & 92.57 & 93.88 & 92.84 & 91.46 & 91.62 \\
\bottomrule
\end{tabular}
}
\vspace{-2mm}
\end{table*}

\section{Limitations and Broader Impact}
\label{app:limitations}
\label{app:broader_impact}

Scoring atomic units rather than holistic text carries inherent limitations. First, \name{} evaluates structural fidelity against our annotated reference system $\mathcal{S}^\star(V)$, not \emph{every} true fact in a video. A valid but peripheral detail outside our annotation scope therefore goes uncredited, a limitation shared by many generative benchmarks. Second, the dataset is bounded to videos under five minutes; far longer videos may require identity-tracking architectures beyond our scope. Third, the Stage~II drafts seeding our reference system come from frontier models, some in the families we later evaluate, so an evaluator--model bias cannot be excluded: a reference inherits whatever blind spot survived review, and a model sharing the drafter's stylistic priors may be scored slightly generously. We mitigate this by drafting with several model families, letting deterministic validators and human annotators arbitrate every unit (Appendix~\ref{app:benchmark_construction}), and restricting the scoring contract to locally verifiable fields, but we do not claim the bias is fully eliminated.

Finally, enforcing a strict, deep-structured JSON schema inherently disadvantages models heavily over-optimized for free-form prose (as discussed in Appendix~\ref{app:specialized_models_failure}). This is a deliberate choice intended to encourage the research community to shift toward controllable, structured video generation rather than opaque text generation. \name{}'s broader impact is providing developers with a highly auditable, traceable diagnostic tool, ensuring that future improvements in MLLMs represent genuine audio-visual reasoning rather than mere linguistic fluency.

\section{Ethical Considerations}
\label{app:ethical_considerations}

Constructing and releasing \name{} raises ethical considerations around privacy, copyright, and the interpretation of diagnostic evaluations.

\textbf{Privacy and Consent:} The benchmark curates videos from publicly accessible media platforms and existing academic datasets. Since video intrinsically contains faces, voices, and potentially sensitive personal expressions, we filter out harmful, explicit, or highly sensitive private content during validation. Researchers using \name{} must strictly adhere to the privacy guidelines of the original source platforms.

\textbf{Copyright and Fair Use:} \name{} respects intellectual property rights by not distributing raw video files. Instead, we distribute the structured annotations alongside deterministic download scripts pointing to the original public URLs. This keeps the benchmark within academic fair use while respecting creators' distribution rights.

\textbf{Evaluation Bias and Capabilities:} Our benchmark is explicitly designed to measure structured audio-visual understanding. Enforcing a strict, rule-based schema intrinsically disadvantages models optimized solely for free-form prose generation. It is ethically imperative to emphasize that \name{} measures a specific, rigorous form of grounded reasoning. A low score on \name{} highlights localized structural deficits, but it does not necessarily imply that a model is entirely incapable of generating helpful video summaries in more relaxed, open-ended conversational settings.

\section{Dataset Licenses and Usage}
\label{app:dataset_licenses}

\name{} is constructed under the MIT License, which permits free use, modification, and distribution for academic and commercial purposes. For the underlying video files, users must comply with the original licenses of the source datasets, which generally restrict usage to non-commercial research purposes. To respect copyright boundaries, we do not host or distribute the raw MP4 video files directly. Instead, we provide a deterministic script to download the publicly available videos using their original URLs.
Table~\ref{tab:dataset_licenses} details the specific licenses for the underlying video data sources and the software models utilized during our evaluation.

\begin{table}[htbp]
\centering
\caption{\textbf{License Information for Scientific Artifacts.} This table details the licenses for the underlying video data sources and software models used in our evaluation.}
\label{tab:dataset_licenses}
\small
\setlength{\tabcolsep}{4pt}
\resizebox{\linewidth}{!}{
\begin{tabular}{@{}llc@{}}
\toprule
\textbf{Data Sources} & \textbf{URL} & \textbf{License} \\
\midrule
FunQA & \url{https://github.com/Jingkang50/FunQA} & MIT \\
LLaVA-Video-178K & \url{https://huggingface.co/datasets/lmms-lab/LLaVA-Video-178K} & Apache-2.0 \\
Tarsier2-Recap & \url{https://huggingface.co/datasets/omni-research/Tarsier2-Recap-585K} & Research Only \\
Video-MME & \url{https://github.com/MME-Benchmarks/Video-MME} & Research Only \\
VisionRewardDB-Video & \url{https://huggingface.co/datasets/zai-org/VisionRewardDB-Video} & Apache-2.0 \\
LongVideoBench & \url{https://huggingface.co/datasets/longvideobench/LongVideoBench} & CC BY-NC-SA 4.0 \\
OmniVideoBench & \url{https://huggingface.co/datasets/NJU-LINK/OmniVideoBench} & CC BY-NC-SA 4.0 \\
\midrule
\textbf{Software / Models} & \textbf{URL} & \textbf{License} \\
\midrule
Gemini Models & \url{https://ai.google.dev/} & Google ToS \\
Qwen Models & \url{https://github.com/QwenLM/Qwen2.5-Omni} & Apache-2.0 \\
Seed Models & \url{https://seed.bytedance.com/} & ByteDance ToS \\
MiMo Models & \url{https://huggingface.co/XiaomiMiMo/MiMo-V2.5} & MIT \\
MiniCPM-o Models & \url{https://huggingface.co/openbmb/MiniCPM-o-2_6} & Apache-2.0 \\
\bottomrule
\end{tabular}
}
\end{table}

\section{Instruction Templates and Prompts}
\label{app:instruction_templates}

\begin{tcolorbox}[
    enhanced,
    colback=uclablue!8,
    colframe=uclablue,
    colbacktitle=uclablue,
    coltitle=black,
    fonttitle={\bfseries\sffamily\small},
    title={\textbf{Listing B.1:} Zero-shot Generation Instruction Example for predicting \name{} units.},
    boxrule=1pt,
    arc=4pt,
    drop fuzzy shadow
]
\begin{lstlisting}[style=chattemplate, language=Python, breaklines=true]
You are an expert audio-visual video analyzer. Watch the video carefully and decompose its contents into a strict JSON structure containing three tracks: `references`, `events`, and `shots`.

1. `references`: List all persistent entities (PERSON, OBJECT, SCENE). Give each a unique `ref_id` (e.g., PERSON_1). Include a `semantic_description` and an `appearance_anchor` detailing their visual features.
2. `events`: List all audio events. Give each an `event_id`, a `time_range` [start, end] in seconds, and `content`. If the event is spoken dialogue, include the exact transcribed `line`.
3. `shots`: Divide the video into non-overlapping temporal shots. For each shot, provide its `time_range`, `visual_description`, and lists of `active_events` (referencing event_ids) and `references_in_shot` (referencing ref_ids).

Output ONLY valid JSON matching this schema. Do not include markdown formatting or conversational text.
\end{lstlisting}
\end{tcolorbox}

\begin{tcolorbox}[
    enhanced,
    colback=uclablue!8,
    colframe=uclablue,
    colbacktitle=uclablue,
    coltitle=black,
    fonttitle={\bfseries\sffamily\small},
    title={\textbf{Listing B.2:} Post-hoc Parser Instruction Example used to extract units from dense captions.},
    boxrule=1pt,
    arc=4pt,
    drop fuzzy shadow
]
\begin{lstlisting}[style=chattemplate, language=Python, breaklines=true]
You are a structural parser. Given a free-form dense video caption, extract all implicit facts and format them into the explicit MTSS JSON schema.
- Identify all mentioned characters/objects and assign them consistent IDs in the `references` list.
- Extract any mentioned sounds or dialogue into the `events` list, attempting to infer timestamps if mentioned.
- Break the narrative into temporal `shots`, linking the extracted references and events. If a detail (like an exact timestamp) is missing from the text, omit it or use a best-guess estimate.

Caption to parse: "{caption_text}"
\end{lstlisting}
\end{tcolorbox}

\begin{tcolorbox}[
    enhanced,
    colback=uclablue!8,
    colframe=uclablue,
    colbacktitle=uclablue,
    coltitle=black,
    fonttitle={\bfseries\sffamily\small},
    title={\textbf{Listing B.3:} Local Semantic Judge Instruction Examples (Subshot Matching and Quality).},
    boxrule=1pt,
    arc=4pt,
    drop fuzzy shadow
]
\begin{lstlisting}[style=chattemplate, language=Python, breaklines=true]
_SUBSHOT_MATCH_SYSTEM_PROMPT: You are a strict binary matcher for video local visual states. Decide whether the TARGET item has at least one true counterpart in the opposite-side candidate list. Use only the provided sub_shot_content, times, ids, and reference mapping. A match must share the main visible subject(s), action/pose/state, and local scene facts. Accept paraphrases and small granularity differences; reject generic overlap, nearby-but-different actions, or contradictory core facts. Do not evaluate hallucination or extra details separately; decide only whether the target's core local facts are covered by at least one candidate. Output exactly <match>1</match> for a match or <match>0</match> for no match, followed by one short reason.

_SUBSHOT_QUALITY_SYSTEM_PROMPT: You are an evidence-based evaluator for local video-state fact coverage. Compare the TARGET local-state facts against the provided matched context and score 1-5. Do not evaluate hallucination or extra details as a separate criterion; score only whether the requested TARGET facts are covered. Ignore minor wording differences and neighboring context that is not relevant to the target. Output format: first write <score>N</score> (N=1-5), then 2-3 short evidence bullets naming the hit/missed target facts.
\end{lstlisting}
\end{tcolorbox}

\newpage
\begin{tcolorbox}[
    enhanced,
    colback=uclablue!8,
    colframe=uclablue,
    colbacktitle=uclablue,
    coltitle=black,
    fonttitle={\bfseries\sffamily\small},
    title={\textbf{Listing B.4:} Generic Local 1--5 Judge System Instruction.},
    boxrule=1pt,
    arc=4pt,
    drop fuzzy shadow
]
\begin{lstlisting}[style=chattemplate, language=Python, breaklines=true]
_JUDGE_SYSTEM_PROMPT: You are a strict, evidence-based evaluator for video description quality. Your job is to compare a PREDICTION against a GROUND TRUTH and score 1-5. Use only the fields shown in the instruction; do not infer from outside knowledge or missing video context.

RULES:
1. Judge factual consistency and coverage of the requested field only.
2. Contradictions to the GT are severe errors; missing details are less severe than wrong details.
3. Ignore wording/order differences when the same fact is expressed. 4. Do not evaluate hallucination as a separate criterion. Ignore facts outside the requested coverage checklist; score only factual coverage of the requested field. 5. Output format: first write <score>N</score> (N=1-5), then 2-3 concise evidence bullets naming the key matched/missed/wrong facts.
\end{lstlisting}
\end{tcolorbox}

\begin{tcolorbox}[colback=uclablue!8, colframe=uclablue, colbacktitle=uclablue, coltitle=black, boxrule=1pt, title={\textbf{System-Level Local Judging Instruction}}]
\begin{lstlisting}[style=chattemplate, language=Python, breaklines=true]
_JUDGE_SYSTEM_PROMPT: You are a strict, evidence-based evaluator for video description quality. Your job is to compare a PREDICTION against a GROUND TRUTH and score 1-5. Use only the fields shown in the instruction; do not infer from outside knowledge or missing video context.

RULES:
1. Judge factual consistency and coverage of the requested field only.
2. Contradictions to the GT are severe errors; missing details are less severe than wrong details.
3. Ignore wording/order differences when the same fact is expressed. 4. Do not evaluate hallucination as a separate criterion. Ignore facts outside the requested coverage checklist; score only factual coverage of the requested field. 5. Output format: first write <score>N</score> (N=1-5), then 2-3 concise evidence bullets naming the key matched/missed/wrong facts.

_SUBSHOT_MATCH_SYSTEM_PROMPT: You are a strict binary matcher for video local visual states. Decide whether the TARGET item has at least one true counterpart in the opposite-side candidate list. Use only the provided sub_shot_content, times, ids, and reference mapping. A match must share the main visible subject(s), action/pose/state, and local scene facts. Accept paraphrases and small granularity differences; reject generic overlap, nearby-but-different actions, or contradictory core facts. Do not evaluate hallucination or extra details separately; decide only whether the target's core local facts are covered by at least one candidate. Output exactly <match>1</match> for a match or <match>0</match> for no match, followed by one short reason.

_SUBSHOT_QUALITY_SYSTEM_PROMPT: You are an evidence-based evaluator for local video-state fact coverage. Compare the TARGET local-state facts against the provided matched context and score 1-5. Do not evaluate hallucination or extra details as a separate criterion; score only whether the requested TARGET facts are covered. Ignore minor wording differences and neighboring context that is not relevant to the target. Output format: first write <score>N</score> (N=1-5), then 2-3 short evidence bullets naming the hit/missed target facts.
\end{lstlisting}
\end{tcolorbox}

\newpage
\begin{tcolorbox}[colback=uclablue!8, colframe=uclablue, colbacktitle=uclablue, coltitle=black, boxrule=1pt, title={\textbf{Reference Recall Instruction Example}}]
\begin{lstlisting}[style=chattemplate, language=Python, breaklines=true]
Entity type: {gt_type}

TASK: Evaluate RECALL — using the GROUND TRUTH as the subject, check whether the PREDICTION covers ALL factual content present in the GT detail_description. Do not evaluate hallucination or extra prediction details; only score coverage of GT facts.

Compare only appearance_anchor.id_features.detail_description. Ignore semantic_description, attributes, timestamp, events, shots, and context.

{field_guidance}

Scoring criteria (recall perspective — does prediction cover GT facts?):
- Count the total number of distinct factual elements in the GT description.
- For each GT fact, check if the prediction mentions it (same meaning, wording differences OK).
- A GT fact is 'missed' if the prediction does not mention it at all.
- A GT fact is 'contradicted' if the prediction states the opposite (worse than missing).

=== GROUND TRUTH detail_description (SUBJECT — all facts here should be covered) === {detail_description(gt_ref)}

=== PREDICTION detail_description (check coverage of GT facts) === {detail_description(pred_ref)}

Score 1-5:
  1 = Prediction covers almost none of the GT facts (<20%), or contradicts the core identity
  2 = Prediction covers a minority of GT facts (20-40%); multiple key attributes are missing
  3 = Prediction covers roughly half of GT facts (40-70%); some important details are missing
  4 = Prediction covers most GT facts (70-90%); only minor details are missing
  5 = Prediction covers essentially all GT facts (>90%); at most trivial omissions
\end{lstlisting}
\end{tcolorbox}

\begin{tcolorbox}[colback=uclablue!8, colframe=uclablue, colbacktitle=uclablue, coltitle=black, boxrule=1pt, title={\textbf{Reference Precision Instruction Example}}]
\begin{lstlisting}[style=chattemplate, language=Python, breaklines=true]
Entity type: {gt_type}

TASK: Evaluate PRECISION — using the PREDICTION as the subject, check whether the GT covers ALL factual content present in the prediction detail_description. Do not evaluate hallucination as a separate criterion; only score coverage of prediction facts by the GT.

Compare only appearance_anchor.id_features.detail_description. Ignore semantic_description, attributes, timestamp, events, shots, and context.

{field_guidance}

Scoring criteria (precision perspective — are prediction facts supported by GT?):
- Count the total number of distinct factual elements in the PREDICTION description.
- For each prediction fact, check if the GT mentions or supports it.
- A prediction fact is 'not covered' if the GT does not mention the same factual content.
- A prediction fact is 'contradicted' if the GT states the opposite; count it as not covered.

=== GROUND TRUTH detail_description (reference for verification) === {detail_description(gt_ref)}

=== PREDICTION detail_description (SUBJECT — all facts here should be verified) === {detail_description(pred_ref)}

Score 1-5:
  1 = GT covers almost none of the prediction facts (<20%)
  2 = GT covers a minority of prediction facts (20-40%)
  3 = GT covers roughly half of prediction facts (40-70%)
  4 = GT covers most prediction facts (70-90%)
  5 = GT covers essentially all prediction facts (>90%)
\end{lstlisting}
\end{tcolorbox}

\newpage
\begin{tcolorbox}[colback=uclablue!8, colframe=uclablue, colbacktitle=uclablue, coltitle=black, boxrule=1pt, title={\textbf{Reference Recall \texttt{field\_guidance}: Subject Branch}}]
\begin{lstlisting}[style=chattemplate, language=Python, breaklines=true]
For SUBJECT (PERSON and OBJECT) entities, the detail_description typically includes:
(for persons) gender, age range, ethnicity, hairstyle, facial features, body build, clothing, held objects;
(for objects) category, size, shape, material, color, texture, state; and any other distinguishing visual traits. Each of these factual elements counts as a separate coverage point.
\end{lstlisting}
\end{tcolorbox}

\begin{tcolorbox}[colback=uclablue!8, colframe=uclablue, colbacktitle=uclablue, coltitle=black, boxrule=1pt, title={\textbf{Reference Recall \texttt{field\_guidance}: Scene Branch}}]
\begin{lstlisting}[style=chattemplate, language=Python, breaklines=true]
For SCENE entities, the detail_description typically includes: location type (indoor/outdoor, specific venue), spatial layout and dimensions, lighting conditions (natural/artificial, brightness, color temperature), dominant colors and textures of surfaces (walls, floor, ceiling), key furniture/fixtures/objects present in the environment, atmosphere/weather conditions if outdoor, signage or text visible, and any other distinguishing environmental traits. Each of these factual elements counts as a separate coverage point.
\end{lstlisting}
\end{tcolorbox}

\begin{tcolorbox}[colback=uclablue!8, colframe=uclablue, colbacktitle=uclablue, coltitle=black, boxrule=1pt, title={\textbf{Reference Recall \texttt{field\_guidance}: Other Branch}}]
\begin{lstlisting}[style=chattemplate, language=Python, breaklines=true]
Focus on all concrete visual/physical attributes mentioned in the detail_description. Each distinct factual element counts as a separate coverage point.
\end{lstlisting}
\end{tcolorbox}

\begin{tcolorbox}[colback=uclablue!8, colframe=uclablue, colbacktitle=uclablue, coltitle=black, boxrule=1pt, title={\textbf{Reference Precision \texttt{field\_guidance}: Subject Branch}}]
\begin{lstlisting}[style=chattemplate, language=Python, breaklines=true]
For SUBJECT (PERSON and OBJECT) entities, the detail_description typically includes:
(for persons) gender, age range, ethnicity, hairstyle, facial features, body build, clothing, held objects;
(for objects) category, size, shape, material, color, texture, state; and any other distinguishing visual traits. Each of these factual elements counts as a separate verification point.
\end{lstlisting}
\end{tcolorbox}

\begin{tcolorbox}[colback=uclablue!8, colframe=uclablue, colbacktitle=uclablue, coltitle=black, boxrule=1pt, title={\textbf{Reference Precision \texttt{field\_guidance}: Scene Branch}}]
\begin{lstlisting}[style=chattemplate, language=Python, breaklines=true]
For SCENE entities, the detail_description typically includes: location type (indoor/outdoor, specific venue), spatial layout and dimensions, lighting conditions (natural/artificial, brightness, color temperature), dominant colors and textures of surfaces (walls, floor, ceiling), key furniture/fixtures/objects present in the environment, atmosphere/weather conditions if outdoor, signage or text visible, and any other distinguishing environmental traits. Each of these factual elements counts as a separate verification point.
\end{lstlisting}
\end{tcolorbox}

\begin{tcolorbox}[colback=uclablue!8, colframe=uclablue, colbacktitle=uclablue, coltitle=black, boxrule=1pt, title={\textbf{Reference Precision \texttt{field\_guidance}: Other Branch}}]
\begin{lstlisting}[style=chattemplate, language=Python, breaklines=true]
Focus on all concrete visual/physical attributes mentioned in the detail_description. Each distinct factual element counts as a separate verification point.
\end{lstlisting}
\end{tcolorbox}

\begin{tcolorbox}[colback=uclablue!8, colframe=uclablue, colbacktitle=uclablue, coltitle=black, boxrule=1pt, title={\textbf{Reference Context Built by \texttt{\_build\_ref\_context}}}]
\begin{lstlisting}[style=chattemplate, language=Python, breaklines=true]
Reference ID Mapping (Pred → GT): {pred_id} ({detail_description(pred_ref)}) → {gt_id} ({detail_description(gt_ref)})
\end{lstlisting}
\end{tcolorbox}

\newpage
\begin{tcolorbox}[colback=uclablue!8, colframe=uclablue, colbacktitle=uclablue, coltitle=black, boxrule=1pt, title={\textbf{Event Quality Instruction Example}}]
\begin{lstlisting}[style=chattemplate, language=Python, breaklines=true]
Event type: {gt_type}

{ref_context}

Compare only content.description. Do not judge dialogue line, speaker field, event type, time_range, shot placement, or reference ID correctness except as context for the description.

An event description typically captures: the main action/activity happening, which entities (persons/objects) are involved and their roles, spatial relationships between entities, cause-effect or sequential sub-actions, environmental/audio context relevant to the event, and the outcome or state change. Each of these factual elements is a separate coverage point.

Scoring approach:
- Identify all distinct factual claims in the GT description.
- Check which are present in the prediction (same meaning, wording differences OK).
- Contradictions (stating the opposite) are worse than omissions.
- Do not evaluate hallucination or extra prediction details; ignore facts outside the GT checklist.

=== GROUND TRUTH content.description === {gt_content.description}

=== PREDICTION content.description === {pred_content.description}

Score 1-5:
  1 = Prediction contradicts or entirely misses the core event action/subject (<20% GT facts covered)
  2 = Prediction captures the broad category but misses or contradicts major factual elements (20-40%)
  3 = Prediction captures the main gist but omits or gets wrong several important details (40-70%)
  4 = Prediction covers most GT facts accurately; only minor details are missing or imprecise (70-90%)
  5 = Prediction faithfully covers essentially all GT factual content (>90%)
\end{lstlisting}
\end{tcolorbox}

\begin{tcolorbox}[colback=uclablue!8, colframe=uclablue, colbacktitle=uclablue, coltitle=black, boxrule=1pt, title={\textbf{Subshot Formatting and \texttt{task\_text} Inserted by the Code}}]
\begin{lstlisting}[style=chattemplate, language=Python, breaklines=true]
mark = " [TARGET]" if focus_key and key == focus_key else "" lines.append( f"- id={key}{mark}; shot={sub.get('_shot_id', '')}; " f"time={sub.get('sub_shot_time', [])}: {sub.get('sub_shot_content', '')}" )

task_text = ( "Task: decide whether the TARGET PREDICTION subshot matches at least one GT candidate in the context." if direction == "precision" else "Task: decide whether at least one prediction candidate matches the TARGET GT subshot." )
\end{lstlisting}
\end{tcolorbox}

\begin{tcolorbox}[colback=uclablue!8, colframe=uclablue, colbacktitle=uclablue, coltitle=black, boxrule=1pt, title={\textbf{Subshot Match Instruction Example}}]
\begin{lstlisting}[style=chattemplate, language=Python, breaklines=true]
{ref_context}

{task_text} A match should preserve the main entities, visible state, action/pose, and local scene facts. Use [TARGET] to identify the item being evaluated. Ignore minor wording differences and slight granularity shifts. Reject candidates that only share a broad scene, are merely adjacent in time, or contradict the target's core subject/action/state. Do not evaluate hallucination or extra details; this binary decision is only about whether the target facts are covered by at least one candidate.

=== GROUND TRUTH local visual states === {gt_text}

=== PREDICTION local visual states === {pred_text}

Output exactly one decision: <match>1</match> if the target has at least one true counterpart in the opposite context <match>0</match> if no candidate matches, or the best candidate is only generic/adjacent
\end{lstlisting}
\end{tcolorbox}

\newpage
\begin{tcolorbox}[colback=uclablue!8, colframe=uclablue, colbacktitle=uclablue, coltitle=black, boxrule=1pt, title={\textbf{Subshot Quality Instruction Example}}]
\begin{lstlisting}[style=chattemplate, language=Python, breaklines=true]
{ref_context}

Compare only local visual state text: sub_shot_content. Use ids/times only to understand which item is the target and nearby context; do not score timing accuracy here.

{scoring_text}

Scoring approach:
- Identify all distinct factual claims in the TARGET item.
- For each fact, check if the opposite-side context contains it (same meaning, wording OK).
- Contradictions count as not covered.
- Nearby context items help disambiguate but are not themselves scored.
- Do not evaluate hallucination or extra details outside the target checklist.

=== GROUND TRUTH sub_shot_content context === {gt_text}

=== PREDICTION sub_shot_content context === {pred_text}

Score 1-5:
  1 = Target local state is essentially not hit; core subject/action/state is absent or contradicted (<20% facts covered)
  2 = A small fraction of target facts are hit (20-40%), but core subject/action/state is missing or wrong
  3 = Core state is partially hit (40-70%), with important target facts missing or inaccurate
  4 = Most target facts are hit (70-90%); only minor misses or imprecision
  5 = Target facts are fully and accurately hit (>90%); at most trivial omissions
\end{lstlisting}
\end{tcolorbox}

\begin{tcolorbox}[colback=uclablue!8, colframe=uclablue, colbacktitle=uclablue, coltitle=black, boxrule=1pt, title={\textbf{Subshot Recall \texttt{scoring\_text} Inserted by the Code}}]
\begin{lstlisting}[style=chattemplate, language=Python, breaklines=true]
Score how well the prediction context covers the TARGET GT subshot facts (RECALL direction). Judge only the GT local-state facts that should be recovered. Do not evaluate hallucination or prediction-side extra details; only score coverage of the target GT facts. A GT target fact that is omitted or contradicted by the prediction context is not covered.

A sub_shot_content typically describes: which entities are visible (persons/objects with ref IDs), their current pose/action/gesture/expression, spatial positions and relationships between entities, interaction with objects or environment, motion direction/speed, and any transient visual state, for instance, a lighting change or an object appearing/disappearing. Each of these factual elements is a separate coverage point.
\end{lstlisting}
\end{tcolorbox}

\begin{tcolorbox}[colback=uclablue!8, colframe=uclablue, colbacktitle=uclablue, coltitle=black, boxrule=1pt, title={\textbf{Subshot Precision \texttt{scoring\_text} Inserted by the Code}}]
\begin{lstlisting}[style=chattemplate, language=Python, breaklines=true]
Score the TARGET PREDICTION subshot against the GT context (PRECISION direction). Judge only whether the target prediction's concrete local-state facts are supported by or accurately hit the GT context. Do not evaluate hallucination or extra target details separately; only score coverage of the target prediction facts by the GT context. A target fact that conflicts with, or fails to match, the GT context is not covered.

A sub_shot_content typically describes: which entities are visible (persons/objects with ref IDs), their current pose/action/gesture/expression, spatial positions and relationships between entities, interaction with objects or environment, motion direction/speed, and any transient visual state, for instance, a lighting change or an object appearing/disappearing. Each of these factual elements is a separate verification point.
\end{lstlisting}
\end{tcolorbox}

\newpage
\begin{tcolorbox}[colback=uclablue!8, colframe=uclablue, colbacktitle=uclablue, coltitle=black, boxrule=1pt, title={\textbf{Shot Fact Instruction Example}}]
\begin{lstlisting}[style=chattemplate, language=Python, breaklines=true]
Compare only camera and shot_transfer fields. Do not judge visual_description, active_events, references_in_shot, event content, or timing.

Camera facts include: shot scale (close-up/medium/wide/extreme), camera angle (high/low/eye-level/bird's-eye/dutch), camera movement (pan/tilt/dolly/zoom/tracking/crane/handheld/static), framing composition (rule-of-thirds/centered/off-center), depth of field (shallow/deep), focus subject, and any special techniques (rack focus, whip pan, etc.). Shot_transfer facts include: transition type (cut/dissolve/fade/wipe/match-cut), transition speed, and any visual continuity or discontinuity noted.

Scoring approach:
- Identify all distinct camera/transfer facts in the GT.
- Check which are present in the prediction (same meaning, terminology differences OK).
- Contradictions are worse than omissions.
- Do not evaluate hallucination or extra prediction details; ignore facts outside the GT camera/transfer checklist.

=== GROUND TRUTH === camera: {json.dumps(gt_camera, ensure_ascii=False)} shot_transfer: {gt_transfer}

=== PREDICTION === camera: {json.dumps(pred_camera, ensure_ascii=False)} shot_transfer: {pred_transfer}

Score 1-5:
  1 = Camera and transfer facts are mostly missing or contradicted (<20% GT facts covered)
  2 = Some overlap but major camera/transfer facts are wrong or missing (20-40%)
  3 = Partly correct; some key camera/transfer details are missing or inaccurate (40-70%)
  4 = Mostly accurate with only minor omissions or imprecision (70-90%)
  5 = Fully consistent with all GT camera and shot_transfer facts (>90%)
\end{lstlisting}
\end{tcolorbox}

\newpage


\end{document}